\documentclass[preprint,12pt,authoryear]{elsarticle}

\usepackage{amssymb}
\usepackage{amsmath}
\usepackage{xcolor}
\usepackage{tikz}
\usepackage{forest}
\usepackage{hyperref}
\usepackage{adjustbox}
\usepackage{booktabs}

\useforestlibrary{edges}
\definecolor{yellowset}{rgb}{1, 0.898, 0.6}  
\definecolor{greenset}{rgb}{0.725, 0.878, 0.647}  
\definecolor{redset}{rgb}{0.945, 0.611, 0.6}  
\definecolor{blueset}{rgb}{0.662, 0.768, 0.921}  
\definecolor{greyset}{rgb}{0.901, 0.901, 0.901}  
\setcitestyle{authoryear,round} % Example: Author (Year)

\tikzset{%
  parent/.style={align=center, text width=3cm, text ragged, rounded corners=4pt, minimum height=1cm},
  child/.style={align=center, text width=3cm, text ragged, rounded corners=4pt,  minimum height=1cm},
  root/.style={align=center, text width=7cm, rounded corners=4pt, draw, minimum height=1cm, font=\bfseries\sffamily,},
  leaf/.style={align=left}, 
  base1/.style={minimum height=4ex, minimum width=2cm,rounded corners=2pt, fill=gray!10},
}

\journal{Image and Vision Computing}

\begin{document}

\begin{frontmatter}



\title{A Survey on Dynamic Neural Networks: from Computer Vision to Multi-modal Sensor Fusion} %% Article title

\author[1]{Fabio Montello}
\ead{fabmo@dtu.dk}

\author[1]{Ronja Güldenring}
\ead{ronjag@dtu.dk}

\author[2]{Simone Scardapane}
\ead{simone.scardapane@uniroma1.it}

\author[1]{Lazaros Nalpantidis}
\ead{lanalpa@dtu.dk}

\affiliation[1]{organization={DTU Electrical and Photonics Engineering, Technical University of Denmark},
% addressline={Radarweg 29},
postcode={2800},
city={Kgs. Lyngby},
country={Denmark}}
\affiliation[2]{organization={DIET Department, Sapienza University of Rome},
% addressline={Radarweg 29},
postcode={00185},
city={Rome},
country={Italy}}

%% Abstract
\begin{abstract}
Model compression is essential in the deployment of large Computer Vision models on embedded devices. However, static optimization techniques (e.g. pruning, quantization, etc.) neglect the fact that different inputs have different complexities, thus requiring different  amounts of computations. Dynamic Neural Networks allow conditioning the number of computations to the specific input. The current literature on the topic is very extensive and fragmented. We present a comprehensive survey that synthesizes and unifies existing Dynamic Neural Networks research in the context of Computer Vision. Additionally, we provide a logical taxonomy based on which component of the network is adaptive: the output, the computation graph or the input. Furthermore, we argue that Dynamic Neural Networks are particularly beneficial in the context of Sensor Fusion for better adaptivity, noise reduction and information prioritization. We present preliminary works in this direction.  We complement this survey with a curated repository listing all the surveyed papers, each with a brief summary of the solution and the code base when available: \href{https://github.com/DTU-PAS/awesome-dynn-for-cv}{https://github.com/DTU-PAS/awesome-dynn-for-cv}.
\end{abstract}

% \begin{figure}[!t]
%     \centering
%     \includegraphics[width=\textwidth]{techniques_horizontal.png}
%     % \caption{\textbf{The three types of Dynamic Neural Networks} we consider in the survey. From left to right: (a) Early Exits networks decide at which point to output, (b) Dynamic routing networks use a Mixture-of-Experts and decide which computational path is optimal according to the input, (c) Token skimming networks decide which subset of tokens will attend the following blocks.}
%     % \label{fig:pagewidthimage}
% \end{figure}

% \begin{highlights}
% \end{highlights}

%% Keywords

\end{frontmatter}

% Introduction
\section{Introduction}
\begin{figure}[!t]
    \centering
    \includegraphics[width=\textwidth]{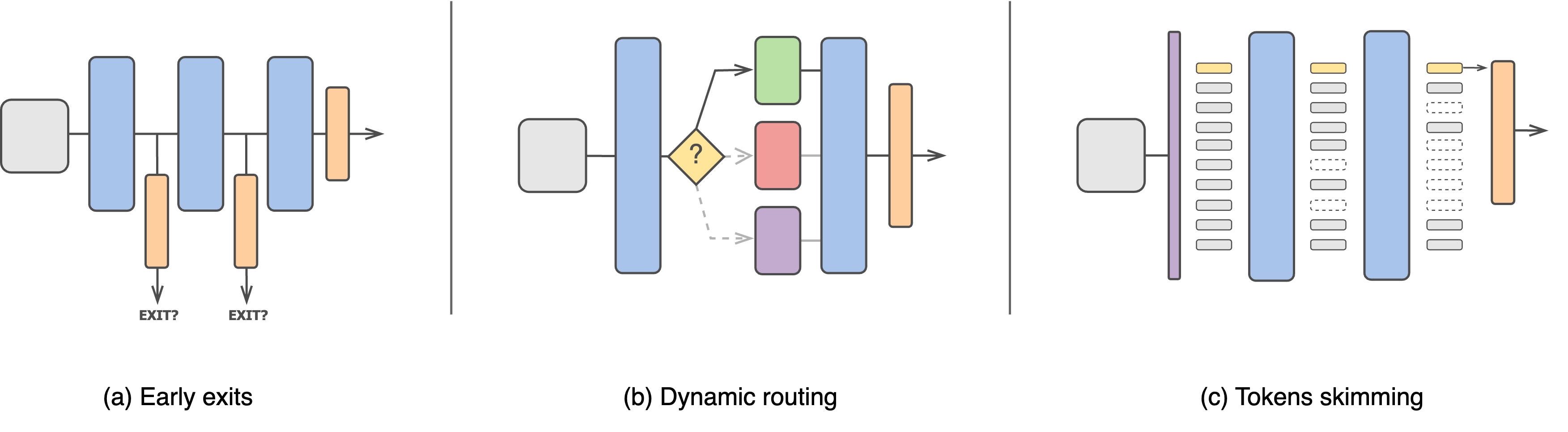}
    \caption{\textbf{The three types of Dynamic Neural Networks} we consider in the survey. From left to right: (a) Early Exits networks decide at which point to output, (b) Dynamic routing networks use a Mixture-of-Experts and decide which computational path is optimal according to the input, (c) Token skimming networks decide which subset of tokens will attend the following blocks.}
    \label{fig:pagewidthimage}
\end{figure}

\begin{figure}[!t]
    \centering
    \includegraphics[width=0.92\textwidth]{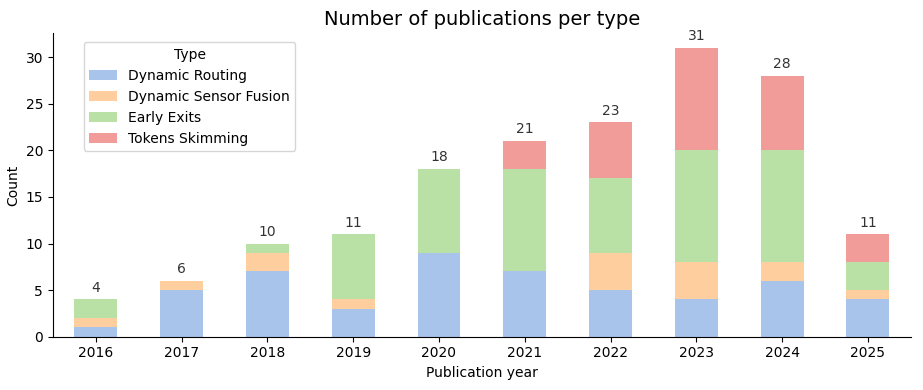}
    \caption{\textbf{Overview of the publications} considered in this survey, grouped by year and topic. In total, 163 publications have been reviewed: 65 for the Early Exits Section, 51 for the Dynamic Routing Section, 31 for the Token Skimming Section, and 16 in the Dynamic Sensor Fusion Section.}
    \label{fig:trenddynncvpapers}
\end{figure}

In the past decade, the introduction of architectural paradigms such as Convolutional Neural Networks (CNNs) and Vision Transformers (ViTs) has enabled substantial advancements in the Computer Vision field. Efficiency then becomes an important factor for deploying these neural networks---especially in the foundation model era. Model compression methods like quantization \citep{gholamiSurveyQuantizationMethods2021}, pruning \citep{chengSurveyDeepNeural2024}, knowledge distillation \citep{gouKnowledgeDistillationSurvey2021}, neural architecture search \citep{whiteNeuralArchitectureSearch2023}, and low-rank factorization \citep{swaminathanSparseLowRank2020} have demonstrated how Neural Networks are often unnecessarily overparameterized  \citep{dengModelCompressionHardware2020}. On the other hand, these optimization techniques do not take into account the different amount of computations needed for each input, disregarding the fact that different inputs have different complexities. Dynamic Neural Networks allow one to work around this problem by conditioning the computation path of a network to its input---refer to Figure \ref{fig:pagewidthimage} for an overview of three different types of Dynamic Neural Networks which are the focus of this survey paper. 
% Biological relation
A parallel can be drawn with the human perception system, which is able to adaptively allocate time and scrutiny for visual recognition, where simple glimpses are sufficient to recognize simple objects and scenes, while more time and attention are required in cases where the elements are occluded or complicated \citep{schiebenerIntegratingVisualPerception2013, waltherFMRIDecodingNatural2010}. \\
% Definition of Dynamic Neural Network
A Dynamic Neural Network is defined as a deep learning model that is able to adjust the number of computations according to the complexity of the input \citep{hanDynamicNeuralNetworks2021a, xuSurveyDynamicNeural2023a}. This is obtained by introducing some elasticity within the network (usually gating or decision modules), which allows one to differentiate between \textit{easy} and \textit{hard} samples and take a computation path decision accordingly. In some cases, the complexity estimation can also be steered by other external factors such as the overall computational budget or the environmental conditions \citep{wangSEESchedulingEarly2019, malawadeEcoFusionEnergyawareAdaptive2022, devotoAdaptiveSemanticToken2024}.\\
% Intro + biological explaination sensor fusion
Furthermore, we argue that the adaptability of Dynamic Neural Networks is a particularly beneficial property in the context of Sensor Fusion---however, this is currently an underexplored research area. Studies suggest that, in the human brain, distinct parts are highly specialized in different tasks \citep{kandelPrinciplesNeuralScience}, e.g. spatial information and object information are processed separately \citep{goodaleSeparateVisualPathways1992}. Furthermore, studies on different living organisms have shown how the loss of a sense gets compensated for by remaining sensory modalities \citep{rabinowitchNeuropeptideDrivenCrossModalPlasticity2016}.
% Definition of sensor fusion
In Computer Vision, Multi-modal Sensor Fusion can be defined as the aptitude of a model to combine, interpret, and reason on information coming from different sensors and express various aspects of the same environment, aiming at a better perception of the surroundings in different contexts and conditions \citep{fungSensorFusionReview2017}. \\
% Why dynn can be beneficial in sensor fusion
We observe that Sensor Fusion models can strongly benefit from the use of adaptive techniques for multiple reasons. Firstly, they allow sample-aware model efficiency. Furthermore, the learned adaptivity produces robustness against the noise generated by non-optimal environmental conditions in which the sensors are exposed. Finally, there is an intrinsic ability of these techniques to identify critical information by design. This allows organizing the processing according to input importance and computation budget constraints.
% The trend of paper is skyrocketing (explained with figures), but something is missing
As can be seen in Figure \ref{fig:trenddynncvpapers}, Dynamic Neural Networks are gaining traction, with the number of papers steadily increasing year by year. While---as will be discussed later in this section---existing survey papers have covered similar directions, there is still a lack of a survey paper and a clear taxonomy that focuses on Computer Vision.\\
% What we present
Thus, we present a survey of Dynamic Neural Networks in the context of Computer Vision, outlining the most representative methods. We categorize concepts, methods, and techniques in this field through the structuring of a clear taxonomy so that it becomes easier to  understand and navigate the field. Our goal is to create a standardized language that supports the exploration of methodologies and applications in the domain through a comparison of recent advancements.

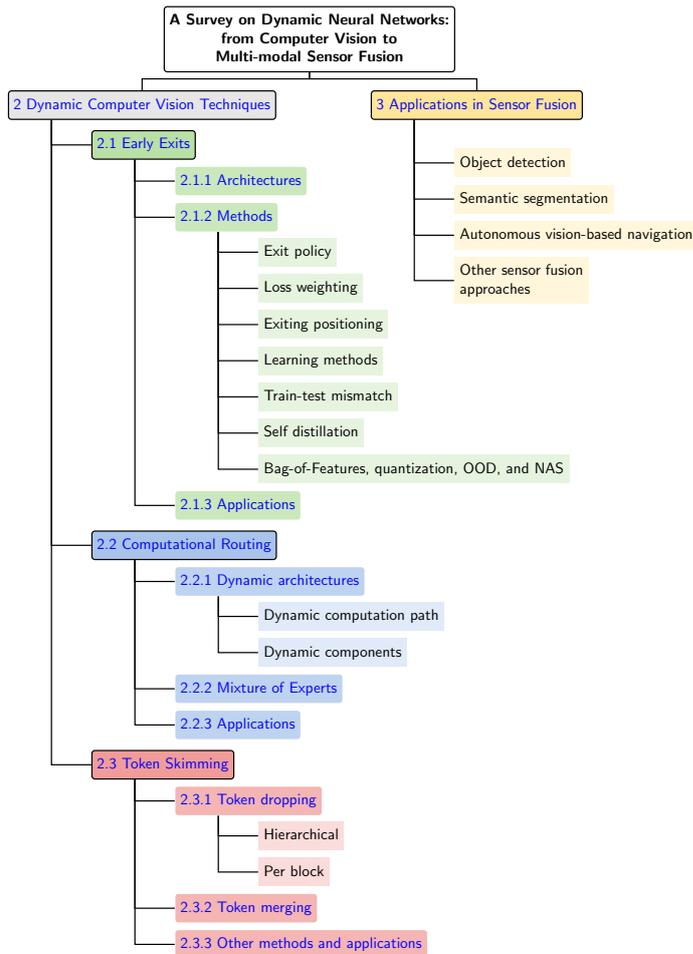
\begin{figure}[!ht]
  \centering
  \resizebox{0.7\textwidth}{!}{%
    \begin{forest}
        forked edges, folder indent=1cm,
        where={level()<1}{}{folder, grow'=east, align=center},
        where={level()>0}{l sep+=1.5cm}{},
        for tree={
            fork sep=1.5mm,
            thick, edge=thick,
            font=\sffamily,
            align=left, % Center-align all node text
            if n children=0{
                if level=2{yshift=-5mm}{},
                for parent={s sep=1.5mm}
            }{
                draw , base1
            }
        }
        [A Survey on Dynamic Neural Networks: \\ from Computer Vision to\\ Multi-modal Sensor Fusion, calign=edge midpoint, s sep=2cm, font=\bfseries\sffamily, fill=white, align=center, draw
            [{\hyperref[section:dycv]{\ref{section:dycv} Dynamic Computer Vision Techniques}}, fill=greyset
                [{\hyperref[section:earlyexits]{\ref{section:earlyexits} Early Exits}}, fill=greenset
                  [{\hyperref[subsubsec:architecture]{\ref{subsubsec:architecture} Architectures}}, base1, fill=greenset!75]
                  [{\hyperref[subsubsec:method]{\ref{subsubsec:method} Methods}}, draw=none, base1, fill=greenset!75, 
                        [\hyperlink{para:exitpolicy}{Exit policy}, fill=greenset!35]
                        [\hyperlink{para:lossweighting}{Loss weighting}, fill=greenset!35]
                        [\hyperlink{para:exitpositioning}{Exiting  positioning}, fill=greenset!35]
                        [\hyperlink{para:learningmethods}{Learning methods}, fill=greenset!35]
                        [\hyperlink{para:traintestmismatch}{Train-test mismatch}, fill=greenset!35]
                        [\hyperlink{para:selfdistillation}{Self distillation}, fill=greenset!35]
                        [\hyperlink{para:bof}{Bag-of-Features{,} quantization{,} OOD{,} and NAS}, fill=greenset!35]
                      ]
                  [{\hyperref[subsubsec:applications]{\ref{subsubsec:applications} Applications}}, base1, fill=greenset!75]
                ]
                [{\hyperref[section:routing]{\ref{section:routing} Computational Routing}}, fill=blueset
                  [{\hyperref[subsubsec:dynamicarchitecture]{\ref{subsubsec:dynamicarchitecture} Dynamic architectures}}, base1, draw=none, fill=blueset!75
                    [\hyperlink{para:dyncomputationpath}{Dynamic computation path}, fill=blueset!35]
                    [\hyperlink{para:dyncomponents}{Dynamic components}, fill=blueset!35]
                  ]
                  [{\hyperref[subsubsec:mixtureofexperts]{\ref{subsubsec:mixtureofexperts} Mixture of Experts}}, base1, fill=blueset!75]
                  [{\hyperref[subsubsec:routingapplications]{\ref{subsubsec:routingapplications} Applications}}, base1, fill=blueset!75]
                ]
                [{\hyperref[section:skimming]{\ref{section:skimming} Token Skimming}}, fill=redset
                  [{\hyperref[subsubsec:dropping]{\ref{subsubsec:dropping} Token dropping}}, base1, draw=none, fill=redset!75,
                      [\hyperlink{para:hierarchical}{Hierarchical}, fill=redset!35]
                      [\hyperlink{para:perblock}{Per-block}, fill=redset!35]
                  ]
                  [{\hyperref[subsubsec:merging]{\ref{subsubsec:merging} Token merging}}, base1, fill=redset!75]
                  [{\hyperref[subsubsec:skimother]{\ref {subsubsec:skimother} Other methods and applications}}, base1, fill=redset!75]
                ]
            ]
            [{\hyperref[section:sensorfusion]{\ref{section:sensorfusion} Applications in Sensor Fusion}}, fill=yellowset,
                [\hyperlink{para:objectdetection}{Object detection}, fill=yellowset!35]
                [\hyperlink{para:semanticsegmentation}{Semantic segmentation}, fill=yellowset!35]
                [\hyperlink{para:autonomous}{Autonomous vision-based navigation}, fill=yellowset!35]
                [\hyperlink{para:othersensor}{Other sensor fusion}\\\hyperlink{para:othersensor}{approaches}, fill=yellowset!35]
            ]
        ]
    \end{forest}
  }
  \vspace{8pt}
  \caption{\textbf{Taxonomy of Dynamic Neural Network techniques presented in this survey}, categorized by application domain (Computer Vision and Sensor Fusion) and specific method. The diagram highlights key methods such as Early Exits (in green), Computational Routing (in blue), Token Skimming (in red), and their applications in various Sensor Fusion tasks (in yellow). }
  \label{fig:dynamic_nn_tree}
\end{figure}

\paragraph{Taxonomy}
The full tree structure of the paper is shown in Figure \ref{fig:dynamic_nn_tree}. For better readability and reachability, we gathered the papers related to Computer Vision methods and applications in the first part of this survey, and those related to Sensor Fusion applications in the second half.
In Section \ref{section:dycv} we introduce the advancements of Dynamic Neural Network techniques in the scope of Computer Vision, dividing them into three logical splits according to where the dynamicity is introduced: 
\begin{enumerate}
\renewcommand{\labelenumi}{(\roman{enumi})}
\item In Early Exits (Section \ref{section:earlyexits}), we discuss those networks that take decisions on where the output is returned within the architecture. We show here the most important solutions, then we present methods that improve certain aspects of Early Exiting and we conclude by introducing some papers that present the application of these methods to specific tasks and use cases. 
\item In Dynamic Routing (Section \ref{section:routing}), we explore all those techniques that work directly on a dynamic computational path, either by constructing it on the fly or routing the computation to different model blocks according to the input needs (as is in the case of Mixture-of-Experts). The last part in this Section is dedicated to the application of these methods to various tasks. 
\item In Token Skimming (Section \ref{section:skimming}), we focus on models that reduce the input to each block of layers within a network. This chapter focuses on Transformer models, and it is split between techniques that drop parts of the tokens and techniques that decide to merge together less significant tokens. Last, we present other methods and the applications of skimming to tasks that differ from image classification.
\end{enumerate}
In Section \ref{section:sensorfusion}, we present publications that apply these types of techniques in the context of Sensor Fusion. We first delve into why this branch of vision can particularly benefit from the introduction of adaptability in the fusion architecture, and then we present the solutions grouped by task.

\paragraph{Scope}
This survey discusses Dynamic Neural Networks only in the context of Computer Vision and their application in Sensor Fusion, independently of the task that the paper tackles. We do not cover papers coming from other fields in deep learning (e.g. Natural Language Processing, Graph Neural Networks), although some of those might be mentioned when important in the context of the discussion. Furthermore, we considered only papers from the last 9 years, i.e., from January 2016 to September 2025.

\paragraph{Related surveys}
This paper is related to existing surveys: \citet{scardapaneWhyShouldWe2020a} is the first survey in the field to set the base for a theoretical description of Early Exits Neural Networks, unifying different methodologies under a common formulation. \citet{scardapaneConditionalComputationNeural2024a} updates and generalizes this formalization also to other Dynamic Neural Network approaches, i.e. Mixture-of-Experts and Token sparsification. 
In contrast, our survey takes a complementary approach where we focus on analyzing all relevant papers on the topic, rather than a theoretical unification of the field. The goal is to provide a more complete insight into all the available ideas and applications of these methods.
\citet{chenChasingSparsityVision2021a} explores sparsity in ViT models, focusing on Token Skimming approaches, while \citet{farinaSparsityTransformersSystematic2024a} is a more recent survey on the same topic. \citet{hanDynamicNeuralNetworks2021a} discusses advancements in Dynamic Routing techniques, focusing mostly on the dynamicity of the computational path. All these survey papers highlight the most relevant works in both fields of Natural Language Processing and Computer Vision. On the other hand, in this survey, we deliberately consider only the Computer Vision field, in order to provide a more in-depth overview of all relevant publications and not only major milestones. \citet{matsubaraSplitComputingEarly2022a} provides a survey that puts in relation Early Exits models with computation splitting between on-device and cloud resources. In \citet{xuSurveyDynamicNeural2023a}, a survey on Dynamic Neural Networks for Natural Language Processing is presented. This paper shares a similar structure with ours but focuses on NLP and presents a limited selection of publications. We provide a survey that is complementary but focuses on Computer Vision papers.

Finally, to the best of our knowledge, no existing survey comprehensively examines the application of Dynamic Neural Networks in the context of Sensor Fusion. Since we consider this a relevant and emerging topic, we dedicate a chapter specifically to all the relevant publications in this subtopic.

 \paragraph{Contribution} Unlike existing surveys, our work offers a comprehensive overview of Dynamic Neural Networks for Computer Vision, while simultaneously connecting them to the crucial field of Sensor Fusion. Our primary contributions are summarized as follows: 
\begin{itemize} 
    \item We carry out a comprehensive and up-to-date survey of Computer Vision methods that dynamically adjust their computational load based on input complexity. In total, we review 163 publications relevant to the domain.
    \item We propose a clear taxonomy categorizing methods based on the adaptive network component. Specifically, we group approaches by where the network yields \textit{output} (Early Exits), how it constructs its \textit{computation path} (Dynamic Routing), or which portion of the \textit{input} is processed (Token Skimming). 
    \item We highlight that Dynamic Neural Network strategies are particularly advantageous for Sensor Fusion. We discuss the seminal works in this area, organizing them according to the specific task they address. 
    \item We propose a standardized evaluation template to address the critical lack of unified benchmarking protocols in the current literature, which aims at providing a fair and rigorous cross-methodology comparison methodology for future publications.
    \item Finally, we provide a selected online repository listing all referenced papers, accompanied by brief descriptions and links to the corresponding codebases (where available) for easy reference. The repository is available at: \href{https://github.com/DTU-PAS/awesome-dynn-for-cv}{https://github.com/DTU-PAS/awesome-dynn-for-cv}. 
\end{itemize}

\section{Dynamic Computer Vision Techniques} \label{section:dycv}

\subsection{Early exits} 
\label{section:earlyexits}
The idea underlying of the development of Early Exits networks is rooted in the assumption that samples in a dataset can vary a lot in complexity. Following this principle, traditional Neural Networks tend to be overparametrized for most of the samples, which could potentially be processed with just a fraction of the parameters used.
Furthermore, in the case of \textit{easy} samples, Convolutional Neural Networks (CNNs) have been shown to be capable of extracting features in the early layers that are already sufficient for a confident classification \citep{pandaConditionalDeepLearning2016}. For these reasons, Early Exits networks propose to evaluate the confidence of the model in the prediction of output at different stages within the network, with the option of exiting the computation path earlier, if the model is confident enough in the prediction. This has a twofold purpose: it allows to optimize the computational cost of the model tailored to the input difficulty, and moreover, it avoids the phenomenon of overthinking, a circumstance in which a correct prediction in the early stages of a network gets corrupted by noise in deeper layers \citep{kayaShallowDeepNetworksUnderstanding2019}. The confidence evaluation of a model is carried out by attaching a number of auxiliary classifiers to the intermediate blocks of the network. Section \ref{subsubsec:method} further illustrates different methods for the determination of the number and position of the exit points, and how to properly construct and evaluate the prediction confidence score of a network in the early stages.
Before exploring the major contributions in the field, we present here a summary of the formulation of the Early Exits models presented in \citet{scardapaneWhyShouldWe2020a} and \citet{scardapaneConditionalComputationNeural2024a}, to which we refer for more details.\\
A Neural Network can be seen as a composition of $b$ sequential blocks:
\begin{equation}
    f(x) = f_b \circ f_{b-1} \circ \cdots \circ f_1(x) 
\end{equation}
where $f_i$ is the $i$-th block of the network with $i \in\{0,1...,b\}$, $f_b$ is the final output of the network and $\circ$ denotes the function composition. We can produce an Early Exits model by attaching an arbitrary amount of auxiliary classifiers. For the sake of simplicity, an example with a classifier after the $i$-th network block can be written as:
\begin{equation}
    y_i(x) = c_i \circ (f_i \circ f_{i-1} \circ \cdots \circ f_1(x))
\end{equation}
where $c_i$ is the classifier attached to the $i$-th block. Generally speaking, in a classification network, an Early Exits auxiliary module is composed of a pooling operation to reduce the feature size followed by a Multi-Layer Perceptron (MLP) which produces the classification output. The design of the classifier is strongly dependent on the network architecture. For CNNs, especially in the early stages, where the feature maps resolution is still large, an ad-hoc downsampling step is required, followed by the MLP head for classification. In the case of ViTs, since the token dimension remains constant throughout the network, the same MLP head or multiple MLP heads of the same shape can be reused.
The most common practice is to train Early Exits models by aggregating the final loss $\lambda_F$ with the losses $\lambda_i$  computed at all the intermediate intervals. The optimization objective then becomes: 
\begin{equation} \label{eq:objective}
    f^* = \arg\min \left\{ \lambda_F + \sum_{i=1}^{b-1}  \alpha_i \lambda_i \right\}
\end{equation}
where $\alpha_i$ weights the contribution of each auxiliary classifier to the overall loss. 
A popular exit policy solution is the one proposed by \citet{teerapittayanonBranchyNetFastInference2016a}. It consists of calculating an entropy score:
\begin{equation} \label{eq:entropy}
H(\textbf{y}) = \sum_{c \in C} y_c \log \frac{1}{y_c}
\end{equation}
where $\textbf{y}$ is a vector containing computed probabilities $y_c$ for all possible class labels and $C$ is the set. The exiting decision is then based on the score as follows:
\begin{equation} \label{eq:exitdecision}
d = 
\begin{cases} 
1, & \text{if } H(\textbf{y}) \leq \tau \\
0, & \text{if } H(\textbf{y}) > \tau
\end{cases}
\end{equation}
\\
where $\tau$ is a defined threshold for entropy and $d$ is the decision variable; $d=0$ indicates an exit and $d=1$ indicates a continuation of the computations.\\
We proceed now to illustrate the major contributions in the field of Early Exits models. Section \ref{subsubsec:architecture} presents the key architectures that have been introduced in the literature, Section \ref{subsubsec:method} discusses papers that try to improve specific aspects of the learning process of these types of networks, while Section \ref{subsubsec:applications} focuses on applications to other tasks rather than classification.\\
Although this survey considers Dynamic Neural Networks only in the context of Computer Vision, the theoretical formulation of Early Exits is not restricted to visual tasks. The idea at the base of these models—that different inputs have different complexities—applies equally to other fields, such as Natural Language Processing (NLP). Consequently, the methods proposed to compute the confidence score, like the entropy calculation of a classifier's output, are independent of the data modality. Furthermore, solutions designed to optimize the exit policy or bridge the train-test mismatch gap present direct cross-domain applicability. For example, formulating the sequential early-exit decision criterion as a Markov Decision Process, or treating the stopping time as a latent variable within a variational Bayes problem, relies entirely on mathematical constructs independent of the input modality.\\
% It should be noted that numerical results for the presented publications are not presented for this section, as the studies report on different datasets and evaluation architectures, which makes direct comparison infeasible.
It should be noted that numerical results for the presented publications are not provided in this section. The domain of Early Exits lacks a standardized experimental protocol. We report on studies whose findings are based on a diverse array of datasets and evaluation architectures. A direct and fair quantitative analysis infeasible without extensive re-implementation. This heterogeneity highlights a critical gap in the current literature: the necessity for a structured benchmark. We argue that such a benchmark is essential to consistently compare different Dynamic Neural Network models, specifically accounting for inference speed and efficiency, the variability of computations in relation to input complexity, and performance quality in edge cases, such as small objects or targets located in the peripheral regions of the image. A further discussion on the topic is presented in the Discussion (Section \ref{section:discussion}).
%%%%%%%%%%%%%%%%%%%%%%%%%%%%%%%%%%%

\subsubsection{Architectures}
\label{subsubsec:architecture}
\begin{figure}[t]
    \centering
    \includegraphics[width=0.7\textwidth]{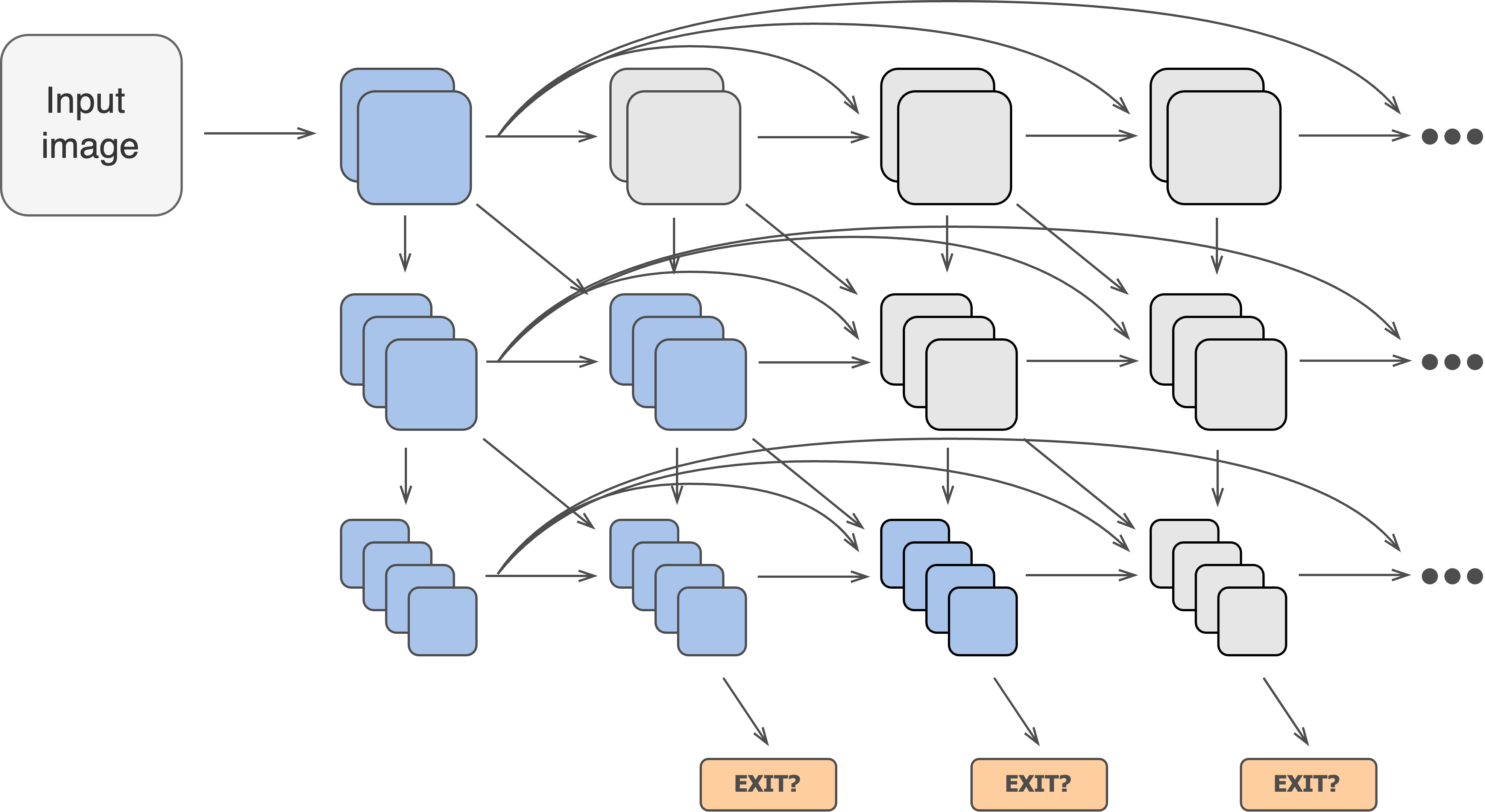}
    \vspace{8pt}
    \caption{ \textbf{Illustrative example of MSDNet (Multi-Scale Dense Network)}. MSDNet processes the input image through a multi-scale architecture, enabling feature extraction at various resolutions.  The feature maps shown in blue and classifiers highlighted in orange are active during this example, while the grey blocks indicate components that are not in use.
    For the complete scheme, please refer to \citet{huangMultiScaleDenseNetworks2018}.}
    \label{fig:msdnet}
\end{figure}

Following the principle that only a small fraction of inputs require the full computational effort of a network, while a large majority of the data samples can be classified correctly even in the very early stages of a network, \citet{pandaConditionalDeepLearning2016} is the first paper to inject Early Exits classifiers into a pretrained network. It takes a trained network and attaches linear classifiers at each layer, to establish whether the layer is good enough to produce a correct classification output with confidence. The paper also includes a study on how to determine a good threshold to stop the computations.\\
BranchyNet \citep{teerapittayanonBranchyNetFastInference2016a} improves on top of \citet{pandaConditionalDeepLearning2016} by presenting an end-to-end training procedure for the Neural Network: instead of having classifiers attached to the already trained network, this solution proposes to train the whole network end-to-end, with exit points composed of MLP branches for classification along the architecture. The network is trained by optimizing the sum of the loss functions associated with the classification at each exit point. The model uses an entropy score of a classifier's output (see equation \eqref{eq:entropy}) as a measurement of the model's confidence in having a correct prediction. The contribution of each loss to the overall loss in \citep{teerapittayanonBranchyNetFastInference2016a} is set manually, and the authors observe that weighting more early branches encourages a more discriminative feature learning in the early layers of the network, allowing more samples to exit with high confidence. A simplified schematic of the architecture can be seen in Figure \ref{fig:pagewidthimage}(a). Other methods for computing the confidence score in different ways are discussed in more detail in Section \ref{subsubsec:method}. \citet{kayaShallowDeepNetworksUnderstanding2019} proposes a slight variation with respect to the original BranchyNet paper, where the pre-trained weights are frozen and only the Early Exits classifiers are trained. \citet{jieAnytimeRecognitionRouting2021} adopts a CNN-based architecture that allows each sample to adaptively select its own optimal exit within a specific time budget. At each exit, there is a module trained through deep Reinforcement Learning which predicts the optimal exit for each sample.\\
\citet{huangMultiScaleDenseNetworks2018} presents MSDNet, a new architecture that builds on top of DenseNet, with each layer working on multiple scales by maintaining multiple filter sizes of diminishing spatial dimensions, but growing depth. Figure \ref{fig:msdnet} presents the overall schematic of the architecture. Within a layer, features are produced for all scales by iteratively striding the fine feature to a coarser level. This process is repeated for each layer of the network, facilitating correct classifications even in the early layers, but also extracting low-level features that can become useful after several layers. The output of each feature is propagated in the respective scale feature in the next layer, as well as downsampled to be passed to coarser classifiers in the next layer. Intermediate classifiers are appended throughout the network to perform Early Exits strategies, while dense connections are also incorporated to prevent the classifiers from altering internal representations. 
% Insert a page-width image
\citet{yangResolutionAdaptiveNetworks2020} proposes some variation in the implementation with respect to MSDNet. Compared to MSDNet, the dense blocks with the smallest scale input are taken into account first and processed throughout all depths of the network. The processing at finer scales happens at a later stage. The key idea is that this low-scale resolution is sufficient to classify easy samples better in the early stages, whereas higher resolution is beneficial only for more difficult inputs later on, which are also more computationally costly. Also, \cite{addadCHASEChannelWiseSpatial2025} further explores the use of MSDNet architecture for early exits by improving the feature representation and aggregation based on self-attention.\\

More recently, with the advent of Transformer-based architectures, \citet{wangNotAllImages2021a} proposes a framework named Dynamic Vision Transformer (DVT) which aims at finding the correct patch size for tokenization at which the image has to be processed to guarantee a confident output. In the Transformer architecture, the complexity of the model is quadratic with respect to the number of tokens that attend the attention block. For this reason, a lower number of tokens translates into a much faster model. The paper trains a cascade of Transformers, where each model uses a larger number of smaller tokens at each scale. In this way, images are processed at a coarse level at first with a low number of tokens, and then more details are processed at each iteration including a larger number of tokens derived from smaller patches. The model exits in the case where the prediction output is sufficiently confident. The paper presents also feature-wise and relationship-wise reuse mechanisms to reduce redundant computations already performed by the layer that processed the coarser tokens.\\
\citet{tangDynamicTokenPruning2023a} proposes an alternative method to perform Early Exits of tokens in the ViT for semantic segmentation. After each Transformer block in the encoder, an auxiliary head determines which of the tokens can be halted and passed through the decoder, and which need more computations. The paper further discusses the case where all the tokens of a given \textit{easy} category get stopped in the early stages, losing the possibility to provide contextual information in the attention operation to the tokens of other classes. To avoid this, for each category, at least one token and at most top-$k$ of them are kept. This paper presents an approach that is in-between Token Skimming and Early Exits, which is possible only due to the repetitive nature of the Tranformer architecture. \\
\citet{hanDynamicPerceiverEfficient2023} proposes a novel architecture called Dynamic Perceiver, which aims at decoupling the feature extraction procedure and the early classification task with a dual-branch architecture. In this model, one branch is dedicated only to the extraction of image features, while a second branch processes the extracted features by performing classification tasks. The two branches progressively exchange information down the network via two specifically designed modules: the feature-to-latent cross-attention and the latent-to-feature cross-attention. A token mixer module is also inserted in the classification branch with the purpose of reducing the token length and expanding the hidden dimension of the latent vectors. A self-distillation loss is used in combination with cross-entropy loss to further improve the accuracy.\\
Finally, \citet{xueAdaptiveComputationElastic2023} introduces a different architecture paradigm type called AdaTape. In this case, the paper presents a foundation model based on a Transformer architecture which has also the ability to read and store tokens from a tape bank which can be either trainable or derived. The dynamic reading from the bank is regulated through a proposed dynamic halting algorithm, called Adaptive Tape Reading.

%%%%%%%%%%%%%%%%%%%%%%%%%%%%%%%%%%%

\subsubsection{Methods}
\label{subsubsec:method}
Different methods have been proposed in the literature to improve the learning process of Early Exits networks, targeting their exit policy, loss weighting, exit positioning, used learning approach, train-test mismatches, self distillation, as well as Bag-of-Features, Out-of-Distribution and Neural Architecture Search aspects. 

\paragraph{Exit policy} 
\hypertarget{para:exitpolicy}{}\label{para:exitpolicy}
The exit policy is the logical component within an Early Exits network which is in charge of determining whether the model has processed the input enough to provide an intermediate output.
\citet{scardapaneDifferentiableBranchingDeep2020a} proposes a method to optimize the Early Exits decision jointly during training by adding a \textit{soft} gate at each exit that decides a weighting strategy. The idea is to learn a weighting scheme to combine the output of the current prediction with the (supposedly improved) output of the next exit's prediction. Essentially, the key idea is for the network to self-weight how much it should rely on the output of the current exit with respect to the output of the next Early Exit. This strategy allows training the model by exploiting a cross-entropy loss with respect to the final output solely; thus, it does not need to create a loss that is a weighted combination of all the losses computed on the intermediate exits. During inference, the added auxiliary gating module can be considered a standard binary classifier and can be thresholded to evaluate the probability of exit. Furthermore, the paper proposes a regularization term for the final loss that induces the training process to take into account also the average computational cost of the network. Similarly, in \citet{chenLearningStopLearning2020} the same idea of learning the stopping policy for Early Exits together with the network parameters  is considered. In this case, the solution views the stopping time as a latent variable conditioned on the input, interpreting the problem as a variational Bayes one. The training procedure to achieve this consists of decomposing the problem in two stages, necessary for learning both the predictive model and the stopping policy. In the first stage the model parameters are learned, assuming that the stop time always follows the best stopping distribution. In the second stage, the exit policy is learned by minimizing the reverse Kullback–Leibler divergence through a Reinforcement Learning algorithm.\\
\citet{tanEmpoweringAdaptiveEarlyExit2021} works on the threshold determination problem proposing an approximate formulation for finding the accuracy-maximum threshold setting that meets a given average latency requirement. Furthermore, a threshold determination method is proposed to solve the formulated non-convex problem. The paper presents a theoretical proof, under certain parameter settings, of an approximate stationary point of the formulated problem, which is then also verified empirically.\\
\citet{pomponiProbabilisticReIntepretationConfidence2022} proposes a method to train jointly all the branches of a multi-exit model by weighting the predictions from each branch with a trained confidence score. Each confidence score is an approximation of the actual confidence score produced by the branch, which is calculated and regularized while training the rest of the model. The key idea is to allow the network to select the best Early Exits for a given input by associating to each auxiliary classifier an additional confidence score of exiting. The paper contribution consists in reinterpreting the output of the network probabilistically adding uncertainty quantification during training. Each confidence score is used as a parameter for a continuous relaxation of the Bernoulli distribution. Based on a similar idea, EERO \citep{valadeEEROEarlyExit2024a} is a methodology that revisits the problem of Early Exiting using multiple classifiers with a reject option to better select the exiting head for each instance. The problem is formalized as a classification with a reject option and establishes the optimal rule for a two-head scenario, deriving a data-driven procedure to define the rejection threshold. The probabilities of exiting at the different heads are calibrated using aggregation with exponential weights, with the aim of guaranteeing a fixed budget. Translated into the classification with reject option vocabulary, the question becomes whether the classifier should classify the instance or reject it.\\
Furthermore, \citet{kimExitNotExit2024} proposes a method based on Markov Decision Process (MDP) for the early-exit decision criterion. The key intuition is to use the sequentiality of the Early Exits decisions to model the exiting strategy as a Markov decision problem. A sequential inference process of each data sample can be represented as a single episode with a finite horizon. A Reinforcement Learning approach is then used to solve the problem by maximizing the accuracy as much as possible while minimizing computational cost. This solution allows the early-exit decisions to be dynamic instead of simply using a fixed early-exit criterion.\\
Very recently, \cite{bajpaiBEEMBoostingPerformance2025} proposes the previous exit responses as experts for an agreement that provides robustness in the exit confidence decision. If the previous exits were leaning toward a different class, the current expert is considered not confident, and the network continues the computations.

\paragraph{Loss weighting} 
\hypertarget{para:lossweighting}{}\label{para:lossweighting}
Loss weighting methods try to determine the correct contribution of each exit output to the overall loss, with the aim of improving the learning of the network. It consists in determining the $\alpha_i$ values in equation \eqref{eq:objective}.
\citet{wangDynExitDynamicEarlyExit2019} proposes DynExit, a dynamic early-exit strategy for ResNet \citep{heDeepResidualLearning2015}.  This strategy introduces a dynamic loss-weight adjustment mechanism contributing to the weighting of the final loss as a weighted sum of all the losses, each computed at the respective exit. Instead of using manually tuned weighting values, in DynExit the weights become a trainable parameter with a dynamic range, which enables the loss-weight of different exit points. For this reason, the correct loss combination can be learned with all the other parameters, thus weighting the overall contribution of single exits to the learning process and increasing the overall accuracy. Furthermore, some hardware-aware optimizations of the implementation of the softmax and the cross entropy loss are presented, which are adapted to achieve high speed with low hardware complexity.\\
Finally, \citet{liuSelfsupervisedEfficientSample2023a} proposes a Self-supervised Efficient Sample Weighting method for balancing the contribution of the loss at each of the exits by predicting weights for input samples based on their losses at each exit. Given a classification loss and exit confidence for each exit, an MLP model uses pseudo labels as a target to generate the sample weights to be used in weighting the final loss. 

\paragraph{Exit positioning}
\hypertarget{para:exitpositioning}{}\label{para:exitpositioning}
In this paragraph, we cover those techniques that focus on the determination of the optimal number and position of the exit points within an Early Exits network. The positioning of the exits is a straightforward problem where in most cases the default option is to have the exits equally distributed within the depth of the network. Nonetheless, some studies have researched the optimal position of these exit points.
\citet{laskaridisHAPIHardwareAwareProgressive2020} presents HAPI, a methodology for creating an optimal Early Exits network design by optimizing the placement of the intermediate exits as well as the number of exits within a network. The Early Exits variants of the network are represented in a graph-matrix form that allows the design to be solved as a mathematical optimization problem. The method changes the number and position of Early Exits along the network by adding and removing nodes in the network graph. \\
Recently, \citet{liuMultipleExitTuningInferenceEfficient2024} proposes a solution to add Early Exits to ViT models, called MET. It introduces mainly two novelties: first, in between Transformer blocks, a devised adapter extracts appropriate representations for different exits in a shared space. To reduce the number of trainable parameters, different adapters in the network share a part of the parameters. Furthermore,  supervised graphs, which characterize the intra-class and inter-class relationships among samples, are adopted to regularize the representation-learning processes of Early Exits.
Furthermore, \citet{liPredictiveExitPrediction2023a} builds on top of the work of \citet{passalisEfficientAdaptiveInference2020} by allowing a fine-grained choice on the exit points; instead of fixing a determined amount of exit points, a learned component decides where to place optimally the position of classifiers for Early Exiting with respect to any block in the network. This approach takes also into account the trade-off between model accuracy and energy cost through the use of a hyperparameter. If the classification determines that the network needs more computation to get to a certain result, the next exit point position will be predicted. Finally, \citet{linCloserLookBranch2024} also further analyzes the effect of different design choices on the positioning of the branches.

\paragraph{Learning methods}
\hypertarget{para:learningmethods}{}\label{para:learningmethods}
Here we group all those methods that facilitate the learning of the objective of the Neural Network, taking also into account the Early Exits design of the architecture.
\citet{liImprovedTechniquesTraining2019} introduces two improvements to the training methods of Early Exits models. The first is an algorithm that tackles the problem of learning from different classifiers: the idea is to rescale the magnitude of gradients along the backpropagation path, allowing them to have a constant scale across the network for better stability. The second contribution is a knowledge transfer in combination with a self-distillation technique: a knowledge transfer path is applied among two adjacent classifiers, which can be considered as a residual connection of the classification. Concerning the self-distillation part, all the intermediate classifiers are also supervised using the output of the last classifier. Other techniques of self-distillation will be discussed later in the eponymous paragraph.\\
In \citet{bakhtiarniaImprovingAccuracyEarly2021}, the authors utilize curriculum learning to train multi-exit architectures, a training strategy for Neural Networks that imitates human learning by sorting the training samples based on their difficulty and subsequently gradually introducing more difficult examples to the network. This method uses the transfer teacher technique, where the losses of two different teacher networks are used to produce the scoring function to determine how hard an image sample is. The results presented show ideas for both the curriculum (easier to harder samples) and the anti-curriculum (harder to easier samples) training methods. \\
\citet{sunMetaGFTrainingDynamicDepth2022a} discusses in depth the problem of gradient conflict when it comes to early-exit networks: in the backpropagation process the gradients that have a negative cosine similarity value can harm the learning process by having conflicting update directions. The paper proposes Meta-GF, a weighted fusion policy to combine the gradients of each exit, where the fusion weights of gradients are also part of the learning process with all the other parameters. \\
Finally, the Zero Time Waste (ZTW) \citep{wolczykZeroTimeWaste2021, wojcikZeroTimeWaste2023a} approach proposes to reuse the output of internal classifiers as inductive prediction information to the later classification blocks of the model, for the cases where the exiting has been contemplated by performing a classification, but not performed. This is done by adding direct connections between classifiers and by combining previous outputs in an ensemble-like manner. Part of the work also demonstrates how the method performance can be improved with knowledge distillation between internal classifiers. 

\paragraph{Train-test mismatch} 
\hypertarget{para:traintestmismatch}{}\label{para:traintestmismatch}
The train-test mismatch is a problem where all the prediction heads are optimized on all types of data in the training phase, while during inference the early exiting of data will make the deeper layers of a network only see \textit{difficult} inputs.
The solution is to allow classifiers to focus more on \textit{hard} samples to better learn how to specialize only to those cases. \citet{hanLearningWeightSamples2022} proposes a method that allows to bridge this gap by imposing sample-wise weights on the loss of multiple exits. The key idea of the method is that the optimization objective should encourage each exit to emphasize different training samples by weighting their loss. To learn the appropriate weights, it leverages a weight prediction network (WPN). The WPN takes the training loss from all exits as input, producing the weights imposed on the samples at every exit. The backbone model and the WPN are optimized in a meta-learning manner. The meta objective for each classifier is then defined as the loss only on the samples that exit at a given classifier. \citet{yuBoostedDynamicNeural2022} proposes an alternative solution to the train-test mismatch problem. The solution starts from the assumption that a model is a weighted sum of weak prediction models and uses gradient boosting algorithms to minimize the loss function by iteratively training each weak model with gradient descent. In each training step, the model that points to the negative gradient of the loss function is selected as a new weak model, allowing each weak model to be considered supplementary to the ensemble of its previous weak models. \\
More recently, also \citet{regolJointlyLearnedExitInference2024} arrives at the same solution with a slightly different technique. The authors present a learning method for the gating and inference mechanisms. The paper proposes to optimize a loss that jointly assesses accuracy and inference cost by formulating the minimization as a bilevel optimization task. Each level of the bilevel optimization is simpler than the overall problem, leading to more stable learning.

\paragraph{Self distillation}
\hypertarget{para:selfdistillation}{}\label{para:selfdistillation}
Instead of the traditional two-model distillation technique, where a teacher model drives the learning process of a student model, in one-step self-distillation, the idea is to add intermediate classifiers within the same network, in an architecture similar to \citet{teerapittayanonBranchyNetFastInference2016a}, and treat those intermediate classifiers (and their network blocks) as student models with respect to deeper sections of the network itself. In \citet{zhangBeYourOwn2019} the training concept of self-distillation is first introduced. Specifically, the paper proposes a combination of three different training losses for the classifiers: a cross-entropy loss on the categorical classification, a Kullback-Leibler divergence on the softmax output, and an L2 loss applied from the feature map. In all three cases, the last classifier of the network is used as a teacher reference. \citet{phuongDistillationBasedTrainingMultiExit2019} arrives independently to the same self-distilling idea as the previous mentioned paper, but with a different temperature-scaled cross-entropy loss between output distributions.\\
\citet{wangHarmonizedDenseKnowledge2021} proposes the Harmonized Dense Knowledge Distillation (HDKD) training method for multi-exit architectures. The principle is that each exit learns from all its later exits through dense knowledge distillation to improve classification accuracy. Furthermore, a harmonized weight learning method is employed to obtain the best trade-off between the multiple objectives of multi-exit classification and dense distillation.\\
\citet{zhangSelfDistillationEfficientCompact2022} experiments with various self-distillation techniques: the Best Teacher Distillation uses the last output as a teacher and all Early Exits as students, the Ensemble Teacher Distillation uses an ensemble of all the classifier outputs as a reference label, in the Transitive Teacher Distillation every exit uses the next exit output as reference label and finally in the Dense Teacher Distillation, given an exit point, the knowledge from all the future classifiers is used. Furthermore, a genetic algorithm is applied to find the best threshold for shallow classifiers, encoding them into binary codes as the genes. The accuracy and acceleration results are utilized to measure the fitness of the threshold.\\
\citet{xuLGViTDynamicEarly2023a} presents a strategy to use self-distillation to train ViT models, to support Early Exits. The paper debates that  the ViT architecture is not ideal by design for Early Exiting strategies due to insufficient feature representation in shallower layers. For this reason, the paper explores the idea of self-distillation to guide intermediate classifiers to behave as much as possible like the final classifier. \\
Finally, \citet{zhangMultilevelCollaborativeSelfdistillation2024a} also proposes to train through self-distillation a multi-exit network. The key contributions of this paper are a dynamic generation of the importance weights for each classifier, a logit-based self-distillation module, and a feature-based self-distillation module.

\paragraph{Bag-of-Features, quantization, OOD, and NAS} 
\hypertarget{para:bof}{}\label{para:bof}
\emph{Bag-of-features} is a representation approach that aims to describe an image as a collection of unordered local features, without considering their spatial relationships. \citet{passalisAdaptiveInferenceUsing2019} and \citet{passalisEfficientAdaptiveInference2020} propose the use of Bag-of-Features to tackle the problem of information loss from previous layers' classifiers and the inability to handle large feature representations efficiently. The architecture integrates a feature extractor module and subsequently, the output is transformed into a histogram representation. These histograms are learned in a common space, allowing for a common classification layer, which is shared among all the Early Exits, further reducing the number of parameters needed for classification. \citet{passalisAdaptiveInferenceFace2021a} applies the method to the task of face recognition. \\
\emph{Quantization} is the process of reducing the precision of weights and activations from higher bit-widths to lower ones, generally to reduce the model size and improve inference speed. \citet{regolPredictingProbabilitiesError2024} proposes QuEE, a solution that combines quantization and Early Exits dynamic network. Rather than gating between exiting or continuing the computations, the decision verges among continuing with full or reduced computation. The proposed architecture allows to control both the number of blocks that are evaluated, through the use of Early Exits, and the amount of computation conducted within each block. This is conducted by using a clustering algorithm over a discretized space of possibilities.\\
\emph{Out-of-distribution (OOD)} detection refers to a model's ability to identify inputs that significantly differ from the distribution of the data the model was trained on. \citet{xiaWindowBasedEarlyExitCascades2023} explores tasks related to uncertainty estimation when it comes to Early Exits models. The proposed idea is to create an exit policy that instead of having a single threshold, uses a window-based approach. The intuition is that many downstream tasks for uncertainty estimation can be formulated as binary classification, thus only samples in a window near the binary decision boundary should be evaluated by later stages of a network. All the other samples outside of this window should be considered as OOD samples which can not be predicted and the network can exit earlier in order to avoid unnecessary computational cost. Also \citet{meronenFixingOverconfidenceDynamic2024a} proposes a Bayesian-based method to quantify and account for samples that are overconfidently wrongly estimated, trying to distinguish between samples where the model is uncertain and samples where the model is confidently wrong through probabilistic modeling. The motivation for improving uncertainty estimation is to allow intermediate classifiers to more accurately estimate the uncertainty in their prediction and thus better distinguish the cases where the overconfidence of the model results in an incorrect prediction in the early layers.\\
\emph{Neural Architecture Search (NAS)}, finally, is a field of research that aims to find the optimal subnetwork, often referred to as a ``winning ticket" \citep{frankleLotteryTicketHypothesis2019}, within an already pre-trained network. Although not discussed in detail here because it is outside the scope of this survey, we want to mention \citet{biggsATHEENAToolflowHardware2023a}, \citet{gambellaEDANASAdaptiveNeural2023} and \citet{gambellaNACHOSNeuralArchitecture2024a} as the first exploratory works in this direction. \\

\begin{table}
    \caption{Summary of Early Exit techniques presented in this survey (1/3).}
    \renewcommand{\arraystretch}{0.88}
    \begin{adjustbox}{width=\textwidth}
    \begin{tabular}{@{\extracolsep\fill}rp{6cm}p{1cm}p{2cm}p{4cm}p{2cm}}
        \toprule
        \textbf{Paper} & \textbf{Main contribution} & \textbf{Arch.} & \textbf{Exit policy} & \textbf{Application} & \textbf{Section} \\
        \midrule
        \citet{pandaConditionalDeepLearning2016} & Preliminary Early Exit work & CNN & Confidence & Image classification & \ref{subsubsec:architecture} \\
        \citet{teerapittayanonBranchyNetFastInference2016a} & First end-to-end network & CNN & Entropy & Image classification & \ref{subsubsec:architecture} \\
        \citet{bolukbasiAdaptiveNeuralNetworks2017} & Combination with nets ensemble & CNN & Learned policy & Image classification & \ref{subsubsec:architecture} \\
        \citet{huangMultiScaleDenseNetworks2018} & Multi-scale architecture & CNN & Confidence + budget cost & Image classification & \ref{subsubsec:architecture} \\
        \citet{kayaShallowDeepNetworksUnderstanding2019} & Slight variation wrt BranchyNet & CNN & Confidence & Image classification & \ref{subsubsec:architecture} \\
        \citet{liImprovedTechniquesTraining2019} & Rescale gradient magnitude & CNN & Confidence & Image classification & \ref{subsubsec:method} \\
        \citet{passalisAdaptiveInferenceUsing2019} & Bag-of-features + single classifier & CNN & - & Image classification & \ref{subsubsec:method} \\
        \citet{phuongDistillationBasedTrainingMultiExit2019} & Introduce self-distillation & CNN & - & Image classification & \ref{subsubsec:method} \\
        \citet{wangDynExitDynamicEarlyExit2019} & Dynamic loss-weight modification & CNN & Learned policy & Image classification & \ref{subsubsec:method} \\
        \citet{wangSEESchedulingEarly2019} & Frame dropping according to budget & CNN & Custom optimiz. problem & Video analysis & \ref{subsubsec:applications} \\
        \citet{zhangBeYourOwn2019} & Improves self-distillation loss & CNN & - & Image classification & \ref{subsubsec:method} \\
        \citet{chenLearningStopLearning2020} & Exit problem seen as variational Bayes & CNN & Learned policy & Image classification, few shot classification, image denoising & \ref{subsubsec:method} \\
        \citet{fangFlexDNNInputAdaptiveOnDevice2020} & Applies Early Exits to video analysis & CNN & Entropy & Activity recognition, video understanding, traffic surveillance & \ref{subsubsec:applications} \\
        \citet{laskaridisSPINNSynergisticProgressive2020} & Split computations between on-device and cloud & CNN & Confidence & Image classification & \ref{subsubsec:applications} \\
        \citet{laskaridisHAPIHardwareAwareProgressive2020} & First to investigate optimal exit positioning & CNN & Confidence, entropy & Image classification & \ref{subsubsec:method} \\
        % \citet{liEdgeAIDemand2020} & Split computations between on-device and cloud & CNN & Entropy & Image classification & \ref{subsubsec:applications} \\
        \citet{passalisEfficientAdaptiveInference2020} & Bag-of-features + single classifier & CNN & - & Image classification & \ref{subsubsec:method} \\
        \citet{scardapaneDifferentiableBranchingDeep2020a} & Weighting method to estimate exit confidence & CNN & Learned policy & Image classification & \ref{subsubsec:method} \\
        \citet{xingEarlyExitNot2020} & Application on compressed image enhancement & CNN & quality confidence & Quality enhancement & \ref{subsubsec:applications} \\
        \citet{yangResolutionAdaptiveNetworks2020} & Processes images at a coarser scale first & CNN & Confidence & Image classification & \ref{subsubsec:architecture} \\
        \citet{bakhtiarniaImprovingAccuracyEarly2021} & Use of curriculum learning to train & CNN & Entropy & Image classification & \ref{subsubsec:method} \\
        \citet{ghodratiFrameExitConditionalEarly2021a} & Offline frame sampling strategy with Early Exits & CNN & Learned policy & Video analysis & \ref{subsubsec:applications} \\
        \citet{juDynamicEarlyExit2021} & Early exits for video analytics & CNN & Learned policy & Video analysis & \ref{subsubsec:applications} \\
        \citet{liuAnytimeDensePrediction2021} & Early Exits for semantic segmentation & CNN & Confidence & Image segmentation & \ref{subsubsec:applications} \\
        \citet{bakhtiarniaImprovingAccuracyEarly2021} & Use of curriculum learning to train & CNN & Entropy & Image classification & \ref{subsubsec:method} \\
        \citet{ghodratiFrameExitConditionalEarly2021a} & Offline frame sampling strategy with Early Exits & CNN & Learned policy & Video analysis & \ref{subsubsec:applications} \\
        \citet{juDynamicEarlyExit2021} & Early exits for video analytics & CNN & Learned policy & Video analysis & \ref{subsubsec:applications} \\
        \citet{liuAnytimeDensePrediction2021} & Early Exits for semantic segmentation & CNN & Confidence & Image segmentation & \ref{subsubsec:applications} \\
        \citet{passalisAdaptiveInferenceFace2021a} & Applies BoF Early Exits to face recognition & CNN & - & Face Recognition & \ref{subsubsec:applications} \\
        \citet{shaoBranchyGNNDeviceEdgeCoInference2021} & Early exits graph neural network for point cloud & GNN & - & point cloud classification & \ref{subsubsec:applications} \\
        \citet{tanEmpoweringAdaptiveEarlyExit2021} & Threshold determination as non-convex problem & CNN & Learned policy & Image classification & \ref{subsubsec:method} \\
        % \botrule
    \end{tabular}
    \end{adjustbox}
\end{table}

\addtocounter{table}{-1}

\begin{table}

    \caption{Summary of Early Exit techniques presented in this survey (2/3). }
    \renewcommand{\arraystretch}{0.88}
    \begin{adjustbox}{width=\textwidth}
    \begin{tabular}{@{\extracolsep\fill}rp{6cm}p{1cm}p{2cm}p{4cm}p{2cm}}
        \toprule
        \textbf{Paper} & \textbf{Main contribution} & \textbf{Arch.} & \textbf{Exit policy} & \textbf{Application} & \textbf{Section} \\
        \midrule
        \citet{wangHarmonizedDenseKnowledge2021} & Dense knowledge distillation for each exit from all the later exits & CNN & Confidence & Image classification & \ref{subsubsec:method} \\
        \citet{wangNotAllImages2021a} & Image elaboration at different scales with Early Exits & Transf. & Confidence & Image classification & \ref{subsubsec:architecture} \\
        \citet{wolczykZeroTimeWaste2021} & Reuse the output of internal classifiers as inductive prediction & CNN & Confidence & Image classification & \ref{subsubsec:method} \\
        \citet{bakhtiarniaSinglelayerVisionTransformers2022} & Early exits audiovisual crowd counting & Transf. & - & Image classification & \ref{subsubsec:applications} \\
        \citet{hanLearningWeightSamples2022} & The loss is provided by a weighting network & CNN & Confidence & Image classification & \ref{subsubsec:method} \\
        \citet{kourisMultiExitSemanticSegmentation2022a} & Framework for Early Exit semantic segmentation & CNN & Confidence & Image segmentation & \ref{subsubsec:applications} \\
        \citet{pomponiProbabilisticReIntepretationConfidence2022} & Train by weighting the prediction with a trained confidence score & CNN & Confidence & Image classification & \ref{subsubsec:method} \\
        \citet{sabetTemporalEarlyExits2022} & Temporal Early Exits for video object detection & CNN & Entropy & Object detection & \ref{subsubsec:applications} \\
        \citet{sunMetaGFTrainingDynamicDepth2022a} & Weighted policy to alleviate gradient conflict problems & CNN & Confidence & Image classification & \ref{subsubsec:method} \\
        \citet{yuBoostedDynamicNeural2022} & Proposes a solution to the train-test mismatch problem & CNN & Confidence & Image classification & \ref{subsubsec:method} \\
        \citet{zhangSelfDistillationEfficientCompact2022} & Experiments with various self-distillation techniques & CNN & Confidence & Image classification & \ref{subsubsec:method} \\
        \citet{hanDynamicPerceiverEfficient2023} & Decouples the features extraction and the early classification & CNN & Confidence & Image classification & \ref{subsubsec:architecture} \\
        \citet{liPredictiveExitPrediction2023a} & learned component to decide where to place the classifiers & CNN & - & Image classification & \ref{subsubsec:method} \\
        \citet{liSEENNTemporalSpiking2023a} & Early exit architecture applies to a Spiking Neural Network & Spiking NN & Confidence & Image classification & \ref{subsubsec:applications} \\
        \citet{liuSelfsupervisedEfficientSample2023a} & Balance the loss contribution by weight prediction & CNN & Confidence & Image classification & \ref{subsubsec:method} \\
        \citet{polinaFIANCEEFasterInference2023} & Early Exits applied to Generative Adversarial Networks & GAN & quality threshold & Image generation & \ref{subsubsec:applications} \\
        \citet{tangDynamicTokenPruning2023a} & Token Early Exit for semantic segmentation & Transf. & - & Image segmentation & \ref{subsubsec:architecture} \\
        \citet{xiaWindowBasedEarlyExitCascades2023} & Study of uncertainty estimation when it comes to Early Exit & CNN & window of confidence & Image classification & \ref{subsubsec:method} \\
        \citet{xuLGViTDynamicEarly2023a} & Self-distillation to train Early Exit ViT models & Transf. & confidence threshold & Image classification & \ref{subsubsec:method} \\
        \citet{xueAdaptiveComputationElastic2023} & ViT  which has also the ability to read and store tokens & Transf. & - & Image classification & \ref{subsubsec:architecture} \\
        \citet{bajpaiCAPEENImageCaptioning2024} & ViT with Early Exits for image captioning & Transf. & Learned policy & Image captioning & \ref{subsubsec:applications} \\
        % \botrule
    \end{tabular}
    \end{adjustbox}
    
\end{table}

\addtocounter{table}{-1}

\begin{table}
    \caption{Summary of Early Exit techniques presented in this survey (3/3). }
    \renewcommand{\arraystretch}{0.88}
    \begin{adjustbox}{width=\textwidth}
        \begin{tabular}{@{\extracolsep\fill}rp{6cm}p{1cm}p{2cm}p{4cm}p{2cm}}
            \toprule
            \textbf{Paper} & \textbf{Main contribution} & \textbf{Arch.} & \textbf{Exit policy} & \textbf{Application} & \textbf{Section} \\
            \midrule
            \citet{gormezClassBasedThresholding2024a} & Early Exits with threshold tailored to the class & CNN \& Transformer & class confidence & Image classification & \ref{subsubsec:applications} \\
            \citet{kimExitNotExit2024} & Method based on the Markov decision process  for the early-exit decision  & CNN & Learned policy & Image classification & \ref{subsubsec:method} \\
            \citet{liuMultipleExitTuningInferenceEfficient2024} & Introduce an adapter for representations in a shared exit space & Transf. & Confidence & Image classification & \ref{subsubsec:method} \\
            \citet{meronenFixingOverconfidenceDynamic2024a} & Bayesian method to highlight out of distribution samples & CNN & - & Image classification & \ref{subsubsec:method} \\
            \citet{regolPredictingProbabilitiesError2024} & Combines quantization and Early Exit dynamic network & Transf. & Confidence & Image classification & \ref{subsubsec:method} \\
            \citet{regolJointlyLearnedExitInference2024} & Presents a loss for both accuracy and inference cost & Transf. & Confidence & Image classification & \ref{subsubsec:method} \\
            \citet{valadeEEROEarlyExit2024a} & The exiting is formalized as a classification with a reject option & CNN & Confidence & Image classification & \ref{subsubsec:method} \\
            \citet{yangAdaDetAdaptiveObject2024} & Early Exits for object detection & CNN & Entropy & Object detection & \ref{subsubsec:applications} \\
            \citet{zhangMultilevelCollaborativeSelfdistillation2024a} & Self distillation with a dynamic generation of the importance weights & CNN & Confidence & Image classification & \ref{subsubsec:method} \\
            \citet{addadCHASEChannelWiseSpatial2025} & Self-attention to improve feature representation and aggregation & CNN & - & Image classification & \ref{subsubsec:architecture} \\
            \citet{bajpaiBEEMBoostingPerformance2025} & Reuse of previous exits responses for exit confidence & CNN & Confidence & Image captioning & \ref{subsubsec:method} \\
            \citet{kuhseYouOnlyLook2025} & Early exit into YOLO architecture & CNN & - & Object detection & \ref{subsubsec:applications} \\
            \citet{wangAdaptiveRectangularConvolution} &  Adjust kernel shape based on the size of different objects &  CNN & - & Image enhancement & \ref{subsubsec:method} \\
            
        % \botrule
        \end{tabular}
    \end{adjustbox}
\end{table}

%%%%%%%%%%%%%%%%%%%%%%%%%%%%%%%%%%%

\subsubsection{Applications}
\label{subsubsec:applications}
%images
Early Exits networks have seen application in numerous computer vision tasks.
In the context of object detection \citet{yangAdaDetAdaptiveObject2024} proposes AdaDets, an architecture that speeds up inference through the use of Early Exits. The model formulates the bounding box prediction as a random variable following a certain probability distribution. This allows one to estimate the reliability of both the \textit{recognition and localization of a detected object}. Furthermore, an entropy score is used for all the detections to estimate the certainty of the network toward the estimation, and make a decision on where to stop the inference. Recently, \cite{kuhseYouOnlyLook2025} discusses a variation of the YOLO architecture to perform object detection with the option of exiting at early stages, in combination with specifically dedicated optimization algorithms. \\
\citet{liuAnytimeDensePrediction2021} proposes an adaptation of the Early Exits framework to \textit{semantic segmentation} bringing two novel solutions to the network: a redesign of the exiting heads to support early segmentation exits and a confidence adaptivity method. Similarly, \citet{kourisMultiExitSemanticSegmentation2022a} proposes a framework for adapting \textit{semantic segmentation} networks to support Early Exiting. The paper introduces a two-stage scheme for training the network, starting with an end-to-end exit-aware pre-training of the backbone thanks to a novel exit dropout loss for the extraction of semantically strong features in the early layers of the network. In the second stage, given the frozen backbone, all candidate Early Exits are further trained through a distillation scheme. More recently, \citet{gormezClassBasedThresholding2024a} improved the \textit{semantic segmentation} on top of \citet{liuAnytimeDensePrediction2021} architecture by including a threshold that is tailored to the class instead of globally, with the idea that some classes are easier to predict than others in this context, an thus Early Exits can save computations for those classes. \\
\citet{bajpaiCAPEENImageCaptioning2024} proposes CapEEN, a Neural Network architecture based on ViT which performs an Early Exits strategy for the task of \textit{image captioning}. The method uses knowledge distillation to the early layers to improve the prediction of early classifiers. Furthermore, an online learning algorithm based on the Multi-Armed Bandits framework is used to learn the exit threshold of the latent representations.\\
\citet{bakhtiarniaSinglelayerVisionTransformers2022} introduces an architecture for Early Exiting based on the ViT, called Single-Layer Vision Transformer (SL-ViT), which has the purpose of performing \textit{crowd counting}. The architecture uses a CNN as the backbone for multiple modalities (images and waveform) and then introduces a transformer-based fusion of global and local scope to determine the Early Exits point. The proposed branch fuses the local receptive field, where the Early Exits branch is attached to the global receptive field of the Early Exit, to effectively aggregate information from the entire input. Furthermore, the paper uses the copycat strategy \citep{correia-silvaCopycatCNNAre2021} for fine-tuning the model. \\
\citet{polinaFIANCEEFasterInference2023} introduces FIANCEE, a method for Early Exits in the context of \textit{image generation} using Generative Adversarial Networks. In this case, the method uses an Early Exits strategy for image synthesis, dynamically routing the computational flow towards the needed exit in accordance to the pictures’ complexity, therefore reducing computational redundancy while maintaining consistent quality. A small database of features is also used in order to condition the image generation and a predictor component is trained on the outputs to indicate the best exit outputting an image of a given quality. \\
\citet{xingEarlyExitNot2020} proposes RBQE, a network for resource-efficient blind \textit{image quality enhancement} for compressed images through the use of Early Exits networks. The model progressively enhances the quality of a previously compressed image through a dynamic deep Neural Network. The architecture is composed of a pyramidal shape series of vanilla CNNs with different depths which are trained with images that are compressed at different compression rates as input. By knowing the compression rate beforehand, the training of the exit strategy can be guided jointly with the denoising task. \\
%videos
The work of \citet{wangSEESchedulingEarly2019} proposes a scheduling solution for Early Exits models, applied to \textit{real-time video analysis}. It takes into account the case where a limited time budget is available due to constrained computing resources. The problem is formulated as an optimal scheduling problem, where the objective is to maximize a general overall utility. The scheduler learns to decide which frames to process and which to discard, as well as how much computation budget needs to be assigned to a given frame given that there is an overall deadline to keep the computations near real-time. Trying to solve the same task, FlexDNN \citep{fangFlexDNNInputAdaptiveOnDevice2020} adopts an architecture search-based scheme for \textit{efficient on-device video analytics}. The proposed solution is able to find the optimal architecture for each Early Exits branch with a trade-off balance between the Early Exits rate and its computational overhead. Moreover, FlexDNN derives the Early Exits insertion plan to identify the optimal number and locations of Early Exits. Also, \citet{ghodratiFrameExitConditionalEarly2021a} proposes an offline simple sampling strategy to process a video in combination with an Early Exits strategy to enable \textit{efficient recognition}. \citet{juDynamicEarlyExit2021} proposes a Dynamic Early Exits method with the aim to find the scheduling of exit points with the best performance both in terms of confidence and low delay. The model targets the application of \textit{video analytics}. At each exit point, the method estimates whether processing the following exit points can improve the performance. The framework consists of three modules that perform the following: jointly save the current estimations of the optimal utility-latency trade-off for each exit point; the exit oracle observes the current estimation as the context at each exit point, and dynamically decides whether to exit or not; if the exit oracle decides to exit the inference, the exit assessment updates and trains the toolbox, and moves on to process the next frame. \\
\citet{sabetTemporalEarlyExits2022} proposes a temporal Early Exits model to reduce the computational complexity of \textit{video object detection}. A specialized module composed of an attention map encodes the semantic differences between consecutive frames. Then, the full computation throughout the network is performed only if the frame is identified as having a significant change with respect to previous frames. This proposed solution can be applied to either per-frame-based pipelines---to identify semantic variation in the early stages of the network and hence avoid deep CNN computations for unchanged video frames---or to feature-propagation-based pipelines---to identify unchanged frames among non-keyframes and hence avoid further computations for those frames.
% point cloud
\citet{shaoBranchyGNNDeviceEdgeCoInference2021} introduces Branchy-GNN, a graph Neural Network for \textit{point cloud} processing, which implements an Early Exits mechanism to save on-device computation and an encoding module that compresses the intermediate feature before transmission to the edge device.\\
%distributed computation
\citet{laskaridisSPINNSynergisticProgressive2020} introduces SPINN, a distributed inference system that spreads the computations between device and cloud with a progressive inference method to deliver refined inference across diverse computational settings. Also in this case, a scheduler is introduced to optimize the exit policy with the network parameters, as well as the network splitting position at run time. The Early Exits mechanism allows the interruption of either the local or remote inference when a confident prediction is reached. A fallback to locally available results is also introduced in order to guarantee the responsiveness of the system even in situations of unavailable or corrupted connectivity. Similarly, \citet{liEdgeAIOnDemand2020} proposes a framework to use Early Exits to automatically decide the correct partition between on-device and on-edge with the idea of coordinating all the available resources for real-time inference.\\
% Spiking neural networks
Finally, Spiking Early-Exit Neural Networks (SEENN) \citep{liSEENNTemporalSpiking2023a} is an Early Exits solution that operates in the context of a \textit{Spiking Neural Network} and allows an adaptively varying number of timesteps from which to sample during inference. The main idea is to adapt the number of timesteps the network processes according to the classification difficulty, resulting in an Early Exit in the time dimension. Two variants are presented, where the stopping policy is determined with either the use of a confidence score thresholding or a Reinforcement Learning-based exit policy.\\

\subsection{Computational Routing} \label{section:routing}
In this Section, we group all the publications that present architectures capable of adjusting their computational path based on the input. We start in Section \ref{subsubsec:dynamicarchitecture} with architectures that are capable of dynamically building their computation path according to needs. In principle, the selection of an ad hoc computation tailored to the input ensures that the network focuses only on the components necessary for processing that specific input.\\
Thereafter, we focus on those dynamic architectures for which different components of a network are dynamically activated or the whole network is composed on the fly---in both cases according to the input. Specifically, we split the discussion according to the different computing units of a CNN. We refer to \citet{osheaIntroductionConvolutionalNeural2015} for a detailed explanation of the role of these components within a Convolutional Network. \\
% the filter (also known as kernel) is a small matrix of learnable weights that, through the convolution operation, slides over the spatial dimension of the input and allows a weighted sum for each input spatial position; the channels, which are a set of features in the intermediate stages and are the output product of the convolutional operation of filters applied to the input; the layers, which are a stack of varying operations that usually include the convolution, a linear activation, a pooling operation, and/or fully connected transformations; finally, blocks are high-level organizational units that comprehend a sequence of layers, and usually encapsulate repeated patterns to improve modularity and reusability.
In Section \ref{subsubsec:mixtureofexperts}, we illustrate various solutions regarding Mixture-of-Experts usage for image processing. \citet{jacobsAdaptiveMixturesLocal1991} introduces the original formulation of Mixture-of-Experts models. The key idea is to insert within a network a collection of \textit{expert} subnetworks, which specialize in solving a subportion of the problem. This technique has been very popular recently with the advent of Large Language Models, where Mixture-of-Experts allows a network to scale the number of parameters while maintaining the computational cost constant. In this context, and differing from the original proposal, a gating module is introduced, which is in charge of selecting, among all available \textit{experts}, a subportion of the most relevant ones to solve the task.  The current use of MoE models differs from what has been presented in \citet{jacobsAdaptiveMixturesLocal1991}: instead of resulting in an ensemble of models, the most popular formulation use the experts as alternative layers activated according to the gating module.
Formally, we can define a Mixture-of-Experts (MoE) model as a combination of the output of $N$ experts $f_1, f_2, ..., f_N$ weighted by the gating function $g(x)$, producing the final output as $y=\sum_{i=1}^{N}g_i(x)f_i(x)$, where $g_i(x)$ represents the weight for the $i$-th expert and usually sums up to 1. Figure \ref{fig:pagewidthimage}(b) provides a visual illustration of this technique.\\
Finally, in Section \ref{subsubsec:routingapplications} we review some of the applications of these techniques to computer vision tasks, such as object detection, image segmentation, video recognition, and some practical solutions for the deployment to resource-constrained hardware.\\
 Although the techniques detailed here are applied to visual processing components, Computational Routing shares deep roots and significant overlap with other deep learning domains. Most notably, the Mixture-of-Experts (MoE) paradigm has recently become unavoidable in the development of Large Language Models (LLMs). This mechanism allows a network to massively scale its parameters and capacity while maintaining a constant computational cost. This universal approach to solving efficiency bottlenecks means that advancements in routing policies or gating stability in vision seamlessly translate across the broader machine learning landscapes.
Similar to the Early Exits domain, numerical results for the publications in this section are not reported here. Also the field of Dynamic Routing is characterized by significant variations in test architectures and benchmark datasets, which prevents a meaningful direct comparison of performance. As previously discussed, this heterogeneity renders a fair quantitative analysis infeasible without extensive re-implementation and underscores the critical need for a standardized experimental protocol. A broader analysis of these limitations, and future directions to overcome these, are provided in the Discussion (Section \ref{section:discussion}).
% Numerical results from the studies in this section are not reported here. As is the case with Early Exits publications, the variations in test architectures and benchmark datasets prevent meaningful direct comparison.
% Define the difference between layers, channels, filters and blocks

\subsubsection{Dynamic architectures} \label{subsubsec:dynamicarchitecture}
\paragraph{Dynamic computation path} 
\hypertarget{para:dyncomputationpath}{}\label{para:dyncomputationpath}
In this paragraph, we present all papers in which the computation path of the network is decided on the fly based on the input.
Among the first papers to introduce an input-based model routing solution is \citet{bolukbasiAdaptiveNeuralNetworks2017} who propose an in-depth ensemble of networks chained in an acyclic computation graph. The image is supplied to the first block composed of Alexnet \citep{krizhevskyImageNetClassificationDeep2012}, after which a decision component evaluates whether the image should be further processed by a ResNet block \citep{heDeepResidualLearning2015}, an Inception block \citep{szegedyGoingDeeperConvolutions2015}, both, or neither (and thus exiting earlier).\\
\citet{rosenbaumRoutingNetworksAdaptive2018} presents a self-organizing Neural Network for the task of Multi-modal Learning. The network is composed of a set of one or more function blocks, which can be constructed from any Neural Network. After the block has been executed, the output is passed back to the router recursively, terminating when a fixed recursion depth is reached. In this way, the routing network dynamically composes different function blocks for each input. A collaborative Multi-Agent Reinforcement Learning approach is used to jointly train the router and function blocks. The router can also decide to take a pass action and leave the state unchanged. A separate Reinforcement Learning agent is created for the resolution of each task. Similarly, \citet{liuDynamicDeepNeural2018} proposed Dynamic Deep Neural Networks (D2NN), a feed-forward Neural Network that allows selective execution with self-defined topology, where the control policy is learned using single-step Reinforcement Learning. In this case, the training of the subparts of the network and the control units is separate. Very similarly, \citet{veitConvolutionalNetworksAdaptive2018} introduces a CNN with adaptive inference graphs (ConvNetAIG) that adaptively define their network topology conditioned on the input image.\\
\citet{liuDynamicallyThrottleableNeural2022} presents an architecture that can self-regulate both its own performance target and the computing resources. The network is composed of plug-and-play modules comprised of Convolutional and Fully-connected layers governed by a controller, with the ability to decide which and how many modules to ``turn off" to obtain the best performance for a given level of utilization. These constraints are enforced by a custom loss. The advantage of using a separate controller module is the possibility of preserving full network performance even in cases where the controller is not present. In this case, the controller is learned as a contextual bandit problem \citep{langfordEpochGreedyAlgorithmMultiarmed2007}.\\
\citet{fengLearningGenerateContentAware2020} proposes the formulation of a routing method based on the mapping of input samples into a shared routing space among all encoded images. The tool explicitly models the relationship between the sample space and the latent routing space by regularizing the distribution of routing paths. The key idea is that samples should follow similar routing paths if they are of similar nature. Furthermore, a differentiable loss is devised to optimize the average computational cost of the model, with the possibility of tailored tuning.\\
\citet{liLearningDynamicRouting2020} introduces an architecture that decides adaptively the scale at which each image gets processed. A scale transform path is decided on the fly by a differentiable conditional gate for end-to-end optimization. For each mode, multi-path propagation and skip connections are also considered. Furthermore, the paper presents the proficiency of the model in the context of semantic segmentation. A similar approach is also adopted by SegBlocks \citep{verelstSegBlocksBlockBasedDynamic2023}, but the variation of resolution happens within the image. The presented method adjusts dynamically the processing resolution of image regions based on their complexity. The image is split into blocks, and a lightweight policy network processes a low-resolution version of the input to decide which portions of the image should be processed at a high resolution. The policy network is trained using Reinforcement Learning, with a reward taking into account both computational complexity and task accuracy. It is interesting to notice that high and low resolution regions are processed by the same Convolutional filters, which therefore perceive objects at different scales.\\

\paragraph{Dynamic components} 
\hypertarget{para:dyncomponents}{}\label{para:dyncomponents}
\label{paragraph:blocklayerchannel}
% block
This paragraph illustrates a subset of solutions where a gating mechanism allows the input (or a portion of the input) to skip blocks of operations, single layers, single channels, or the operations are constrained to specific spatial positions in the context of CNNs.\\ 

\textbf{Filters and channels}. \citet{linRuntimeNeuralPruning2017} performs dynamic channel pruning according to the input image and current feature maps adaptively. The model consists of a CNN and a decision network. The pruning is performed layer-by-layer and modeled as a Markov Decision Process using Reinforcement Learning for training. The agent judges the importance of each convolutional kernel and conducts channel-wise pruning. The Markov Chain is conditioned on the previous state consisting of the previous feature maps. Also \citet{gaoDynamicChannelPruning2019} proposes to accelerate the Convolution operation by dynamically skipping unimportant input and output channels. The contribution of each channel in the layer is predicted, and subsequently, channels are dynamically amplified or suppressed with a parametric function based on the output of the previous layer. In this case, the dynamic channel pruning component is learned within the network through Stochastic Gradient Descent.\\
GaterNet \citep{chenYouLookTwice2019} starts from a similar idea of a backbone network used in combination with a global gating module to generate binary gates for selectively activating filters in the backbone network based on each input. In this case, the novelty is based on a discretization technique called Improved SemHash \citep{kaiserDiscreteAutoencodersSequence2018} to enable differentiable training of input-dependent binary gates such that the backbone and the gating network can be trained jointly via backpropagation. A key difference with other solutions is that the decision on each gate (for each filter) is made based on a shared global view of the current input.\\
\citet{verelstDynamicConvolutionsExploiting2020} introduces a method to spatially execute convolutional filters on important image locations only, instead of uniformly applying the filters all over the image. The method consists of a model trained end-to-end without explicit spatial supervision, where the gating decisions are trained using the Gumbel-Softmax.\\
\citet{liDynamicDualGating2021} presents a method to reduce the model complexity at runtime called dual gating. For each Convolutional block, the dual gating module is tasked with highlighting the informative features in both the channel and spatial dimensions. The essential areas are estimated by a spatial gating module, while the channel gating module identifies the most contributing channels to the final predictions. This approach allows to avoid background computations.\\
More recently, \citet{wangAdaptiveRectangularConvolution} presents a module that treats the height and width of the convolutional kernel as learnable parameters, allowing the shape of
the kernel to adjust dynamically based on the size of different objects.\\

% layer
\textbf{Layers}. \citet{figurnovSpatiallyAdaptiveComputation2017} proposes an architecture based on ResNet that dynamically adjusts the number of executed layers for certain regions of the image. The solution learns a deterministic policy that stops computations at a given spatial position as soon as the features computed become ``good enough" for prediction. For each spatial position, the Convolutional Filters are applied as long as the computational cost does not pass a fixed ponder cost, after which the value of the previous spatial position is copied. This is equivalent, in a residual block, to maintaining only the residual pass while zeroing all other computations. \\
Dynamic Fractional Skipping (DFS) \citep{shenFractionalSkippingFinerGrained2020} introduces layer-wise adaptive quantization as an intermediate alternative in a spectrum where performing full computation and skipping a layer are the two extremes. For each input, DFS dynamically assigns a bitwidth to both weights and activations of each layer, instead of assigning full bitwidth and zero bitwidth.  Using a two-step training procedure, the method learns to dynamically assign different bitwidths to both weights and activations of different layers based on the complexity of the input. The learning objective is based on the optimal trade-off between the maximization of the prediction accuracy and the minimization of the computational cost.\\

\textbf{Blocks}. SkipNet \citep{wangSkipNetLearningDynamic2018} present a revision of the Residual Network \citep{heDeepResidualLearning2015} which includes a gating mechanism to selectively skip Convolutional blocks based on the activations of the previous layer. The block-skipping problem is formulated as sequential decision-making and solved through a hybrid learning algorithm combining both Supervised Learning and Reinforcement Learning. Each group of layers has a gating module, which performs the binary decision to skip or execute the subsequent layer. The gating modules are trained in two stages: first, a pre-trained version is obtained through the use of a softmax relaxation of the binary skipping decision; since the parameters are not directly optimized for discrete selection for inference, a fine-tuning stage is performed using REINFORCE \citep{williamsSimpleStatisticalGradientfollowing1992} to refine the policy without relaxation. EnergyNet \citep{wangEnergyNetEnergyEfficientDynamic2018} adopts a very similar approach with two major variations: the use of a dynamic routing policy based on a hybrid energy loss that captures both computational and data movement costs, and the use of a single Recurrent Neural Network with shared parameters for the skip decision of all the blocks.\\
BlockDrop \citep{wuBlockDropDynamicInference2018} also presents a variation of the idea based on ResNet, where the selection of which residual blocks to evaluate for a given input is made on the fly. In this case, a policy network is trained through Reinforcement Learning, with a dual reward based on minimizing the number of blocks while preserving recognition accuracy. The policy network is trained using Curriculum Learning and later the pre-trained ResNet is further jointly fine-tuned with the policy network.\\
More recently, with the advent of ViT, the concept of block skipping has also been studied in relation to this architecture. One example is \citet{mengAdaViTAdaptiveVision2022}, where a light-weight multi-head subnetwork is attached to the ViT network, combining different ideas from both the skimming and routing worlds. The introduced module has the task of learning inference strategies to decide which patches to keep, which self-attention heads to activate, and which Transformer blocks to skip for each image, all within the same training. Also, in this case, the Gumbel-Softmax trick is used to make the whole framework end-to-end trainable. The computational cost of the network can be controlled by an hyperparameter, corresponding to the percentage of computational cost with respect to the full model. The ablation study demonstrates that the method tends to allocate less and less computations gradually through the backbone network, indicating that most of the redundancy resides in the later stages of the ViT backbone. \\ 
Given the nature of ViTs and their flexibility in treating portions of the input image as independent tokens that interact with one another, the idea of performing block skipping at a token level (selection for each token instead of the whole image) has gained popularity, allowing more granularity based on the content of portions of the image. This is conceptually equivalent to deciding which tokens attend the Transformer block and which are skipping it, a technique which we discuss in detail in Section \ref{section:skimming}. \\

\subsubsection{Mixture of Experts} \label{subsubsec:mixtureofexperts}
As already discussed, Mixture-of-Experts (MoE) (cf. Figure \ref{fig:pagewidthimage}(b)) is a machine learning technique that introduces multiple specialized models within a network to solve complex tasks. %A particular gating module has the task of deciding which experts to consult for a given input. This allows to grow the number of (task-specific) parameters in a network while keeping the computation burden constant. 
\citet{ahmedNetworkExpertsLargeScale2016} is among the first papers to introduce the concept of MoE for CNNs. The presented network is composed of a common backbone (referred to as ``trunk" in the paper) and a number of different branches that are specialized in the classification of a subset of categories that are difficult to tell apart. The training of the solution is completely end-to-end. The model is trained in two stages, where in the first part the multi-classification problem is decomposed into K groups of labels, one for each specialty. Secondly, the trunk network is learned by a generalist network, and the expert heads are attached, with a softmax layer putting the results together.\\
\citet{odenaChangingModelBehavior2017} presents a very early proposal of a dynamic routing network composed of both MoE and Early Exits. The network allows a controller component to adaptively choose which expert to redirect to the next block, or it can decide to reduce computation and exit earlier the network. The controller is implemented as an Recurrent Neural Network by making the controller functions share weights and a hidden state, and it is trained using the REINFORCE algorithm \citep{williamsSimpleStatisticalGradientfollowing1992}. \citet{grossHardMixturesExperts2017} presents a CNN-based architecture in which all experts are first trained separately and then combined into a common decoder. Only one expert gets picked at a time according to a $k$-means clustering solution, which associates the choice of the expert classifier with a location in the feature space.\\
HydraNets \citep{shazeerHydraNetsSpecializedDynamic2018} introduces a network architecture that contains distinct experts specialized at computing features for visually similar classes. Only a small number of these components are then dynamically selected to evaluate for any one input image. HydraNet maintains accuracy by having a large capacity that is semantically specialized to aspects of the input domain. Both the gating component and the branches are trained in a supervised fashion using ground truth labels for image classification.\\ 
\citet{wangDeepMixtureExperts2020} presents an architecture based on MoE that composes the filter of a CNN on the fly from a set of (expert) channels. The input is fed into both a base Convolutional Network and a shallow embedding network. The embedding network outputs a latent mixture of weights, which is used by the convolutional network to select the expert channels to use for each specific layer. The paper also demonstrates that this gating method is equivalent to constructing a MoE for each output channel. A loss based on the computational cost is also used to encourage sparsity in the gating network. Also, \citet{huaChannelGatingNeural2019} presents a similar idea and introduces a revision of the MoE method to learn specialized convolutional kernels for each example. This increases the size and capacity of the network while maintaining efficient inference. The idea is to compute convolutional kernels as a linear combination of $n$ learned experts. In this way, the convolutional kernel is constructed on the fly based on the input needs.\\
In \citet{abbasBiasedMixturesExperts2020} a MoE is proposed, where some experts are favored over others by design through an inductive prior bias. This allows the architecture to meet different constraints based on an external hyperparameter. The paper presents two methods to perform the training such that input space encourages expert usage to maximize performance while meeting a specified bias. \citet{jiangLearningLayerSkippableInference2020} proposes a coupled MoE network with layer-skip capability that dynamically carries out coarse-to-fine object categorization. The network is composed of two homogeneous subnets that process input visual information asymmetrically: one branch is designed to predict global-level classes, while the other provides more fine classes. A layer-skipping mechanism allows to dynamically skip layers for computational cost reduction. The layer-skipping mechanism in this case is optimized with a rank-based loss function. \\
In \citet{hazimehDSelectkDifferentiableSelection2021a} the authors raise the problem of slow convergence in the case of sparse gates (like top-$k$), which results in performance issues when training with gradient-based methods. The authors then introduce a reformulation of the binary selection based on first-order methods (such as SGD). \citet{zhangRobustMixtureofExpertTraining2023a} presents a method to strengthen the MoE CNNs. The paper talks mostly about adversarial attacks and robustness, trying to understand what, among all the factors (network, router), is mostly responsible for the lack of robustness and proposing a method to train robust MoE CNNs.\\
% tranformers
\citet{riquelmeScalingVisionSparse2021a} is the first to introduce the use of MoE in the context of ViT models. The specialization in this case can be performed on a token level, where each token gets assigned to one (or more) experts, and experts can accept tokens up to a capacity level. Generally, in ViTs, the experts is applied to the MLP portion of the network. The paper also introduces a routing algorithm that repurposes model sparsity to skip the computation of some patches, reducing computations on uninformative image regions. \citet{liuRoutersVisionMixture2024} further studies different routing approaches of MoE in the context of ViTs. The comparison considers binary routers for hard assignment among experts, and soft routers for soft assignment between experts and weighted combinations of the expert output tokens. Further assignment divisions can be made, either matching experts to tokens or vice versa. The study arrives at the conclusion that the soft MoE outperforms the binary solution in terms of accuracy and thus is favorable. M3ViT \citep{liangM3ViTMixtureofExpertsVision2022} introduces the use of a MoE framework to accelerate multi-task ViTs. Sparse task-specific experts get activated during training, which disentangles the parameter spaces to avoid different tasks’ training conflicts. The solution proposes tokens to per-expert queues to enable expert-by-expert computation rather than token-by-token. At inference time, this allows for activating only the task-corresponding sparse expert. A computation reordering scheme is also introduces to switch between tasks for deployment on memory-constrained devices.\\
In \cite{jainMixtureNestedExperts2024}, the idea of a nested experts structure is presented, where the model learns to dynamically process less important tokens with a less detailed representation. A novel routing algorithm for the tokens' assignment to experts is also proposed. \cite{xieMoDEMixtureofExpertsModel2024} tries to overcome the limited amount of tokens that each expert sees by introducing a tailored mutual distillation among experts to allow each expert to get a better perception of the global task. \\
We refer the reader to \cite{hanViMoEEmpiricalStudy2024} for an up-to-date, in-depth study on the design of MoE in the context of ViTs.

\subsubsection{Applications} \label{subsubsec:routingapplications}
In this section, we go through publications that present the application of computational routing techniques to solve specific applicational problems.
\citet{parkLearningDynamicNetwork2021a} explore the use of dynamic routing networks in the context of \textit{object detection}. The proposed architecture is based on MSDNet \citep{huangMultiScaleDenseNetworks2018} and automatically generates a sample-adaptive routing path on the fly. Two additional metrics for a global budget constraint and local path similarity regularization are added. The first serves the purpose of inducing sparsity in the network, while the second aims to generate more diverse routing paths. \cite{liuAccurateEfficient3D2025} explores the use of MoE models to dynamically achieve low latency and high accuracy as needed in the context of \textit{3D object detection}. After multiscale preprocessing, the system dispatches the input to different experts based on object distance and visibility.\\
\citet{pavlitskayaUsingMixtureExpert2020} investigates a MoE architecture to achieve interpretability in the context of \textit{semantic segmentation} while retaining comparable result quality. The principle consists in being able to observe both the overall model output as well as individual expert outputs, where the agreement or disagreement between those partial solutions can be used for insights into the decision process.\\
In \citet{sunDynamicNetworkQuantization2021a}, a framework for \textit{efficient video recognition} is presented. The solution selects the optimal precision for each frame conditioned on the input. Through Gumbel-Softmax sampling, an end-to-end differentiable approach selects optimal precision, taking both accuracy and efficiency into account. The solution is composed of a lightweight policy network (composed of a feature extractor and a LSTM module) and a video recognition network. The method also allows to skip frames in addition to selecting the precision dynamically. Furthermore, knowledge distillation from the full-precision recognition network is performed to guide the training of lower precisions and to improve accuracy. \citet{yangDyFADetDynamicFeature2024} presents DyFADet, a module that can adapt both kernel weights and receptive fields at different timestamps in the task of \textit{temporal action detection}. The architectures dynamically adjust detector parameters when applied at different pyramid levels, which corresponds to detecting the actions with different time ranges. \cite{sunHotMoEExploringSparse2025} applies a MoE strategy to reduce the computational complexity of elaborating hyperspectral images in the task of \textit{object tracking}.\\
Furthermore, \citet{wangAdaptiveFocusEfficient2021} introduces an adaptive patch localization solution for \textit{video recognition}. By using a Reinforcement Learning approach, the patch localization problem is defined as a sequential decision task, and a CNN is used to quickly process the full video sequence. Then, the selected patches are used by a larger network to get the final prediction. The solution takes into account also the possibility of skipping some less valuable frames in the training policy. \citet{wangAdaFocusV2EndtoEnd2022} reformulates AdaFocus as one-stage algorithm by introducing a differentiable interpolation-based patch selection approach. Furthermore, in this case, a conditional-exit technique is included to perform temporal adaptive computation. \cite{wangAdaFocusV3UnifiedSpatialTemporal2022} improves on top of the second version of AdaFocus by activating the expensive high-capacity network only on some small but informative 3D video cubes. \cite{huangPSRegPriorguidedSparse2025} applies MoE to the \textit{point cloud registration} task, introducing a prior-guided module to increase feature distinctiveness.
\\
Finally, \citet{panGradMDMAdversarialAttack2023a} studies the effect of \textit{adversarial attacks} on Neural Networks. The method proposes an algorithm to adjust the direction and the magnitude of the gradients with the malicious purpose of activating more computational units of dynamic models during inference. \\

\begin{table}
    \caption{Summary of Dynamic Routing techniques presented in this survey (1/2). }
    \renewcommand{\arraystretch}{0.88}
    \begin{adjustbox}{width=\textwidth}
        \begin{tabular}{@{\extracolsep\fill}rp{7cm}p{2cm}p{3cm}l}
            \toprule
            \textbf{Paper} & \textbf{Summary} & \textbf{Archit.} & \textbf{Dynamic component} & \textbf{Section}\\
            \midrule
            \citet{ahmedNetworkExpertsLargeScale2016} & Introduces MoE architecture for CNN & CNN & MoE & \ref{subsubsec:mixtureofexperts} \\
            \citet{bolukbasiAdaptiveNeuralNetworks2017} & Ensemble of networks chained in an acyclic computation graph & CNN & Computational path & \ref{paragraph:blocklayerchannel} \\
            \citet{figurnovSpatiallyAdaptiveComputation2017} & Dynamically adjusts the number of  layers for certain regions of the image & CNN & Layer & \ref{paragraph:blocklayerchannel} \\
            \citet{grossHardMixturesExperts2017} & MoE of pretrained experts & CNN & MoE & \ref{subsubsec:mixtureofexperts} \\
            \citet{linRuntimeNeuralPruning2017} & Dynamic channel pruning  & CNN & Channel & \ref{paragraph:blocklayerchannel} \\
            \citet{odenaChangingModelBehavior2017} & Combination of MoE and Early Exits  & CNN & MoE & \ref{subsubsec:mixtureofexperts} \\
            \citet{liuDynamicDeepNeural2018} & Selective execution with self-defined topolog & MLP & Computational path & \ref{paragraph:blocklayerchannel} \\
            \citet{rosenbaumRoutingNetworksAdaptive2018} & Self-organizing network for multi-modal learning & CNN & Computational path & \ref{paragraph:blocklayerchannel} \\
            \citet{shazeerHydraNetsSpecializedDynamic2018} & MoE for features of visually similar classes & CNN & MoE & \ref{subsubsec:mixtureofexperts} \\
            \citet{veitConvolutionalNetworksAdaptive2018} & Network with adaptive inference graphs & CNN & Computational path & \ref{paragraph:blocklayerchannel} \\
            \citet{wangEnergyNetEnergyEfficientDynamic2018} & RNN for skip decision of CNN blocks & CNN & Block & \ref{paragraph:blocklayerchannel} \\
            \citet{wangSkipNetLearningDynamic2018} & Selectively skip Convolutional blocks based on the previous layer & CNN & Block & \ref{paragraph:blocklayerchannel} \\
            \citet{wuBlockDropDynamicInference2018} & RL policy network for residual blocks skipping & CNN & Block & \ref{paragraph:blocklayerchannel} \\
            \citet{chenYouLookTwice2019} & Global gating module for channel selection & CNN & Channel & \ref{paragraph:blocklayerchannel} \\
            \citet{gaoDynamicChannelPruning2019} & Skip negligible input and output channels & CNN & Channel & \ref{paragraph:blocklayerchannel} \\
            \citet{huaChannelGatingNeural2019} & Learn specialized convolutional kernels as combination of learned experts & CNN & MoE & \ref{subsubsec:mixtureofexperts} \\
            \citet{abbasBiasedMixturesExperts2020} & MoE with inductive prior bias to certain experts & CNN & MoE & \ref{subsubsec:mixtureofexperts} \\
            \citet{fengLearningGenerateContentAware2020} & Models the relationship between the sample space and the latent routing space & CNN & Computational path & \ref{paragraph:blocklayerchannel} \\
            \citet{jiangLearningLayerSkippableInference2020} & MoE combined with layer skipping & CNN & MoE & \ref{subsubsec:mixtureofexperts} \\
            \citet{liLearningDynamicRouting2020} & Adapts the scale at which each image gets processed & CNN & - & \ref{paragraph:blocklayerchannel} \\
            \citet{pavlitskayaUsingMixtureExpert2020} & MoE for interpretability of Semantic Segmentation & CNN & MoE & \ref{subsubsec:routingapplications} \\
            \citet{shenFractionalSkippingFinerGrained2020} & Layer-wise adaptive quantization and skip & CNN & Layer & \ref{paragraph:blocklayerchannel} \\
            \citet{verelstDynamicConvolutionsExploiting2020} & Spatially execute convolutional filters only on important image patches & CNN & Channel & \ref{paragraph:blocklayerchannel} \\
            \citet{wangDeepMixtureExperts2020} & MoE at the level of convolutional filters, composed on-the-fly & CNN & MoE & \ref{subsubsec:mixtureofexperts} \\
            \citet{wangDualDynamicInference2020} & Layer and channel skipping for IoT applications & CNN & Layer \& channel & \ref{subsubsec:routingapplications} \\
            \citet{collemanProcessorArchitectureOptimization2021} & Investigates hardware constraints in the context of DyNN & CNN & - & \ref{subsubsec:routingapplications} \\
            % \botrule
        \end{tabular}
    \end{adjustbox}
\end{table}

\addtocounter{table}{-1}

\begin{table}
    \caption{Summary of Dynamic Routing techniques presented in this survey (2/2). }
    \renewcommand{\arraystretch}{0.88}
    \begin{adjustbox}{width=\textwidth}
        \begin{tabular}{@{\extracolsep\fill}rp{7cm}p{2cm}p{3cm}l}
            \toprule
            \textbf{Paper} & \textbf{Summary} & \textbf{Archit.} & \textbf{Dynamic component} & \textbf{Section}\\
            \midrule
            \citet{hazimehDSelectkDifferentiableSelection2021a} & Address the problem of slow convergence in sparse gates & CNN & MoE & \ref{subsubsec:mixtureofexperts} \\
            \citet{liDynamicDualGating2021} & Highlight the informative features in both the channel and spatial dimensions & CNN & Channel & \ref{paragraph:blocklayerchannel} \\
            \citet{parkLearningDynamicNetwork2021a} & Adaptive routing path for Object Detection & CNN & Computational path & \ref{subsubsec:routingapplications} \\
            \citet{riquelmeScalingVisionSparse2021a} & MoE in the context of ViT & Transformer & MoE & \ref{subsubsec:mixtureofexperts} \\
            \citet{sunDynamicNetworkQuantization2021a} & Adaptive model precision and frame skipping for Video Recognition & CNN & - & \ref{subsubsec:routingapplications} \\
            \citet{wangAdaptiveFocusEfficient2021} & Adaptive patch location for Video Recognition & CNN & Patch location & \ref{subsubsec:routingapplications} \\
            \citet{liangM3ViTMixtureofExpertsVision2022} & Method to accelerate MoE for multi-task ViT & Transformer & MoE & \ref{subsubsec:mixtureofexperts} \\
            \citet{liuDynamicallyThrottleableNeural2022} & Self-regulate computations according to performance target and resources available & CNN & Computational path & \ref{paragraph:blocklayerchannel} \\
            \citet{mengAdaViTAdaptiveVision2022} & Block skipping in Transformer & Transformer & Block & \ref{paragraph:blocklayerchannel} \\
            \citet{wangAdaFocusV2EndtoEnd2022} & One stage adaptive patch location for Video Recognition & CNN & Patch location & \ref{subsubsec:routingapplications} \\
            \citet{wangAdaFocusV3UnifiedSpatialTemporal2022} & Adaptive patch location for Video Recognition with conditional exits & CNN & Patch location & \ref{subsubsec:routingapplications} \\
            \citet{gaoDPACSHardwareAccelerated2023a} & Spatial and channel pruning hardware accelerator & CNN & Channel & \ref{subsubsec:routingapplications} \\
            \citet{panGradMDMAdversarialAttack2023a} & Studies adversarial attacks on dynamic models & CNN & - & \ref{subsubsec:routingapplications} \\
            \citet{verelstSegBlocksBlockBasedDynamic2023} & Adjusts dynamically the processing resolution of image regions & CNN & Computational path & \ref{paragraph:blocklayerchannel} \\
            \citet{zhangRobustMixtureofExpertTraining2023a} & Discuss adversarial attacks and robustness in the context of MoE & CNN & MoE & \ref{subsubsec:mixtureofexperts} \\
            \citet{devotoAdaptiveLayerSelection2024} & Layer skipping in the fine-tuning of ViT & Transformer & Block & \ref{paragraph:blocklayerchannel} \\
            \citet{liuRoutersVisionMixture2024} & Studies different approaches of MoE in the context of ViT & Transformer & MoE & \ref{subsubsec:mixtureofexperts} \\
            \citet{yangDyFADetDynamicFeature2024} & Adapts kernel weights and receptive field for different frames in Temporal Action Detection & CNN & Channel & \ref{subsubsec:routingapplications} \\
            \citet{hanViMoEEmpiricalStudy2024} & Empirical study on MoE design & Transformer  & MoE & \ref{subsubsec:mixtureofexperts} \\
            \citet{jainMixtureNestedExperts2024} & The model learns to dynamically process less important tokens with a less detailed representation & Transformer & MoE & \ref{subsubsec:mixtureofexperts} \\
            \citet{xieMoDEMixtureofExpertsModel2024} & Distillation among experts to allow each expert to get a better perception  & Transformer & MoE & \ref{subsubsec:mixtureofexperts} \\
            \citet{huangPSRegPriorguidedSparse2025} & MoE applied to Point Cloud registration task & Transformer & MoE & \ref{subsubsec:routingapplications} \\
            \citet{liuAccurateEfficient3D2025} & Multiscale MoE for low latency and high accuracy in 3D object detection & Transformer & MoE  & \ref{subsubsec:routingapplications} \\
            \citet{sunHotMoEExploringSparse2025} & MoE strategy for reducing hyperspectral images computational complexity & Transformer & MoE & \ref{subsubsec:routingapplications} \\
            \citet{chenMultimodalMedicalDiagnosis}  &  \citet{chenMultimodalMedicalDiagnosis} MoE framework for Vision Language medical diagnosis  &  Transformer & MoE & \ref{subsubsec:routingapplications} \\ 
    
            % \botrule
        \end{tabular}
    \end{adjustbox}
\end{table}

\subsection{Token Skimming} \label{section:skimming}
In the context of sparsification of Transformer models for computation efficiency, skimming techniques consist of avoiding redundant computations by dynamically recognizing among all tokens the ones that are useless or redundant to the final task. In the case of ViTs, the assumption is that not all parts of an image contribute equally to solving the task, and many tokens contain irrelevant information. As an example, we can think of an image where most of the background is composed of a blue sky. Although the presence of a \textit{sky} token can be semantically meaningful for a better understanding of the image context, the high number of tokens representing this information may harm the computational efficiency without necessary providing a better solution to the problem. The biological intuition is related to how humans efficiently perceive images and extract regions of interest from them \citep{zhouEyeTrackingData2017}. By focusing only on the important information within a sequence, skimming helps to achieve faster inference speed and reduce noise within the model.\\
We use token skimming as an umbrella term to denote two distinct token reduction strategies: token dropping and token merging. More formally, given a token matrix \(X \in R^{N \times D}\), where \(N\) is the number of tokens and \(D\) is the dimension of each token, the skimming operation can be defined as a function \(f: R^{N \times D} \rightarrow R^{N' \times D}\), where \(N' \leq N\). The construction of the function \(f\) can be implemented in different ways:
\begin{itemize}
    \item In Section \ref{subsubsec:dropping} we discuss the major contributions in \textit{token dropping}, where \(f\) becomes a selection function on the subset of tokens that can either be dropped (hierarchical dropping) or considered for selection again in the next block (per-block dropping).
    \item In Section \ref{subsubsec:merging} we present many texts related to \textit{token merging}, where the function \(f\) is in charge of combining the information of multiple tokens into one.
\end{itemize}
Subsequently, in Section \ref{subsubsec:skimother} we present some solutions of Token Skimming techniques applied to other tasks rather than only image classification.\\
Among all presented methods, Token Skimming is the most reusable across domains. This is primarily due to the underlying Transformer architecture treating inputs as sets of separate tokens that interact with each other in the self-attention block, regardless of what those tokens represent. This allows the algorithms for Token Skimming to be agnostic to the data's origin. Exceptions exist where the skimming process relies explicitly on 2D spatial properties; for instance, methods that utilize spatial adjacency to determine propagation graphs or group neighboring pixel patches. These methods restrict their direct application to non-visual tasks without some structural adaptations.

\begin{table}[h!]
    \centering
    \caption{Performance Comparison of Token Skimming methods on Imagenet validation set. For each publication, only the DeiT small architecture has been considered, and the results are presented as reported. If the small version of the architecture was not present, it has not been reported. In the case where the models presented have different skimming ratios, only the model with the highest accuracy has been reported.}
    \label{tab:performance_comparison}
    \begin{adjustbox}{width=\textwidth}
    \begin{tabular}{rp{1.6cm}p{1.6cm}p{1.6cm}p{1.6cm}p{1.6cm}p{1.2cm}}
    \toprule
    \textbf{Method} &  \textbf{Accuracy (top-1)} & \textbf{GFLOPS} & \textbf{Image Throughput} & \textbf{Params (M)} & \textbf{Type} \\
    \midrule
    % DynamicViT-LV-S/0.5 & 82 & 3.7 & - & 26.9 & \\
    DynamicViT-LV-S/0.7 \citep{raoDynamicViTEfficientVision2021} & 83.0 & 4.6 & - & 26.9 & Drop\\
    S2ViTE-Small \citep{chenChasingSparsityVision2021} &  79.2 & - & - & 14.6 & Drop\\
    % AdaViT \citep{mengAdaViTAdaptiveVision2022}& 81.1 & 3.9 & - & - & Merge\\
    Evo-ViT \citep{xuEvoViTSlowFastToken2022} &  79.4 & - & - & - & Merge\\
    % TokenLearner S/32 \citep{ryooTokenLearnerWhatCan2022a}&  76.3 & 1.9 & - & - & bottleneck\\
    E-ViT-DeiT-S \citep{liangNotAllPatches2022}&  81.3 & - & 4385 & - & Merge\\
    ATS \citep{fayyazAdaptiveTokenSampling2022}& 79.7 & 2.9 & - & 22.1 & Drop\\
    % SiT \citep{zongSelfslimmedVisionTransformer2022} &  83.2 & - & 1892 & - & bottleneck\\
    SaiT \citep{liSaiTSparseVision2022} &  83.1 & 4.3 & - & 26.2 & Drop\\
    A-ViT \citep{yinAViTAdaptiveTokens2022} &  78.6 & 3.6 & - & 22 & Drop\\
    \citep{longAttentiveTokensIncorporating2023} &  79.6 & 3.0 & - & 22.1 & Merge\\
    ToMe \citep{bolyaTokenMergingYour2023} &  79.3 & - & 1564 & - & Merge\\
    dTPS \citep{weiJointTokenPruning2023} &  80.1 & 3.0 & - & 22.8 & Merge\\
    MSViT \citep{havtornMSViTDynamicMixedscale2023} &  79.4 & 4.1 & - & - & Token number\\
    ToMeCIS \citep{seolTokenMergingClass2023a}&  78.9 & 2.7 & 1950 & 22.1 & Merge\\
    NEPAM \citep{jiangNeighborPatchesMerging2023a} &  79.4 & 3.7 & 2452 & - & Merge\\
    ToFu \citep{kimTokenFusionBridging2024} &  79.6 & 2.7 & 1561 & - & Drop \& merge\\
    GTP \citep{xuGTPViTEfficientVision2024} &  79.5 & - & 1581 & 22.1 & Merge\\
    IdleViT \citep{xuNoTokenLeft2024a} &  79.9 & - & 3031 & 22.1 & Drop\\
    Soft Token Merging \cite{yuanEfficientTransformerAdaptation2024} & 79.3 & 2.3 & 4566 & 24.0 & Merge\\
    \bottomrule
    \end{tabular}
    \end{adjustbox}
\end{table}

\subsubsection{Token dropping}
% hierarchical
\label{subsubsec:dropping}

\paragraph{Hierarchical dropping} 
\hypertarget{para:hierarchical}{}\label{para:hierarchical}
In one of the early works on hierarchical dropping, \citet{raoDynamicViTEfficientVision2021} introduces DynamicViT, with the idea of pruning redundant tokens progressively and dynamically. The sparsification of the tokens happens at predetermined layers within the network. During the training phase, a customized binary decision mask is produced. The purpose of the mask is to separate the informative and uninformative tokens. A local account of the token importance is achieved by reprojecting the tokens into a smaller latent space, while a global account is obtained by average pooling over the local importance. The two importance tokens are then concatenated, processed by a MLP, and reprojected through a softmax to obtain the probability for each original token. The Gumbel-Softmax trick is used to overcome the non-differentiable problem of sampling from a discrete distribution, while an attention-masking strategy guarantees that the excluded tokens are kept and can again be influential in the following operation down the model during the training process. This serves the purpose of keeping the correct batch size among all images during training, while in the inference phase, the unimportant tokens can be dropped to speed up the process. %gumbel-softmax 
On the other hand, \citet{panIARED2InterpretabilityAwareRedundancy2021} proposes a different method to hierarchically skim redundant tokens. It optimizes the network in a Curriculum Learning manner, training a router component for each block of heads and subsequently fixing it and moving on to the next block. The optimization of the routers is done through the REINFORCE method \citep{williamsSimpleStatisticalGradientfollowing1992}, where the reward function takes into account both efficiency and accuracy, and subsequently fine-tunes the Transformer blocks with a gradient computed based on the cross-entropy loss. The paper highlights how these types of methods are useful to provide an interpretability of the model behavior given the input. The model has been tested not only for classification but also for segmentation and video action recognition, showing improvement in all tasks. %hierarchical, reinforcement learning
\citet{liSaiTSparseVision2022} proposes an adaptive pruning strategy that dynamically drops the number of tokens based on the weights of the attention block. In this case, a Token Importance Score (TIS) is calculated during training and inferred starting from the attention score at a given head. The tokens are then filtered based on a top-$k$ strategy considering a fixed number of tokens, or with a dynamic selection where a fixed threshold is set and only the tokens above the threshold pass. Knowledge distillation from a teacher model is also introduced to improve the ability of early layers to learn the TIS.\\ 
\citet{fayyazAdaptiveTokenSampling2022} proposes ATS, an approach that automatically selects an adequate number of tokens at each stage based on the attention weights of the classification token in the self-attention layer. Firstly, a significance score for the input tokens is assigned based on the attention weights. Subsequently, inverse transform sampling is applied over the scores to sample a subset of all the tokens. If a token is sampled more than once, only one instance of the token is kept. For this reason, it is likely to happen that the real number of $K'$ unique tokens is often lower than the picked hyperparameter $K$, which represents the maximum number of tokens in the subset.\\ %dropping
In A-ViT \citep{yinAViTAdaptiveTokens2022} the first component of each token is scaled and passed through a logistic sigmoid to provide a probability of dropping. Accumulative importance is used to halt tokens as inference progresses into a deeper layer. Furthermore, the paper introduces a distributional prior regularization value such that tokens are expected to exit at a target depth, of which the probability density function is a Gaussian around a N target layer. This allows more control of the exit strategy.\\ %hierarchical \\
Finally, \citet{chenChasingSparsityVision2021} proposes a combination of methods to induce token sparsity in ViTs. Firstly, it introduces SViTE, a modification of ViT with a module that dynamically extracts and trains sparse subnetworks as opposed to training the full model. With a fixed computational cost, the technique jointly optimizes model parameters and explores connectivity through the entire training process. A second version further improves by introducing sparsity through a customized first-order importance approximation. In a third variation, the model plugs in a token to the patch embeddings in the current input sample, which is used to select the most important top-$k$ patches through Gumbel-Softmax and straight-through tricks.\\ 

\paragraph{Per-block dropping} \hypertarget{para:perblock}{}\label{para:perblock}\citet{mershaDynamicTransformerNetworks2022} introduces the same concept of block skipping for Transformer architectures, but on a token level instead of the entire image.
This is possible due to the different nature of the Transformer architecture: an image is treated as a set of separate tokens that interact with each other only in the Self-Attention Block. Skipping blocks on a per-token case is conceptually equivalent to deciding which ones among all available tokens attend the Transformer block. This allows an oracle function to evaluate all the tokens before each Transformer block and decide which ones should be processed by the block and which ones should skip to the next block. In the presented paper, the generated score for each sample is stored and fed to the oracle of the successive layer, which allows the network to propagate these decisions up to the final layer, where the quality of the decision is estimated and summed to the final loss. Also, \citet{xuNoTokenLeft2024a} proposes a very similar method, where tokens are also selected to either attend the block or skip to the next block. In this case, the model uses the attention score towards the classification token to determine the amount of tokens that will pass to the block. Furthermore, it proposes an optimization problem based on the binary graph cut technique, where the optimization tries to find the minimum normalized cut of tokens. \\
A variation of this concept is proposed in \citet{leiConditionalAdaptersParameterefficient2023a}, where each block passes through an MLP layer, which outputs a score. All the generated scores are then passed through a softmax, and top-$k$ are selected to attend the Tranformer block. The paper tests the solution on a large variety of tasks from very different domains.\\

\subsubsection{Token merging}
\label{subsubsec:merging}
Evo-ViT \citep{xuEvoViTSlowFastToken2022} is among the first to propose the merging of the unnecessary tokens together: the routing scheme divides the tokens into two parts where the top-$k$ important tokens are selected and go through the attention block (informative tokens), while the least important tokens are summarized and updated by a representative token (placeholder token), which then is also passed through the Transformer block. Centered Kernel Alignment \citep{kornblithSimilarityNeuralNetwork2019} is used to measure the similarity of the intermediate token feature in each layer with respect to the classification token, starting from the assumption that the classification token can be used individually since it is strongly correlated with classification. The informative tokens and the placeholder tokens are determined by a global class attention that gets updated across various layers. \\ %merging
E-ViT \citep{liangNotAllPatches2022} performs skimming of the tokens in between the Attention and MLP layers by observing the average of the attention weights of multiple heads at specific blocks (3 blocks, spread in the network). First, the top-$k$ most important tokens are selected, and then all the remaining tokens are merged into one, taking into account a weighted average. The attention weights are used as an indicator of the contribution (and thus importance) of each token to the classification. The importance is estimated with respect to the classification token. \citet{longAttentiveTokensIncorporating2023} improves on top of E-ViT by merging tokens, considering both importance and global token diversity. The proposed technique first splits the tokens into two different sets according to the attention output: attentive tokens and inattentive tokens, as seen also in E-ViT. However, instead of merging all the inattentive tokens, it performs a density peak clustering algorithm for the merging of the inattentive tokens, while a custom cosine similarity-based algorithm is used to cluster the top-$k$ most similar tokens in the attentive group. \\%merging
On the other hand, \citet{bolyaTokenMergingYour2023} proposes ToMe, a method that gradually combines similar tokens in a Transformer using a general and lightweight matching algorithm. It combines tokens with a custom matching algorithm that uses the keys matrix $K$ in the attention block as a denoised summarization of the information present in each token, and cosine similarity to determine which of the keys contain similar information. The reduction of the number of tokens per block is fixed to a number $r$ regardless of the image content to avoid batch mismatch in the number of tokens per image. This method directly uses the keys matrix to produce the selection of the top-$k$ tokens without the need of a score embedded in the network. Thus, the method can be applied without the need for further training, with a little drop in performance, or it can be further fine-tuned for better results. \citet{bolyaTokenMergingFast2023} extends this work by applying ToMe to stable diffusion \citep{rombachHighResolutionImageSynthesis2022}. %merging
Additionally, \citet{seolTokenMergingClass2023a} proposes ToMeCIS, a method that also builds on top of ToMe. The merging technique is the same, but in this case, the values of the new tokens are obtained by a weighted average according to a computed importance score of the token.\\
\citet{weiJointTokenPruning2023} proposes a joint Token Pruning Squeezing (TPS) module. The idea is to have a scoring method based on Gumbel-Softmax to evaluate the tokens, which are then divided into two subsets. An alternative is also explored, where the class token attention values are used to measure the tokens' importance---as already seen in EViT. The subset with pruned tokens is then fused into the subset with tokens reserved. An unidirectional nearest-neighbor matching algorithm is used to identify the pruned tokens that are most similar to the reserved tokens. The pruned tokens are then fused with a weighting scheme based on the similarity of the reserved tokens. The similarity score used is the cosine similarity, while two solutions for placement of the TPS module are proposed: intrablocks (in between ViT blocks) and in-blocks (between MHSA and MLP). %merging
NEPAM \citep{jiangNeighborPatchesMerging2023a} performs token merging based on the similarity of pixel patches: tokens get divided into groups of neighbors, and the similarity of tokens within the group is then computed, using as a reference token the one in the upper-left corner of each group. The groups with the most similar tokens get merged. Furthermore, the method proposes a multi-scale relative position embedding based on the addition of a bias matrix to the tokens to solve the problem of corrupted position information of the merged embeddings. \\
The method proposed in \citet{xuGTPViTEfficientVision2024} tries to solve the challenge of balancing model efficiency and information preservation for efficient ViTs through the use of Graph-based Token Propagation (GTP). In the paper, the authors propose two different metrics to measure the token's importance: regeneration difficulty and broadcasting ability. The first one represents the ability of a token to be aggregated by other tokens during the self-attention process, while the second score represents the ability of a token to contribute to other tokens in the self-attention computations. The self-attention map is used to compute these scores, so the method can be applied without learning of other tokens or needing to fine-tune the solution. Based on the spatial position (adjacency) and semantic position (cosine similarity) of tokens, a propagation graph of each image is determined for each input, and the nodes with low scores get merged according to the graph direction. Furthermore, in the paper, it is observed that discarding tokens tends to yield a smoother attention map after Softmax activation. To resolve this issue, attention map sparsification is applied as an anti-oversmoothing mechanism.\\
\cite{yuanEfficientTransformerAdaptation2024} introduces a token merging system based on a self-attentive method. The introduced block creates a score matrix that compares each input token with each other, subsequently allowing the merging of tokens through a weighted sum. An inflation block is also introduced to reconstruct the tokens after the transformer block.

\begin{table}
    \caption{Summary of Token Skimming techniques presented in this survey (1/2).}
    \label{tab:contributions_2}
    \renewcommand{\arraystretch}{0.88}
    \begin{adjustbox}{width=\textwidth}
        \begin{tabular}{lp{6.5cm}ll}
        \toprule
        \textbf{Paper} & \textbf{Summary} & \textbf{Type} & \textbf{Section} \\
        \midrule
        \citet{chenChasingSparsityVision2021} & Combination of methods to induce token sparsity & Hierarchical dropping & \ref{subsubsec:dropping} \\
        \citet{panIARED2InterpretabilityAwareRedundancy2021} & Train per block in a curriculum-learning manner with RL & Hierarchical dropping & \ref{subsubsec:dropping} \\
        \citet{raoDynamicViTEfficientVision2021} & Pruning of redundant tokens progressively and dynamically & Hierarchical dropping & \ref{subsubsec:dropping} \\
        \citet{fayyazAdaptiveTokenSampling2022} & Selection based on attention weights of the classification token  & Hierarchical dropping & \ref{subsubsec:dropping} \\
        \citet{liangNotAllPatches2022} & Merges according to the attention weights of multiple heads at specific blocks & Merging & \ref{subsubsec:merging} \\
        \citet{liSaiTSparseVision2022} & Tokens selection based on the weights of the attention block & Hierarchical dropping & \ref{subsubsec:dropping} \\
        \citet{mershaDynamicTransformerNetworks2022} & At each block, an function evaluates which tokens should attend it & Per-block dropping & \ref{subsubsec:dropping} \\
        \citet{xuEvoViTSlowFastToken2022} & The least important tokens are summarized by a representative token & Merging & \ref{subsubsec:merging} \\
        \citet{yinAViTAdaptiveTokens2022} & Halt tokens by accumulative importance, with bias target exit depth. & Hierarchical dropping & \ref{subsubsec:dropping} \\
        \citet{bolyaTokenMergingYour2023} & Gradually combines similar tokens with a custom matching algorithm & Merging & \ref{subsubsec:merging} \\
        \citet{bolyaTokenMergingFast2023} & Applies \citet{bolyaTokenMergingYour2023} algorithm to diffusion models & Merging & \ref{subsubsec:merging} \\
        \citet{chenEfficientVideoAction2023a} & Token dropout for video task recognition & Hierarchical dropping & \ref{subsubsec:skimother} \\
        \citet{chenSparseViTRevisitingActivation2023} & Introduce activation sparsity for Swin-based models & Hierarchical dropping & \ref{subsubsec:skimother} \\
        \citet{havtornMSViTDynamicMixedscale2023} & Selects the optimal token scale for every image region & - & \ref{subsubsec:skimother} \\
        \citet{jiangNeighborPatchesMerging2023a} & Token merging based on the similarity of pixel patches & Merging & \ref{subsubsec:merging} \\
        \citet{leiConditionalAdaptersParameterefficient2023a} & Per block skip according to the output of inferred score & Per-block dropping & \ref{subsubsec:dropping} \\
        \citet{longAttentiveTokensIncorporating2023} & Merge tokens according to importance and global token diversity & Merging & \ref{subsubsec:merging} \\
        \citet{luContentawareTokenSharing2023a} & Token reduction for ViTs semantic segmentation & - & \ref{subsubsec:skimother} \\
        \citet{seolTokenMergingClass2023a} & Skimming based on weighted average according to a importance score of the token & Merging & \ref{subsubsec:merging} \\
        \citet{weiJointTokenPruning2023} & Scoring method based on Gumbel-Softmax for merging tokens & Merging & \ref{subsubsec:merging} \\
        \citet{devotoAdaptiveSemanticToken2024} & Token skimming for variable latency and bandwidth constraints in communication channels & Hierarchical dropping & \ref{subsubsec:skimother} \\
        \bottomrule
        \end{tabular}
    \end{adjustbox}
\end{table}

\addtocounter{table}{-1}

\begin{table}
    \caption{Summary of Token Skimming techniques presented in this survey(2/2).}
    \label{tab:contributions}
    \renewcommand{\arraystretch}{0.88}
    \begin{adjustbox}{width=\textwidth}
        \begin{tabular}{lp{6.5cm}ll}
        \toprule
        \textbf{Paper} & \textbf{Summary} & \textbf{Type} & \textbf{Section} \\
        \midrule
        \citet{kimTokenFusionBridging2024} & Aggregation based on both token pruning and token merging & Dropping \& merging & \ref{subsubsec:skimother} \\
        \citet{liuRevisitingTokenPruning2024a} & Token Skimming for object detection and semantic segmentation & Hierarchical dropping & \ref{subsubsec:skimother} \\
        \citet{pengSceneAdaptiveSparse2024} & Token dropout for Event cameras & Hierarchical dropping & \ref{subsubsec:skimother} \\
        \citet{xuATFTransAttentionweightedToken2024a} & Token merging for Object Tracking & Merging & \ref{subsubsec:skimother} \\
        \citet{xuGTPViTEfficientVision2024} & Uses a Graph-based Token Propagation (GTP) as a merging policy & Merging & \ref{subsubsec:merging} \\
        \citet{xuNoTokenLeft2024a} & Per block skip based on the attention score & Per-block dropping & \ref{subsubsec:dropping} \\
        \citet{yuanEfficientTransformerAdaptation2024} & Token merging system base on self-attentive method & Merging & \ref{subsubsec:merging} \\
        \citet{chengNotAllTokens2025} & Controls the token sparsity through the use of the importance distribution for object detection & Merging & \ref{subsubsec:skimother} \\
        \citet{fangAttendNotAttended2025} & Merge redundant features in the context of diffusion models & Merging & \ref{subsubsec:skimother} \\
        \citet{luToMATokenMerge2025} & Token merging for diffusion transformers with GPU friendly operations & Merging & \ref{subsubsec:skimother} \\
        \bottomrule
        \end{tabular}
    \end{adjustbox}
\end{table}

\subsubsection{Other methods and applications}
\label{subsubsec:skimother}

There are numerous other token skimming approaches that do not fall within the token dropping or token merging categories. 
For example, \citet{chenSparseViTRevisitingActivation2023} is a model that revisits the activation sparsity for Swin-based models \citep{liuSwinTransformerHierarchical2021}. It introduces a sparsity-aware adaptation and applies an evolutionary search to efficiently find the optimal sparsity for each layer configuration within the vast search space. The importance of each window is given by its L2 activation magnitude. The windows with the highest score are gathered, then passed through the transformer block, and finally, the output is scattered back with the unprocessed windows. Different sparsity levels are applied at different layers due to the principle that early layers need to see more tokens than the later layers. In order to find the optimal sparsity for each layer, the network is first trained to perform at any random sparsity level, then an evolution algorithm is applied to find the optimal sparsity for each layer, and finally, the model is fine-tuned with the identified optimal sparsity configuration.\\ 
MSViT \citep{havtornMSViTDynamicMixedscale2023} presents a unified single-stage model that applies a conditional gating mechanism to select the optimal token scale for every image region, such that the number of tokens is dynamically determined per input. The tokenization scale for each image region is predetermined in a preprocessing step before the Transformer model, and it is based on a lightweight MLP. In the presented solution, a binary decision is taken for each region to choose whether this has to be processed at either the coarse or fine scale. To avoid the problem of a dynamic number of tokens per batch per image, in the training phase only the top-$k$ tokens are kept, where $k$ has a lower bound that corresponds to the number of tokens at fine scale. \\ 
ToFU \citep{kimTokenFusionBridging2024} proposes an aggregation based on both token pruning and token merging according to the model's `functional linearity' (or the degree to which a model's output for interpolated inputs aligns with a linear behavior). The argument is that the function that describes the merging of two points may not necessarily be linear, and thus it requires some correction weights. The Maximum Norm Linear Interpolation merging technique is introduced---a method constructed to merge multiple tokens while conserving the norm distribution. It computes the normalized average of the features, and then scales this normalized average using the maximum norm of the individual vectors.\\
While all the methods seen above solve the task of \textit{image classification}, \citet{luContentawareTokenSharing2023a} focuses on \textit{semantic segmentation}. It is based on token reduction in the context of ViT networks. The paper proposes a class-agnostic policy network that predicts whether image patches contain the same semantic class, and lets them share a token if they do. The grouping is done on rectangular neighboring regions to ease the reassembling of tokens at the output of the ViT backbone, necessary to perform semantic segmentation on each reconstructed token.\\
\citet{chenEfficientVideoAction2023a} proposes a spatiotemporal token dropout to improve the efficiency of \textit{video task recognition}, preserving keyframes in a video for context and the important motion parts for all other frames. Two pruning stages are performed sequentially: first, for all the non-keyframe tokens, the importance score is calculated with the relative keyframe, and only the top-$k$ tokens are kept. Subsequently, an approach similar to E-ViT \citep{liangNotAllPatches2022} is used to further skim irrelevant tokens to the prediction. \\
More recently, \citet{liuRevisitingTokenPruning2024a} applies the idea of per-block Token Dropping to the tasks of \textit{object detection} and \textit{semantic segmentation}. In this case, the selection is performed through sampling from a distribution noised with the Gumbel-Softmax trick to make the decision differentiable. \citet{chengNotAllTokens2025} controls the token sparsity through the use of the importance distribution. This allows the solution to preserve both spatial details and the global receptive field when used with a DETR head \cite{carionEndtoEndObjectDetection2020a}. \citet{pengSceneAdaptiveSparse2024} introduces an hierarchical dropout solution for \textit{object detection} in scenes recorded by event cameras.\\
\citet{xuATFTransAttentionweightedToken2024a} proposes a token fusion method for the task of \textit{object tracking}. The input of the encoder is composed of search tokens and exemplar tokens, where the search token represents the object that needs to be tracked and the exemplar token can contain either tracking object tokens, which are useful, or background tokens, which get fused together in order to reduce the overall number of tokens. The selection of the tokens is based on the attention weights. \\
\citet{fangAttendNotAttended2025} investigates how to prune feature redundancies in the context of diffusion models, introducing a structure-then-detail token merging approach. \citet{luToMATokenMerge2025} focuses on token merging for diffusion transformers, taking into account only GPU-friendly operations.\\
Last, \citet{devotoAdaptiveSemanticToken2024} considers the case of a deep learning model where, in between the encoder and the decoder components, there is a channel layer for communication that happens with variable \textit{latency and bandwidth constraints}. In this case, the design of a Token Skimming mechanism helps reduce the number of tokens to an amount that can be transmitted within the communication layer constraints. The filtering rate is a hyperparameter that can be set by the user. \\

\section{Dynamic Sensor Fusion in the Context of Computer Vision} \label{section:sensorfusion}

In this section, we provide an overview of methods that use any of the Dynamic Neural Network techniques in the context of Sensor Fusion. We consider almost exclusively publications where Sensor Fusion is performed for Vision purposes, with the exception of two papers in paragraph ``Other Sensor Fusion approaches", that we still consider relevant enough to be included.\\
Analyzing the number of publications, there is clearly a research gap when compared to other applications of Adaptive Networks. On the other hand, we can argue that Dynamic Neural Networks are very beneficial in the context of Sensor Fusion for several reasons. First, they serve efficiency purposes, and usually, Sensor Fusion applications run on low-computing devices or target real-time applications. Introducing some sort of elasticity in elaborating the different inputs allows for reduced computations. It enables to remove redundant and uninformative inputs and relates the use of certain sensors only to cases where the conditions benefit the overall prediction. In this sense, the benefit is derived from a more controlled (and tunable) trade-off between achieving a desired accuracy and constraining the computation resources. Furthermore, dynamically excluding certain sensors can help reduce the noise generated by the said sensors, in cases where the working conditions are suboptimal or a sensor failure is present. In these cases, a dynamic model can perform an improved selection of the optimal subset of sensors, taking into account also the environmental conditions \citep{chenSelectiveSensorFusion2019, malawadeEcoFusionEnergyawareAdaptive2022}. Finally, these methods are able to identify and prioritize the elaboration of critical information by design. This is particularly useful in contexts where computation is a limited resource and the system needs to be responsive. We can imagine a setting where information can be elaborated either on-device or on the cloud. The introduced dynamicity in this case can help to better differentiate between information that is critical and needs to be computed on-premises for real-time and all-time responsiveness, and the information that can be sent out of the device to improve the overall model accuracy at the cost of latency or possibly information loss.\\

 When it comes to Dynamic Neural Networks for Sensor Fusion the field remains relatively underexplored compared to unimodal Computer Vision. Furthermore, the deployment of dynamic multi-modal architectures introduces specific failure modes, operational constraints, and evaluation bottlenecks that require further investigation. The primary vulnerability of dynamic fusion architectures lies in the gating or routing mechanisms. Gating modules are usually trained to assess sensor reliability based on the environmental context. In cases of Out-of-Distribution (OOD) conditions or extreme domain shifts, the gating network can become miscalibrated. This leads to a failure in identifying corrupted modalities, routing noisy data into the computation path. Furthermore, the dynamic router can disproportionately favor a high-bandwidth sensor (e.g., RGB camera) during the training phase. This leads to modality dominance, effectively reducing the dynamic fusion to a static unimodal inference path and neutralizing the benefits of the multi-sensor setup. Lastly, Sensor Fusion involves dealing with sensor synchronization. The architecture has to process asynchronous data streams operating at heterogeneous frequencies (e.g., cameras at 30Hz, LiDAR at 10Hz, IMU at 200Hz). While dynamic networks theoretically offer the ability to adaptively handle variable-rate streams, synchronizing these inputs dynamically on the fly introduces significant computational overhead. In latency-critical applications such as autonomous navigation, the temporal cost of evaluating the gating function to select the optimal sensor subset is not completely clear and could potentially negate the latency saved by skipping computations. Furthermore, in distributed systems where computation is offloaded to edge units, fluctuating communication bandwidth and transmission latency need to be taken into account on top of theoretical FLOP reduction.\\
Finally, it is clear that the field is missing proper evaluation tests to correctly identify the robustness in case of missing or degraded modalities. It is important to highlight the behavior of the model under varying degrees of sensor noise, inducing random modality dropouts, and testing under artificial latency constraints. These types of measurements are fundamental for estimating the network's ability to operate under safety requirements in case of individual or multiple sensor failures. Generally speaking, the transition toward practical applicability and rigorous real-world testing remains a missing step in the current research.\\
In the following paragraphs, we present the application of Adaptive Sensor Fusion techniques for object detection, semantic segmentation, odometry, steering prediction and finally we introduce three different use cases of Sensor Fusion in the context of stress detection, power grid attack detection and multi-modal medical diagnosis.

\paragraph{Object Detection}{ 
\hypertarget{para:objectdetection}{}\label{para:objectdetection}
\citet{meesChoosingSmartlyAdaptive2016b} is the first to propose a novel adaptive fusion approach for object detection. The principle is to learn the weighting of predictions of different sensor modalities in an online manner with an architecture based on a CNN-based MoE. The model works with RGB, depth and motion information. The network performs late fusion: it uses different CNN encoders for each modality and then a gating network which provides both a selection and the weighting of the experts outputs in order to get to the final classification values.\\
\citet{malawadeHydraFusionContextAwareSelective2022} presents HydraFusion, a framework that applies dynamic late fusion (after Faster-RCNN sensor processing) in a context-aware manner to perform object detection. It selects a fixed amount of $k$ sensors out of $n$, and uses a novel attention-based module to determine which are the most important sensors. The execution time of the model is on par with any late fusion method due to the fact that all inputs are processed but only $k$ of them are selected. This selection brings improvements in accuracy due to the induced robustness of the architecture in cherry-picking only a subset of inputs. It is also noticeable that in the case where $k=1$ (only one sensor selected among all), the ablation study shows that a statistical method based on domain knowledge (based on the weather conditions) is sufficient while increasing $k$ (multiple sensor selected and fused) deep learning-based selection performs better. \\ % task: object detection, techniques: top-k selection, type of fusion: hybrid fusion (early and late)
Similarly, EcoFusion \citep{malawadeEcoFusionEnergyawareAdaptive2022} proposes an energy-aware Sensor Fusion approach which uses the environmental context to adapt the fusion method to avoid the negative impact of the fusion process while reducing the energy consumption. The method is able to dynamically switch between different sensor combinations. The optimization is formulated as a joint minimization problem between energy consumption and model loss. A hyperparameter allows to represent the maximum deviation from the maximum performing configuration, exploring more efficient configurations. The workflow goes as follows: sensor measurements are passed through modality-specific stem models, which produce an initial set of features. A gate model then estimates the loss based on the set of features and the possible fusion configurations. The best optimization approaches are then selected, running the select branches for object detection, the outputs of which are then fused in a late fusion block. Four different gating approaches have been tested: knowledge-based, MLP-based, attention-based, and loss-based. This flexibility allows the model to dynamically choose between early fusion, late fusion, no fusion, or a combination of the three. \\% task: object detection, techniques: top-k selection, type of fusion: hybrid fusion (early and late)
ROMANUS \citep{chenRomanusRobustTask2022} tries to solve the problem of offloading multi-sensor process pipelines to edge computing units, either on the side of the road or on cellular base station, supported by low latency and high bandwidth communications. The methodology introduces efficient offloading points along the execution path of the deep learning models, while a runtime solution based on Reinforcement Learning adapts the operating mode according to the scene complexity, leveraging both contextual and temporal correlations. It focuses on a late fusion, ensuring that at least one pipeline is always processed locally in the vehicle for safety reasons, while dynamically routing or avoiding computations according to the total energy consumption and the execution latency for processing the input given the modality. The architecture is designed in an encoder-decoder fashion to accommodate the shrinking of the input data to a lower dimensional space for reducing energy footprint and latency in the transmission phase to the edge unit. A contextual encoder is embedded in the architecture, which allows the decision on the processing pipeline for a fixed window, with the backing idea that successive frames share a similar driving context. Once the current state has been identified, deep Reinforcement Learning is utilized to create a policy that makes the offloading decision. \\
% task: object detection, techniques: reinforcement learning, type of fusion: late fusion
\citet{caoMultimodalGatedMixture2023} proposes a dynamic image fusion technique with a multi-modal mixture of local-to-global experts to perform object detection. The local experts specialize in the learning of multi-modal local features within the input, while the global experts focus on the information of the fused image, with the overall texture details and contrast. FasterRCNN is used as an auxiliary detection network with a spatial attention module to learn attention maps useful to guide the mixture of local experts.
% task: object detection, techniques: (local and global) mixture of experts type of fusion: hybrid (early and late fusion)
}
\paragraph{Semantic Segmentation}{
\hypertarget{para:semanticsegmentation}{}\label{para:semanticsegmentation}
\citet{valadaAdapNetAdaptiveSemantic2017} propose AdapNet a dynamic architecture to perform semantic segmentation. It incorporates a convolutional MoE to dynamically fuse different modalities, where each expert is previously trained in a specific subset of the input space. Specifically, the paper proposes the fusion of RGB and depth modalities through the use of an adaptive gating network. Both modalities produce as intermediate output a segmentation mask. The gating mechanism then proceeds to act as a decision agent, which is allowed to pick one of the modality outputs for each class in the segmentation task. The network is trained through a multi-stage training scheme, where the single experts are trained individually, and in a second phase, the gating parameters are also learned. Every dimensionality-reducing stage is split into convolutions applied at different scales based on ResNet. \\ % task: segmentation, techniques: moe, type of fusion: late fusion 
Furthermore, \citet{blumModularSensorFusion2018} proposes a statistical late fusion approach to perform semantic segmentation starting from the output of expert networks. The method allows to have different separately trained experts (even on different datasets) to make a per-pixel proposal, which is then fused without the need to retrain the fusing method. Both Bayesian fusion and Dirichlet fusion are presented in the paper, showing how, in general, the latter seems to generalize better. \\ % task: segmentation, gating: moe, bayesian, dirichlet, type of fusion: late fusion 
More recently, \citet{zhangTMFormerTokenMerging2024b} introduces the technique of token merging in the case of missing modalities for the task of brain tumor segmentation. The core idea is to reduce the model size by extracting and merging accessible modalities into more compact token sequences. The architecture incorporates two main blocks: a unimodal token merging block focuses on information extraction from individual modalities by compressing spatially redundant tokens, and a multi-modal token merging block, responsible for the fusion of the tokens from the different modalities to enhance global representation.
}

\paragraph{Autonomous Vision-based Navigation}{\hypertarget{para:autonomous}{}\label{para:autonomous}\citet{johnEstimationSteeringAngle2018} proposes to solve the problem of estimating the steering angle in autonomous driving by fusing the RGB image with a second-order particle filtering algorithm output. The particle filter is modeled for varying road scenes and driving patterns. Given a road scene, individual proposals and likelihood distributions are modeled with a Deep Learning-based regression model for both normal driving and collision avoidance, and then the final proposal distribution is modeled using a MoE framework based on Long Short Term Memory. Multiple experts model the regression function, while a gating network weights out the output of different experts. The research problem is formulated as a vision-based filtering problem, where the expert steering angles are estimated from a posterior distribution. % task: regression, techniques: moe, type of fusion: late fusion

\citet{chenSelectiveSensorFusion2019} propose an end-to-end Sensor Fusion framework for monocular visual-inertial odometry that fuses selectively monocular images and inertial measurements (IMU). This allows to estimate the trajectory taking into account missing and corrupted data. The model proposes two fusion methods: a deterministic soft fusion and a hard fusion method based on the Gumbel-Softmax resampling. In particular, this last one allows to generate a binary mask that either propagates the feature or blocks it according to the environment and its reliability. % task: odometry, techniques: gumbel-softmax, type of fusion: late fusion

A more recent work that also uses odometry is MIXO \citep{morraMIXOMixtureExpertsBased2023}. The paper proposes an architecture that combines odometry from multiple cameras to obtain a more accurate and robust global estimate in the context of autonomous driving. Each camera is processed by a visual odometry algorithm, and then these estimates are mixed by a gating network, which selects the locally optimal experts in the current operational conditions and weights their contributions. The method can be implemented on top of any visual odometry algorithm. The selection phase of the optimal camera and respective visual odometry is done through the use of a Mixture of Experts technique, choosing the best experts to solve the problem locally given the operation conditions present in the moment. % task: odometry, techniques: moe, type of fusion: late fusion

CARMA \citep{zhangCARMAContextAwareRuntime2023} is a framework proposing a fusion approach that uses the context to dynamically reconfigure the computation flow on a field-programmable gate array (FPGA) at runtime. This is performed by clock-gating unused sensors and model sub-components, with the aim of reducing the energy consumption. The architecture allows an optimization of the whole system, including sensors, model architecture, and hardware platform. A context identification process is enabled intermittently. The advantage of such a solution is to have global control over the overall power consumption of the whole system and not only the model part. % task: any, techniques: clock gating, type of fusion: -
 }
 \paragraph{Other Sensor Fusion approaches}{
 \hypertarget{para:othersensor}{}\label{para:othersensor}
 \label{paragraph:sfothersensors}
In this paragraph, we present also two papers that involve Sensor Fusion outside of the scope of vision. Nevertheless, we feel these solutions are relevant and easily applicable in the context of Vision. SELF-CARE \citep{rashidSELFCARESelectiveFusion2022, rashidStressDetectionUsing2023a} is a framework that introduces context-aware Sensor Fusion for stress detection on embedded devices. It selectively allows fusion based on motion to maximize classification performance while ensuring energy efficiency. Furthermore, the late fusion for classification incorporates also th temporal dynamic based on a Kalman filter. The model first extracts the features that are directly related to motion. They are then processed by a gating model to select the best-performing branch. A decision tree classifier is used for the gating model since lightweight architectures are a strict requirement for wearable devices. Furthermore, a term is introduced that allows to induce some direction on the conservativeness of the gating fusion module. After the gating has been inferred, early fusion is performed with the selected features to a specifically selected classifier for the specific combination of sensors picked. Different machine learning classifiers (decision tree, random forest, AdaBoost, linear discriminant analysis, and KNN) are evaluated in the experiments, and the classifiers with minimum loss are selected to be included in the architecture. Furthermore, a late fusion module of the output based on Kalman filters is also included in order to keep track of the temporal dynamics.

 % task: classification, techniques: kalman filters, mixture of experts, type of fusion: hybrid (early and late fusion)
AdaCoMed \citep{chenMultimodalMedicalDiagnosis} presents a framework that combines different sized models for medical diagnosis. The paper presents a mix of Vision and Language models through the use of a MoE architecture. It combines features extracted from multiple
single-modal medical large models in combination with multi-modal small models, through a co-learning mechanism which properly weights the strengths of each type of architecture.

 AstroFusion \citep{mortlockAdaptiveDataFusion2024} is a data fusion framework that performs power grid state estimation in order to create robustness towards power grid attacks. It uses an ensemble of ML-aided estimators to perform late adaptive fusions, effectively learning to avoid counting in the estimation areas of the grid that may be under attack. In this use case, the techniques also allow to identify which of the sensors is subject to attack, allowing a hypothetical operator to verify the source of the problem. The training of the model is performed by adding randomly strong noise to some of the sensors used to make the prediction and driving the network to learn to avoid these cases. % task: regression, techniques: ensemble, type of fusion: late fusion
 }

% \begin{table}
%     \caption{Summary of Dynamic Routing techniques presented in this survey (1/2). }
%     \renewcommand{\arraystretch}{0.88}
%     \begin{adjustbox}{width=\textwidth}
%         \begin{tabular}{@{\extracolsep\fill}rp{7cm}p{2cm}p{3cm}l}
        
\begin{table}
    \caption{Summary of Dynamic Sensor Fusion techniques presented in this survey.}
    \label{tab:contributions_3}
    \renewcommand{\arraystretch}{0.88} 
    \begin{adjustbox}{width=\textwidth}
    \begin{tabular}{p{2cm}p{4cm}p{3cm}ll}
        \toprule
        \textbf{Paper} & \textbf{Summary} & \textbf{Sensors} & \textbf{Method} & \textbf{Task} \\
        \midrule
        \citet{meesChoosingSmartlyAdaptive2016b} & Convolutional MoE for RGB, depth and motion  & RGB, Depth & Mixture-of-Experts & Object detection \\
        \citet{valadaAdapNetAdaptiveSemantic2017} & Convolutional MoE to dynamically fuse different modalities  & RGB, Depth, EVI & Mixture-of-Experts & Semantic segmentation \\
        \citet{blumModularSensorFusion2018} & Late fusion approach for semantic segmentation from the output of separately trained experts & RGB, Depth & Mixture-of-Experts & Semantic segmentation \\
        \citet{johnEstimationSteeringAngle2018} & Road scenes and driving patterns based fusion for steering angle estimation & RGB & Mixture-of-Experts & Steering prediction \\
        \citet{chenSelectiveSensorFusion2019} & Selective fusion of images and IMU & RGB, IMU & Gumbel-Softmax fusion & Odometry \\
        \citet{chenRomanusRobustTask2022} & Dynamic offload of sensor process to edge computing units & RGB Stereo, LiDAR, Radar & Early Exits & Object detection \\
        \citet{malawadeEcoFusionEnergyawareAdaptive2022} & Adds the environmental context to Dynamic Sensor Fusion & RGB, LiDAR, Radar & Dynamic Architecture & Object detection \\
        \citet{malawadeHydraFusionContextAwareSelective2022} & Sensor fusion selection to perform object detection & RGB, LiDAR, Radar & Dynamic Architecture & Object detection \\
        \citet{caoMultimodalGatedMixture2023} & Multi-modal mixture of local-to-global experts & RGB, IR & Mixture-of-Experts & Object detection \\
        \citet{morraMIXOMixtureExpertsBased2023} & Local optimal expert selection for multicamera visual odometry & RGB & Mixture-of-Experts & Odometry \\
        \citet{rashidStressDetectionUsing2023a} & Context-aware Sensor Fusion for stress detection on embedded devices & Wearable Sensors & Mixture-of-Experts & Stress detection \\
        \citet{zhangCARMAContextAwareRuntime2023} & Fusion approach  to dynamically reconfigure FPGA at runtime & Any & Dynamic Architecture & Any \\
        \citet{mortlockAdaptiveDataFusion2024} & Fusion for power grid state estimation to create robustness towards attacks & Grid Power Sensors & Dynamic Architecture & Regression \\
        \citet{zhangTMFormerTokenMerging2024b} & Token merging for brain tumor segmentation & FLAIR, T1ce, T1, and T2 & Token merging & Semantic segmentation \\
        \bottomrule
    \end{tabular}
    
    \end{adjustbox}
\end{table}

\section{Discussion}
\label{section:discussion}
 % Status & positive aspects
Since the rise of deep learning solutions in Computer Vision, Dynamic Neural Network models have seen increasing adoption, as highlighted by the increasing trend of publications over time (cf. Figure \ref{fig:trenddynncvpapers}). 
% Status of the literature
The topic of Early Exits is the most popular within the relevant literature, with a large portion of the research effort is spent in understanding both the different types of architectures and the different improvement aspects of those: from optimal exiting policies and positioning, correct weighting for the loss contribution of intermediate classifiers, mismatches between the training and inference phase, and so on. Based on the analysis conducted in this survey, the number of publications in this area has consistently grown since 2016, with activity stabilizing after 2020. \\
For techniques involving dynamic computational paths, the research landscape is extensive but more diverse in its approaches. Different types of solutions have been presented: from networks that are constructed on the fly by composing different computation blocks according to the input, to the dynamic skip of certain computation units, to the sub-specialization of specialized ``expert" blocks activated only to solve certain situations, in the case of MoE. The topic has seen a peak in publications around 2018, targeting more the dynamicity of CNN blocks, followed by a peak in 2020, with the rise of MoE techniques aimed at enabling efficient training of larger models.\\
Token Skimming is a relatively recent topic due to the fact that it emerged alongside the Transformer architecture. This topic has caught on very quickly, due to the simple conceptual idea, as well as the possibility to apply it easily in the fine-tuning process, or sometimes even without the need of further training at all. \\
Finally, the application of these techniques to Dynamic Sensor Fusion has been fluctuating. The field saw an initial rise of papers on the topic between 2017 and 2019, and a second wave of renewed attention since 2022.

\subsection{Standardized evaluation template for Dynamic Neural Networks}
As discussed throughout this survey, the domain of Dynamic Neural Networks currently lacks a standardized experimental protocol. The current literature largely relies on comparisons within the same category of methodologies and often utilizing a diverse array of datasets and evaluation baselines which makes comparisons unfeasible. This directly translates into infeasible quantitative analysis without extensive re-implementation. Furthermore, conventional metrics, such as static accuracy and overall FLOPs, are often not sufficient to capture the adaptive nature of these models. We argue that a structured benchmark is essential to consistently compare different Dynamic Neural Network models. Such an evaluation must specifically account for actual inference speed and efficiency, the variability of computations in relation to input complexity, and performance quality in edge cases. To address this critical gap and facilitate future cross-study analysis, we propose a draft formulation of a standardized evaluation protocol for reporting quantitative results.
We believe that there are four core aspects that a standardized template must address: the trade-off between performance and efficiency, theoretical versus hardware efficiency, a quantification of the dynamicity and a direct comparison with a static equivalent baseline. We proceed by detailing these proposed criteria below.

\paragraph{Performance-efficiency trade-off} Static networks yield a single operational point in terms of accuracy and computational cost (FLOPs). In contrast, Dynamic Neural Networks inherently yield a continuous trade-off curve based on their budget thresholds (e.g., confidence levels in Early Exits). To establish a benchmark that is strictly table-comparable without relying on arbitrary discrete "cherry picked" configurations, we propose standardizing evaluations around two rigorous metrics: the Area Under the Curve (AUC) of the accuracy-computation trade-off, and the computational variance. Utilizing the AUC summarizes the network's overall resilience across all computational budgets into a single metric. Reporting the average computational cost (e.g., Average GFLOPs) alongside its standard deviation allows to quantify the network's dynamicity, proving that it actively redistributes computations based on sample complexity. 

\paragraph{Computational efficiency versus hardware reality} There is a known discrepancy in the current literature between theoretical computational cost and actual inference speed. A dynamic router might theoretically save 30\% of FLOPs, but inadvertently slow down the actual inference due to fragmented memory access. Hardware accelerators, such as GPUs and TPUs, inherently favor static, dense computations. Furthermore, memory bandwidth and data movement costs often dominate the actual inference latency in edge devices. We argue benchmarks must explicitly state the hardware platform utilized and measure actual wall-clock time. To further standardize these empirical metrics, we encourage conducting evaluations on integrated System-on-Chip (SoC) platforms (e.g., the NVIDIA Jetson family). The use of these integrated platforms allows to reduce the measurement noise generated by the different hardware components (e.g. CPU, GPU, memory) present in standard desktop setups. Furthermore, when evaluating models intended for embedded systems, the energy consumption should also be reported.

\paragraph{Quantifying dynamicity} A critical, yet often overlooked, aspect of evaluation is the actual dynamic behavior of the network. A network is dynamic if the computational path can be adjusted to the input complexity. To effectively quantify this adaptability, we propose the introduction of specific metrics evaluated on a fixed, representative network configuration (e.g., a default or optimal budget threshold). Specifically, we suggest reporting the compute variance across the test set. Evaluating a single operational point allows to isolate and measure the true adaptivity of the model to the complexity of individual samples, rather than the variance generated by different hyperparameter settings. 

\paragraph{Fair baselines} When evaluating Dynamic Neural Networks, a direct comparison with their static counterpart enables a better quantification of the type and magnitude of the improvement introduced by the dynamic components. The direct comparison offers a clear intuition of the structural and computational advantages provided by the dynamicity with respect to a well-known architecture. Furthermore, the static baseline can serve as an effective proxy to compare different dynamic methodologies. In a heterogeneous literature landscape where training setups frequently vary, evaluating the relative improvements of different dynamic networks over the exact same static backbone (e.g., analyzing two distinct dynamic routing methods that both build upon a standard ResNet-50) allows to contextualize the results and provides a fairer ground for cross-paper comparisons.

\begin{table}[h!]
    \centering
    \caption{A proposed standardized evaluation template applied to landmark Dynamic Neural Networks. Extracting these metrics highlights historical inconsistencies: while recent Vision Transformer methods correctly report empirical Image Throughput, earlier literature often poorly standardized latency, and crucially, \textit{none} report AUC, computational variance, or energy metrics. Standardizing these fields is essential for fair, comprehensive cross-family comparisons. \textbf{Abbreviations:} \textbf{EE}: Early Exits, \textbf{DR}: Dynamic Routing, \textbf{TD}: Token Drop, \textbf{TM}: Token Merge. \textbf{RN}: ResNet, \textbf{DN}: DenseNet. \textbf{C-10}: CIFAR-10, \textbf{IN-1K}: ImageNet-1K. \textbf{GF}: GFLOPs. \textbf{$\sigma$}: Standard Deviation. \textbf{AUC}: Area Under the Curve.}
    \label{tab:standardized_evaluation_template}
    \begin{adjustbox}{width=\textwidth}
    \begin{tabular}{@{} l c l c l c c c c c c @{}}
    \toprule
    \textbf{Method} & \textbf{Type} & \textbf{Arch.} & \textbf{Data} & \textbf{Hardware} & 
    \begin{tabular}{@{}c@{}}\textbf{Static} \\ \textbf{(Acc / GF)}\end{tabular} & 
    \begin{tabular}{@{}c@{}}\textbf{Dynamic} \\ \textbf{(Acc / Avg GF $\pm \sigma$)}\end{tabular} & 
    \begin{tabular}{@{}c@{}}\textbf{AUC} \\ \textbf{(Acc-GF)}\end{tabular} & 
    \begin{tabular}{@{}c@{}}\textbf{Comput.} \\ \textbf{Savings}\end{tabular} & 
    \begin{tabular}{@{}c@{}}\textbf{Inference} \\ \textbf{Speed}\end{tabular} & 
    \begin{tabular}{@{}c@{}}\textbf{Energy} \\ \textbf{(mJ/im)}\end{tabular} \\
    \midrule
    BranchyNet & EE & RN-110 & C-10 & Titan X & 80.7\% / -- & -- / -- ($\pm$ --) & -- & -- & 37.2 ms & -- \\
    MSDNet & EE & Custom & IN-1K & -- & -- & $\sim$75.0\% / 1.7 ($\pm$ --) & -- & -- & -- & -- \\
    DVT & EE & T2T-ViT-12 & IN-1K & 2080 Ti & 76.7\% / 1.78 & 76.2\% / 1.00 ($\pm$ --) & -- & $\sim$43\% & 1128 im/s & -- \\
    SkipNet & DR & RN-101 & IN-1K & -- & 77.4\% / -- & -- / -- ($\pm$ --) & -- & 30\% & -- & -- \\
    BlockDrop & DR & RN-101 & IN-1K & P6000 & 76.4\% / 15.6 & 76.4\% / 12.5 ($\pm$ 0.43) & -- & 19.9\% & -- & -- \\
    ConvNet-AIG & DR & RN-101 & IN-1K & -- & 76.4\% / 7.6 & 77.5\% / 5.1 ($\pm$ --) & -- & 38\% & -- & -- \\
    DynamicViT & TD & LV-ViT-S & IN-1K & RTX 3090 & 83.3\% / 6.6 & 83.0\% / 4.6 ($\pm$ --) & -- & 31\% & 1417 im/s & -- \\
    EVIT & TM & DeiT-S & IN-1K & A100 & 79.8\% / 4.6 & 79.5\% / 3.0 ($\pm$ 0.0) & -- & 35\% & 4385 im/s & -- \\
    ToMe & TM & ViT-B (Aug) & IN-1K & V100 & 84.6\% / -- & 83.9\% / -- ($\pm$ 0.0) & -- & -- & 406 im/s & -- \\
    \bottomrule
    \end{tabular}
    \end{adjustbox}
\end{table}

To demonstrate the usefuleness of the proposed framework, Table \ref{tab:standardized_evaluation_template} presents a retrospective application of our criteria to a selection of landmark Dynamic Neural Networks. For this analysis, we selected three of the most representative publications for each main methodological category. Extracting these metrics directly from the original manuscripts highlights several inconsistencies and gaps in standard evaluation protocols. As evident from the missing values, none of the considered foundational works report the Area Under the Curve (AUC) or energy consumption, and only a minority explicitly detail the computational variance ($\sigma$). Furthermore, the table illustrates a clear evolution in how empirical inference speed is quantified. In earlier literature, such as ResNet-era routing architectures, GFLOPs and latency (e.g., in milliseconds) were often poorly standardized or omitted entirely. In contrast, more recently image throughput (images/sec) has become the standardized metric over single-image latency. Our proposed evaluation template highlights what should be reported to bridge reporting gaps, ensuring that future architectures are benchmarked comprehensively across theoretical sparsity, dynamic adaptivity, and practical hardware utility.

\subsection{Other challenges and limitations}
Even though Dynamic Neural Networks have exhibited a lot of merit and potential in Computer Vision and Sensor Fusion, their current state of the art does not come without challenges and limitations.\\
\textbf{Stability and convergence}. Incorporating dynamic behavior within a network often translates in the need of performing discrete decisions. The training of discrete decision modules tends to lead to instability during the training process, hindering the convergence toward an optimal result or in some cases destabilizing the whole process. Furthermore, the discretization sometimes induces different parts of the network to learn from different number of subsamples related to the specific task, resulting in a poor learning on certain subtasks. Exploring techniques that ensure that all network paths are adequately trained is a task that require further investigation.\\
\textbf{Deployability}. The deployments of Dynamic Neural Networks in production environments pose some additional challenges when compared with their static counterparts. Retaining the dynamicity of a model while still processing samples in batches for optimal execution is a non-trivial task and often deployment frameworks (e.g. TensorRT, ONNX) are not optimized for the dynamic operations.\\
\textbf{Efficiency}. Due to their adaptability, Dynamic Neural Network models are designed to contain their computation cost within an upper and a lower bound. Precisely controlling the amount of computation is still an open challenge.\\
% \textbf{Benchmarks and metrics}. The lack of standardized metrics and benchmarks tailored to Dynamic Neural Networks complicates the comparison among different methods. Conventional metrics (e.g., accuracy, FLOPs) often fail to adequately compare the adaptive nature of these model.\\
% \textbf{Benchmarks and metrics}. The lack of standardized metrics and benchmarks tailored to Dynamic Neural Networks complicates the comparison among different methods. Conventional metrics (e.g., accuracy, FLOPs) are often not sufficient to capture the adaptive nature of these models. Currently, the literature relies on comparisons within the most similar architecture, making a broader comparison among different approaches hard to make.\\
\textbf{Robustness}. Robustness to adversarial attacks is an underexplored area. Preliminary studies \citep{zhangRobustMixtureofExpertTraining2023a, panGradMDMAdversarialAttack2023a} provide some insights but are limited in scope. The full extent of possible attacks on any Dynamic Neural Network has yet to be explored. Furthermore, there are no studies comparing the robustness to attacks of these networks to static networks.

\subsection{Future directions}
The field of Dynamic Neural Networks has come a long way and many possible direction have arisen for future directions.\\
\textbf{Benchmark}
To enable fair cross-study analysis, the field would highly benefit from a standardized benchmark. As previously discussed, cross technique comparison is currently a challenge. We argue that such a benchmark must consistently evaluate not only accuracy but also inference speed, the variability of computations in relation to input complexity, and robustness in edge cases (e.g., small objects or targets located in the peripheral regions of the image).\\
\textbf{Architecture agnosticism}. Future research should prioritize broader diffusion of these techniques across diverse architectures. A crucial step is enabling any static pretrained network architecture to be converted to a dynamic model with minimal effort. For example, \citet{bolyaTokenMergingYour2023} and \citet{seolTokenMergingClass2023a} demonstrate approaches that introduce dynamicity with no extra parameters added to the original pretrained ViT, no need for model fine-tuning and a contained drop in accuracy with respect to the original model. This enables the solution to be plug-and-play to all ViTs classification models after the training phase with little additional effort. We argue that the development of post-training solutions that are architecture agnostic is to be preferred, mirroring the success of static efficiency methods like quantization and pruning.\\
\textbf{Dynamicity for large models}. Adaptive techniques can potentially also be used for fine-tuning large pretrained models. For example, \citet{devotoAdaptiveLayerSelection2024} introduces a fine-tuning method for ViT models based on a computed budget calculated at each step. An importance score for each layer is estimated, and only a meaningful subset of layers is selected for fine-tuning.\\
\textbf{Improve model elasticity}. As seen already in some preliminary studies, the dynamicity can be entangled to the resources capacity. This idea could be expanded to unexplored fields of Dynamic Neural Networks. For example, further improvements could work toward having a top-$k$ selection of experts with a variable $k$ according to both the input and the available processing power.\\
\textbf{Improved gating}. Research should focus on the improvement of the learning of gating modules, a fundamental component of Dynamic Neural Networks which is prone to instability during the training process. Advancements in this area could involve developing alternative gating mechanisms, specific loss functions, and robust optimization algorithms.\\
\textbf{Less represented classes}. Datasets frequently contain classes with limited representation. This can harm the recognition of those classes, especially when components of a network are dedicated to learning that specific subtask. Exploring few-shot learning in the context of Dynamic Neural Networks could help cover the edge cases for those components of a network that are subject to data scarcity.\\
\textbf{Sensor Fusion}. When it comes to Dynamic Neural Networks in the context of Sensor Fusion, there are many unexplored areas. The temporal aspect has been largely explored \citep{sabetTemporalEarlyExits2022, yangDyFADetDynamicFeature2024, chenEfficientVideoAction2023a}, and Dynamic Sensor Fusion could benefit from the ability to adaptively handle asynchronous or variable-rate data streams. Furthermore, different sensors produce data at different levels of detail, also depending on the context. Introducing dynamically the option of adapting the granularity of fused representations could help both the efficiency and the performance. A fully modular solution should also be explored, where components of a network can be added and removed according to sensor availability, and the network can orchestrate among both sensors available and sensors needed for the task. Finally, ensuring safety and trust in multi-sensor systems is a key element in certain fields, like autonomous driving, and it requires both some understanding and robustness towards unknown events.

\section{Conclusions}\label{conclusions}
% This survey focuses on Dynamic Neural Networks techniques for Computer Vision and their applications in Sensor Fusion. We presented a taxonomy based on a logical division of the techniques according to the parts of the networks that are dynamically influenced by the input. \\
This survey provides a comprehensive analysis of 163 Dynamic Neural Network methodologies for Computer Vision and their applications in Sensor Fusion. We propose a taxonomy categorizing the techniques according to their adaptive network component: \textit{Early Exits} (output-level), \textit{Dynamic Routing} (path-level), and \textit{Token Skimming} (input-level). Furthermore, we examine the specific advantages of these architectures for multi-modal tasks and supplement this review with a comprehensive, up-to-date repository of implementations to facilitate future research.
All models that constrained the network output position with respect to the input complexity fall under the Early Exits Section, where we presented the most important architectural solutions, relevant methods for improving the training, and the application of these methods to specific tasks and use cases. Then we proceeded with the review of Dynamic Routing techniques, where the computational path is created dynamically, either by constructing it on the fly or routing the computation to different model blocks. Also, in this case, a we present the application of these methods to various tasks. Lastly, in the Token Skimming section, we focused on Transformer-based models that either drop certain tokens along the block of the network (temporarily or permanently) or merge them together, based on an estimated token contribution to the desired task. The last part is dedicated to other methods and the applications of skimming to tasks that differ from image classification.\\
In the second half of the paper, we argue that Sensor Fusion is a field that could largely benefit from these adaptive techniques for multiple reasons: efficiency purposes, noise reduction due to different environmental conditions, and the ability of these techniques to prioritize the elaboration of critical information by design.\\
Furthermore, we provided a discussion on the current limitations and future directions. A considerable highlight is that Token Skimming allows for direct comparison on the ImageNet benchmark, where SaiT currently holds the best overall accuracy, whereas Early Exits and Dynamic Routing still lack a shared standard for consistent evaluation. For other method, a straight comparison is not trivial. We formulate a standardized evaluation template to address the lack of unified benchmarking protocols, with the scope of providing a fair and rigorous cross-methodology comparison methodology for future publications.\\
In conclusion, we hope this survey can encourage further progress in the field by providing a clear and structured taxonomy of Dynamic Neural Network techniques from a Computer Vision perspective, as well as highlighting their applicability to Sensor Fusion.

% \backmatter

\section*{Acknowledgements}

This work has been supported by Innovation Fund Denmark through the project ``Safety and Autonomy for Vehicles in Agriculture (SAVA)", 2105-00013A.

\section*{Data availability} 
The authors declare that all data supporting the findings of this survey are publicly available in the cited literature, as referenced throughout the manuscript.

% Example citation, See \citet{lamport94}.

% \bibliographystyle{elsarticle-harv} 
% \bibliography{refs}

\begin{thebibliography}{197}
\expandafter\ifx\csname natexlab\endcsname\relax\def\natexlab#1{#1}\fi
\providecommand{\url}[1]{\texttt{#1}}
\providecommand{\href}[2]{#2}
\providecommand{\path}[1]{#1}
\providecommand{\DOIprefix}{doi:}
\providecommand{\ArXivprefix}{arXiv:}
\providecommand{\URLprefix}{URL: }
\providecommand{\Pubmedprefix}{pmid:}
\providecommand{\doi}[1]{\href{http://dx.doi.org/#1}{\path{#1}}}
\providecommand{\Pubmed}[1]{\href{pmid:#1}{\path{#1}}}
\providecommand{\bibinfo}[2]{#2}
\ifx\xfnm\relax \def\xfnm[#1]{\unskip,\space#1}\fi
%Type = Article
\bibitem[{Abbas and Andreopoulos(2020)}]{abbasBiasedMixturesExperts2020}
\bibinfo{author}{Abbas, A.}, \bibinfo{author}{Andreopoulos, Y.}, \bibinfo{year}{2020}.
\newblock \bibinfo{title}{Biased {{Mixtures}} of {{Experts}}: {{Enabling Computer Vision Inference Under Data Transfer Limitations}}}.
\newblock \bibinfo{journal}{IEEE Transactions on Image Processing} \bibinfo{volume}{29}, \bibinfo{pages}{7656--7667}.
\newblock \DOIprefix\doi{10.1109/TIP.2020.3005508}.
%Type = Inproceedings
\bibitem[{Addad et~al.(2025)Addad, Lechervy and Jurie}]{addadCHASEChannelWiseSpatial2025}
\bibinfo{author}{Addad, Y.}, \bibinfo{author}{Lechervy, A.}, \bibinfo{author}{Jurie, F.}, \bibinfo{year}{2025}.
\newblock \bibinfo{title}{{{CHASE}}: {{Channel-Wise}} and {{Spatial Attention}} for {{Early Exiting}} in {{Image Classification}}}, in: \bibinfo{booktitle}{{{ICASSP}} 2025 - 2025 {{IEEE International Conference}} on {{Acoustics}}, {{Speech}} and {{Signal Processing}} ({{ICASSP}})}, pp. \bibinfo{pages}{1--5}.
\newblock \DOIprefix\doi{10.1109/ICASSP49660.2025.10888714}.
%Type = Inproceedings
\bibitem[{Ahmed et~al.(2016)Ahmed, Baig and Torresani}]{ahmedNetworkExpertsLargeScale2016}
\bibinfo{author}{Ahmed, K.}, \bibinfo{author}{Baig, M.H.}, \bibinfo{author}{Torresani, L.}, \bibinfo{year}{2016}.
\newblock \bibinfo{title}{Network of {{Experts}} for {{Large-Scale Image Categorization}}}, in: \bibinfo{editor}{Leibe, B.}, \bibinfo{editor}{Matas, J.}, \bibinfo{editor}{Sebe, N.}, \bibinfo{editor}{Welling, M.} (Eds.), \bibinfo{booktitle}{Computer {{Vision}} -- {{ECCV}} 2016}, \bibinfo{publisher}{Springer International Publishing}, \bibinfo{address}{Cham}. pp. \bibinfo{pages}{516--532}.
\newblock \DOIprefix\doi{10.1007/978-3-319-46478-7_32}.
%Type = Misc
\bibitem[{Bajpai and Hanawal(2024)}]{bajpaiCAPEENImageCaptioning2024}
\bibinfo{author}{Bajpai, D.J.}, \bibinfo{author}{Hanawal, M.}, \bibinfo{year}{2024}.
\newblock \bibinfo{title}{{{CAPEEN}}: {{Image Captioning}} with {{Early Exits}} and {{Knowledge Distillation}}}.
\newblock \DOIprefix\doi{10.48550/arXiv.2410.04433}.
%Type = Misc
\bibitem[{Bajpai and Hanawal(2025)}]{bajpaiBEEMBoostingPerformance2025}
\bibinfo{author}{Bajpai, D.J.}, \bibinfo{author}{Hanawal, M.K.}, \bibinfo{year}{2025}.
\newblock \bibinfo{title}{{{BEEM}}: {{Boosting Performance}} of {{Early Exit DNNs}} using {{Multi-Exit Classifiers}} as {{Experts}}}.
\newblock \DOIprefix\doi{10.48550/arXiv.2502.00745}, \href{http://arxiv.org/abs/2502.00745}{{\tt arXiv:2502.00745}}.
%Type = Misc
\bibitem[{Bakhtiarnia et~al.(2021)Bakhtiarnia, Zhang and Iosifidis}]{bakhtiarniaImprovingAccuracyEarly2021}
\bibinfo{author}{Bakhtiarnia, A.}, \bibinfo{author}{Zhang, Q.}, \bibinfo{author}{Iosifidis, A.}, \bibinfo{year}{2021}.
\newblock \bibinfo{title}{Improving the {{Accuracy}} of {{Early Exits}} in {{Multi-Exit Architectures}} via {{Curriculum Learning}}}.
\newblock \href{http://arxiv.org/abs/2104.10461}{{\tt arXiv:2104.10461}}.
%Type = Article
\bibitem[{Bakhtiarnia et~al.(2022)Bakhtiarnia, Zhang and Iosifidis}]{bakhtiarniaSinglelayerVisionTransformers2022}
\bibinfo{author}{Bakhtiarnia, A.}, \bibinfo{author}{Zhang, Q.}, \bibinfo{author}{Iosifidis, A.}, \bibinfo{year}{2022}.
\newblock \bibinfo{title}{Single-layer vision transformers for more accurate early exits with less overhead}.
\newblock \bibinfo{journal}{Neural Networks} \bibinfo{volume}{153}, \bibinfo{pages}{461--473}.
\newblock \DOIprefix\doi{10.1016/j.neunet.2022.06.038}.
%Type = Inproceedings
\bibitem[{Biggs et~al.(2023)Biggs, Bouganis and Constantinides}]{biggsATHEENAToolflowHardware2023a}
\bibinfo{author}{Biggs, B.}, \bibinfo{author}{Bouganis, C.S.}, \bibinfo{author}{Constantinides, G.}, \bibinfo{year}{2023}.
\newblock \bibinfo{title}{{{ATHEENA}}: {{A Toolflow}} for {{Hardware Early-Exit Network Automation}}}, in: \bibinfo{booktitle}{2023 {{IEEE}} 31st {{Annual International Symposium}} on {{Field-Programmable Custom Computing Machines}} ({{FCCM}})}, \bibinfo{publisher}{IEEE Computer Society}. pp. \bibinfo{pages}{121--132}.
\newblock \DOIprefix\doi{10.1109/FCCM57271.2023.00022}.
%Type = Inproceedings
\bibitem[{Blum et~al.(2018)Blum, Gawel, Siegwart and Cadena}]{blumModularSensorFusion2018}
\bibinfo{author}{Blum, H.}, \bibinfo{author}{Gawel, A.}, \bibinfo{author}{Siegwart, R.}, \bibinfo{author}{Cadena, C.}, \bibinfo{year}{2018}.
\newblock \bibinfo{title}{Modular {{Sensor Fusion}} for {{Semantic Segmentation}}}, in: \bibinfo{booktitle}{2018 {{IEEE}}/{{RSJ International Conference}} on {{Intelligent Robots}} and {{Systems}} ({{IROS}})}, pp. \bibinfo{pages}{3670--3677}.
\newblock \DOIprefix\doi{10.1109/IROS.2018.8593786}.
%Type = Misc
\bibitem[{Bolukbasi et~al.(2017)Bolukbasi, Wang, Dekel and Saligrama}]{bolukbasiAdaptiveNeuralNetworks2017}
\bibinfo{author}{Bolukbasi, T.}, \bibinfo{author}{Wang, J.}, \bibinfo{author}{Dekel, O.}, \bibinfo{author}{Saligrama, V.}, \bibinfo{year}{2017}.
\newblock \bibinfo{title}{Adaptive {{Neural Networks}} for {{Efficient Inference}}}.
\newblock \href{http://arxiv.org/abs/1702.07811}{{\tt arXiv:1702.07811}}.
%Type = Inproceedings
\bibitem[{Bolya et~al.(2023)Bolya, Fu, Dai, Zhang, Feichtenhofer and Hoffman}]{bolyaTokenMergingYour2023}
\bibinfo{author}{Bolya, D.}, \bibinfo{author}{Fu, C.Y.}, \bibinfo{author}{Dai, X.}, \bibinfo{author}{Zhang, P.}, \bibinfo{author}{Feichtenhofer, C.}, \bibinfo{author}{Hoffman, J.}, \bibinfo{year}{2023}.
\newblock \bibinfo{title}{Token merging: Your vit but faster}, in: \bibinfo{booktitle}{The Eleventh International Conference on Learning Representations}.
\newblock \URLprefix \url{https://iclr.cc/virtual/2023/oral/12533}.
%Type = Inproceedings
\bibitem[{Bolya and Hoffman(2023)}]{bolyaTokenMergingFast2023}
\bibinfo{author}{Bolya, D.}, \bibinfo{author}{Hoffman, J.}, \bibinfo{year}{2023}.
\newblock \bibinfo{title}{Token {{Merging}} for {{Fast Stable Diffusion}}}, in: \bibinfo{booktitle}{2023 {{IEEE}}/{{CVF Conference}} on {{Computer Vision}} and {{Pattern Recognition Workshops}} ({{CVPRW}})}, \bibinfo{publisher}{IEEE}, \bibinfo{address}{Vancouver, BC, Canada}. pp. \bibinfo{pages}{4599--4603}.
\newblock \DOIprefix\doi{10.1109/CVPRW59228.2023.00484}.
%Type = Inproceedings
\bibitem[{Cao et~al.(2023)Cao, Sun, Zhu and Hu}]{caoMultimodalGatedMixture2023}
\bibinfo{author}{Cao, B.}, \bibinfo{author}{Sun, Y.}, \bibinfo{author}{Zhu, P.}, \bibinfo{author}{Hu, Q.}, \bibinfo{year}{2023}.
\newblock \bibinfo{title}{Multi-modal {{Gated Mixture}} of {{Local-to-Global Experts}} for {{Dynamic Image Fusion}}}, in: \bibinfo{booktitle}{2023 {{IEEE}}/{{CVF International Conference}} on {{Computer Vision}} ({{ICCV}})}, \bibinfo{publisher}{IEEE}, \bibinfo{address}{Paris, France}. pp. \bibinfo{pages}{23498--23507}.
\newblock \DOIprefix\doi{10.1109/ICCV51070.2023.02153}.
%Type = Inbook
\bibitem[{Carion et~al.(2020)Carion, Massa, Synnaeve, Usunier, Kirillov and Zagoruyko}]{carionEndtoEndObjectDetection2020a}
\bibinfo{author}{Carion, N.}, \bibinfo{author}{Massa, F.}, \bibinfo{author}{Synnaeve, G.}, \bibinfo{author}{Usunier, N.}, \bibinfo{author}{Kirillov, A.}, \bibinfo{author}{Zagoruyko, S.}, \bibinfo{year}{2020}.
\newblock \bibinfo{title}{End-to-{{End Object Detection}} with {{Transformers}}}. \bibinfo{publisher}{Springer International Publishing}, \bibinfo{address}{Cham}.
\newblock pp. \bibinfo{pages}{213--229}.
\newblock \DOIprefix\doi{10.1007/978-3-030-58452-8_13}.
%Type = Misc
\bibitem[{Chen et~al.(2019a)Chen, Rosa, Miao, Lu, Wu, Markham and Trigoni}]{chenSelectiveSensorFusion2019}
\bibinfo{author}{Chen, C.}, \bibinfo{author}{Rosa, S.}, \bibinfo{author}{Miao, Y.}, \bibinfo{author}{Lu, C.X.}, \bibinfo{author}{Wu, W.}, \bibinfo{author}{Markham, A.}, \bibinfo{author}{Trigoni, N.}, \bibinfo{year}{2019}a.
\newblock \bibinfo{title}{Selective {{Sensor Fusion}} for {{Neural Visual-Inertial Odometry}}}.
\newblock \href{http://arxiv.org/abs/1903.01534}{{\tt arXiv:1903.01534}}.
%Type = Inproceedings
\bibitem[{Chen et~al.(2022)Chen, Odema and Faruque}]{chenRomanusRobustTask2022}
\bibinfo{author}{Chen, L.}, \bibinfo{author}{Odema, M.}, \bibinfo{author}{Faruque, M.A.A.}, \bibinfo{year}{2022}.
\newblock \bibinfo{title}{Romanus: {{Robust Task Offloading}} in {{Modular Multi-Sensor Autonomous Driving Systems}}}, in: \bibinfo{booktitle}{Proceedings of the 41st {{IEEE}}/{{ACM International Conference}} on {{Computer-Aided Design}}}, pp. \bibinfo{pages}{1--8}.
\newblock \DOIprefix\doi{10.1145/3508352.3549356}, \href{http://arxiv.org/abs/2207.08865}{{\tt arXiv:2207.08865}}.
%Type = Inproceedings
\bibitem[{Chen et~al.(2023a)Chen, Tong, Song, Wu and Wang}]{chenEfficientVideoAction2023a}
\bibinfo{author}{Chen, L.}, \bibinfo{author}{Tong, Z.}, \bibinfo{author}{Song, Y.}, \bibinfo{author}{Wu, G.}, \bibinfo{author}{Wang, L.}, \bibinfo{year}{2023}a.
\newblock \bibinfo{title}{Efficient {{Video Action Detection}} with {{Token Dropout}} and {{Context Refinement}}}, in: \bibinfo{booktitle}{2023 {{IEEE}}/{{CVF International Conference}} on {{Computer Vision}} ({{ICCV}})}, \bibinfo{publisher}{IEEE}, \bibinfo{address}{Paris, France}. pp. \bibinfo{pages}{10354--10365}.
\newblock \DOIprefix\doi{10.1109/ICCV51070.2023.00953}.
%Type = Inproceedings
\bibitem[{Chen et~al.(2021a)Chen, Cheng, Gan, Yuan, Zhang and Wang}]{chenChasingSparsityVision2021a}
\bibinfo{author}{Chen, T.}, \bibinfo{author}{Cheng, Y.}, \bibinfo{author}{Gan, Z.}, \bibinfo{author}{Yuan, L.}, \bibinfo{author}{Zhang, L.}, \bibinfo{author}{Wang, Z.}, \bibinfo{year}{2021}a.
\newblock \bibinfo{title}{Chasing {{Sparsity}} in {{Vision Transformers}}: {{An End-to-End Exploration}}}, in: \bibinfo{booktitle}{Advances in {{Neural Information Processing Systems}}}, \bibinfo{publisher}{Curran Associates, Inc.}. pp. \bibinfo{pages}{19974--19988}.
%Type = Inproceedings
\bibitem[{Chen et~al.(2021b)Chen, Cheng, Gan, Yuan, Zhang and Wang}]{chenChasingSparsityVision2021}
\bibinfo{author}{Chen, T.}, \bibinfo{author}{Cheng, Y.}, \bibinfo{author}{Gan, Z.}, \bibinfo{author}{Yuan, L.}, \bibinfo{author}{Zhang, L.}, \bibinfo{author}{Wang, Z.}, \bibinfo{year}{2021}b.
\newblock \bibinfo{title}{Chasing {{Sparsity}} in {{Vision Transformers}}: {{An End-to-End Exploration}}}, in: \bibinfo{booktitle}{Advances in {{Neural Information Processing Systems}}}, \bibinfo{publisher}{Curran Associates, Inc.}. pp. \bibinfo{pages}{19974--19988}.
%Type = Article
\bibitem[{Chen et~al.(2025)Chen, Zhao, Yao, Zhang, Bu and Wang}]{chenMultimodalMedicalDiagnosis}
\bibinfo{author}{Chen, W.}, \bibinfo{author}{Zhao, Z.}, \bibinfo{author}{Yao, J.}, \bibinfo{author}{Zhang, Y.}, \bibinfo{author}{Bu, J.}, \bibinfo{author}{Wang, H.}, \bibinfo{year}{2025}.
\newblock \bibinfo{title}{Multi-modal {{Medical Diagnosis}} via {{Large-small Model Collaboration}}} .
%Type = Misc
\bibitem[{Chen et~al.(2020)Chen, Dai, Li, Gao and Song}]{chenLearningStopLearning2020}
\bibinfo{author}{Chen, X.}, \bibinfo{author}{Dai, H.}, \bibinfo{author}{Li, Y.}, \bibinfo{author}{Gao, X.}, \bibinfo{author}{Song, L.}, \bibinfo{year}{2020}.
\newblock \bibinfo{title}{Learning to {{Stop While Learning}} to {{Predict}}}.
\newblock \href{http://arxiv.org/abs/2006.05082}{{\tt arXiv:2006.05082}}.
%Type = Inproceedings
\bibitem[{Chen et~al.(2023b)Chen, Liu, Tang, Yi, Zhao and Han}]{chenSparseViTRevisitingActivation2023}
\bibinfo{author}{Chen, X.}, \bibinfo{author}{Liu, Z.}, \bibinfo{author}{Tang, H.}, \bibinfo{author}{Yi, L.}, \bibinfo{author}{Zhao, H.}, \bibinfo{author}{Han, S.}, \bibinfo{year}{2023}b.
\newblock \bibinfo{title}{{{SparseViT}}: {{Revisiting Activation Sparsity}} for {{Efficient High-Resolution Vision Transformer}}}, in: \bibinfo{booktitle}{2023 {{IEEE}}/{{CVF Conference}} on {{Computer Vision}} and {{Pattern Recognition}} ({{CVPR}})}, \bibinfo{publisher}{IEEE}, \bibinfo{address}{Vancouver, BC, Canada}. pp. \bibinfo{pages}{2061--2070}.
\newblock \DOIprefix\doi{10.1109/CVPR52729.2023.00205}.
%Type = Inproceedings
\bibitem[{Chen et~al.(2019b)Chen, Li, Bengio and Si}]{chenYouLookTwice2019}
\bibinfo{author}{Chen, Z.}, \bibinfo{author}{Li, Y.}, \bibinfo{author}{Bengio, S.}, \bibinfo{author}{Si, S.}, \bibinfo{year}{2019}b.
\newblock \bibinfo{title}{You {{Look Twice}}: {{GaterNet}} for {{Dynamic Filter Selection}} in {{CNNs}}}, in: \bibinfo{booktitle}{2019 {{IEEE}}/{{CVF Conference}} on {{Computer Vision}} and {{Pattern Recognition}} ({{CVPR}})}, \bibinfo{publisher}{IEEE}, \bibinfo{address}{Long Beach, CA, USA}. pp. \bibinfo{pages}{9164--9172}.
\newblock \DOIprefix\doi{10.1109/CVPR.2019.00939}.
%Type = Article
\bibitem[{Cheng et~al.(2024)Cheng, Zhang and Shi}]{chengSurveyDeepNeural2024}
\bibinfo{author}{Cheng, H.}, \bibinfo{author}{Zhang, M.}, \bibinfo{author}{Shi, J.Q.}, \bibinfo{year}{2024}.
\newblock \bibinfo{title}{A {{Survey}} on {{Deep Neural Network Pruning}}: {{Taxonomy}}, {{Comparison}}, {{Analysis}}, and {{Recommendations}}}.
\newblock \bibinfo{journal}{IEEE Transactions on Pattern Analysis and Machine Intelligence} \bibinfo{volume}{46}, \bibinfo{pages}{10558--10578}.
\newblock \DOIprefix\doi{10.1109/TPAMI.2024.3447085}.
%Type = Inproceedings
\bibitem[{Cheng et~al.(2025)Cheng, Yao, Yuan and Han}]{chengNotAllTokens2025}
\bibinfo{author}{Cheng, J.}, \bibinfo{author}{Yao, X.}, \bibinfo{author}{Yuan, X.}, \bibinfo{author}{Han, J.}, \bibinfo{year}{2025}.
\newblock \bibinfo{title}{Not {{All Tokens Matter All The Time}}: {{Dynamic Token Aggregation Towards Efficient Detection Transformers}}}, in: \bibinfo{booktitle}{Forty-Second {{International Conference}} on {{Machine Learning}}}.
%Type = Inproceedings
\bibitem[{Colleman et~al.(2021)Colleman, Verelst, Mei, Tuytelaars and Verhelst}]{collemanProcessorArchitectureOptimization2021}
\bibinfo{author}{Colleman, S.}, \bibinfo{author}{Verelst, T.}, \bibinfo{author}{Mei, L.}, \bibinfo{author}{Tuytelaars, T.}, \bibinfo{author}{Verhelst, M.}, \bibinfo{year}{2021}.
\newblock \bibinfo{title}{Processor {{Architecture Optimization}} for {{Spatially Dynamic Neural Networks}}}, in: \bibinfo{booktitle}{2021 {{IFIP}}/{{IEEE}} 29th {{International Conference}} on {{Very Large Scale Integration}} ({{VLSI-SoC}})}, pp. \bibinfo{pages}{1--6}.
\newblock \DOIprefix\doi{10.1109/VLSI-SoC53125.2021.9607013}.
%Type = Article
\bibitem[{{Correia-Silva} et~al.(2021){Correia-Silva}, Berriel, Badue, De~Souza and {Oliveira-Santos}}]{correia-silvaCopycatCNNAre2021}
\bibinfo{author}{{Correia-Silva}, J.R.}, \bibinfo{author}{Berriel, R.F.}, \bibinfo{author}{Badue, C.}, \bibinfo{author}{De~Souza, A.F.}, \bibinfo{author}{{Oliveira-Santos}, T.}, \bibinfo{year}{2021}.
\newblock \bibinfo{title}{Copycat {{CNN}}: {{Are}} random non-{{Labeled}} data enough to steal knowledge from black-box models?}
\newblock \bibinfo{journal}{Pattern Recognition} \bibinfo{volume}{113}, \bibinfo{pages}{107830}.
\newblock \DOIprefix\doi{10.1016/j.patcog.2021.107830}.
%Type = Article
\bibitem[{Deng et~al.(2020)Deng, Li, Han, Shi and Xie}]{dengModelCompressionHardware2020}
\bibinfo{author}{Deng, L.}, \bibinfo{author}{Li, G.}, \bibinfo{author}{Han, S.}, \bibinfo{author}{Shi, L.}, \bibinfo{author}{Xie, Y.}, \bibinfo{year}{2020}.
\newblock \bibinfo{title}{Model {{Compression}} and {{Hardware Acceleration}} for {{Neural Networks}}: {{A Comprehensive Survey}}}.
\newblock \bibinfo{journal}{Proceedings of the IEEE} \bibinfo{volume}{108}, \bibinfo{pages}{485--532}.
\newblock \DOIprefix\doi{10.1109/JPROC.2020.2976475}.
%Type = Article
\bibitem[{Devoto et~al.(2025)Devoto, Alvetreti, Pomponi, {Di Lorenzo}, Minervini and Scardapane}]{devotoAdaptiveLayerSelection2024}
\bibinfo{author}{Devoto, A.}, \bibinfo{author}{Alvetreti, F.}, \bibinfo{author}{Pomponi, J.}, \bibinfo{author}{{Di Lorenzo}, P.}, \bibinfo{author}{Minervini, P.}, \bibinfo{author}{Scardapane, S.}, \bibinfo{year}{2025}.
\newblock \bibinfo{title}{Adaptive layer and token selection for efficient fine-tuning of vision transformers}.
\newblock \bibinfo{journal}{Neurocomputing} \bibinfo{volume}{654}, \bibinfo{pages}{131216}.
\newblock \URLprefix \url{https://www.sciencedirect.com/science/article/pii/S0925231225018880}, \DOIprefix\doi{https://doi.org/10.1016/j.neucom.2025.131216}.
%Type = Inproceedings
\bibitem[{Devoto et~al.(2024)Devoto, Pomponi, Petruzzi, Di~Lorenzo and Scardapane}]{devotoAdaptiveSemanticToken2024}
\bibinfo{author}{Devoto, A.}, \bibinfo{author}{Pomponi, J.}, \bibinfo{author}{Petruzzi, S.}, \bibinfo{author}{Di~Lorenzo, P.}, \bibinfo{author}{Scardapane, S.}, \bibinfo{year}{2024}.
\newblock \bibinfo{title}{Adaptive {{Semantic Token Selection}} for {{AI-native Goal-oriented Communications}}}, in: \bibinfo{booktitle}{2024 {{IEEE Globecom Workshops}} ({{GC Wkshps}})}, pp. \bibinfo{pages}{1--6}.
\newblock \DOIprefix\doi{10.1109/GCWkshp64532.2024.11100347}.
%Type = Inproceedings
\bibitem[{Fang et~al.(2020)Fang, Zeng, Zhang, Xu and Zhang}]{fangFlexDNNInputAdaptiveOnDevice2020}
\bibinfo{author}{Fang, B.}, \bibinfo{author}{Zeng, X.}, \bibinfo{author}{Zhang, F.}, \bibinfo{author}{Xu, H.}, \bibinfo{author}{Zhang, M.}, \bibinfo{year}{2020}.
\newblock \bibinfo{title}{{{FlexDNN}}: {{Input-Adaptive On-Device Deep Learning}} for {{Efficient Mobile Vision}}}, in: \bibinfo{booktitle}{2020 {{IEEE}}/{{ACM Symposium}} on {{Edge Computing}} ({{SEC}})}, pp. \bibinfo{pages}{84--95}.
\newblock \DOIprefix\doi{10.1109/SEC50012.2020.00014}.
%Type = Inproceedings
\bibitem[{Fang et~al.(2025)Fang, Tang, Cao, Zhang, Tang and Lee}]{fangAttendNotAttended2025}
\bibinfo{author}{Fang, H.}, \bibinfo{author}{Tang, S.}, \bibinfo{author}{Cao, J.}, \bibinfo{author}{Zhang, E.}, \bibinfo{author}{Tang, F.}, \bibinfo{author}{Lee, T.Y.}, \bibinfo{year}{2025}.
\newblock \bibinfo{title}{Attend to {{Not Attended}}: {{Structure-then-Detail Token Merging}} for {{Post-training DiT Acceleration}}}, in: \bibinfo{booktitle}{Proceedings of the {{Computer Vision}} and {{Pattern Recognition Conference}}}, pp. \bibinfo{pages}{18083--18092}.
%Type = Article
\bibitem[{Farina et~al.(2024)Farina, Ahmad, Taha, Younes, Mesbah, Yu and Pedrycz}]{farinaSparsityTransformersSystematic2024a}
\bibinfo{author}{Farina, M.}, \bibinfo{author}{Ahmad, U.}, \bibinfo{author}{Taha, A.}, \bibinfo{author}{Younes, H.}, \bibinfo{author}{Mesbah, Y.}, \bibinfo{author}{Yu, X.}, \bibinfo{author}{Pedrycz, W.}, \bibinfo{year}{2024}.
\newblock \bibinfo{title}{Sparsity in transformers: {{A}} systematic literature review}.
\newblock \bibinfo{journal}{Neurocomputing} \bibinfo{volume}{582}, \bibinfo{pages}{127468}.
\newblock \DOIprefix\doi{10.1016/j.neucom.2024.127468}.
%Type = Inproceedings
\bibitem[{Fayyaz et~al.(2022)Fayyaz, Koohpayegani, Jafari, Sengupta, Joze, Sommerlade, Pirsiavash and Gall}]{fayyazAdaptiveTokenSampling2022}
\bibinfo{author}{Fayyaz, M.}, \bibinfo{author}{Koohpayegani, S.A.}, \bibinfo{author}{Jafari, F.R.}, \bibinfo{author}{Sengupta, S.}, \bibinfo{author}{Joze, H.R.V.}, \bibinfo{author}{Sommerlade, E.}, \bibinfo{author}{Pirsiavash, H.}, \bibinfo{author}{Gall, J.}, \bibinfo{year}{2022}.
\newblock \bibinfo{title}{Adaptive {{Token Sampling}} for~{{Efficient Vision Transformers}}}, in: \bibinfo{editor}{Avidan, S.}, \bibinfo{editor}{Brostow, G.}, \bibinfo{editor}{Ciss{\'e}, M.}, \bibinfo{editor}{Farinella, G.M.}, \bibinfo{editor}{Hassner, T.} (Eds.), \bibinfo{booktitle}{Computer {{Vision}} -- {{ECCV}} 2022}, \bibinfo{publisher}{Springer Nature Switzerland}, \bibinfo{address}{Cham}. pp. \bibinfo{pages}{396--414}.
\newblock \DOIprefix\doi{10.1007/978-3-031-20083-0_24}.
%Type = Misc
\bibitem[{Feng et~al.(2020)Feng, Hua, Lai, Huang, Li and Hua}]{fengLearningGenerateContentAware2020}
\bibinfo{author}{Feng, J.}, \bibinfo{author}{Hua, J.}, \bibinfo{author}{Lai, B.}, \bibinfo{author}{Huang, J.}, \bibinfo{author}{Li, X.}, \bibinfo{author}{Hua, X.s.}, \bibinfo{year}{2020}.
\newblock \bibinfo{title}{Learning to {{Generate Content-Aware Dynamic Detectors}}}.
\newblock \DOIprefix\doi{10.48550/arXiv.2012.04265}, \href{http://arxiv.org/abs/2012.04265}{{\tt arXiv:2012.04265}}.
%Type = Inproceedings
\bibitem[{Figurnov et~al.(2017)Figurnov, Collins, Zhu, Zhang, Huang, Vetrov and Salakhutdinov}]{figurnovSpatiallyAdaptiveComputation2017}
\bibinfo{author}{Figurnov, M.}, \bibinfo{author}{Collins, M.D.}, \bibinfo{author}{Zhu, Y.}, \bibinfo{author}{Zhang, L.}, \bibinfo{author}{Huang, J.}, \bibinfo{author}{Vetrov, D.}, \bibinfo{author}{Salakhutdinov, R.}, \bibinfo{year}{2017}.
\newblock \bibinfo{title}{Spatially {{Adaptive Computation Time}} for {{Residual Networks}}}, in: \bibinfo{booktitle}{2017 {{IEEE Conference}} on {{Computer Vision}} and {{Pattern Recognition}} ({{CVPR}})}, \bibinfo{publisher}{IEEE}, \bibinfo{address}{Honolulu, HI}. pp. \bibinfo{pages}{1790--1799}.
\newblock \DOIprefix\doi{10.1109/CVPR.2017.194}.
%Type = Misc
\bibitem[{Frankle and Carbin(2019)}]{frankleLotteryTicketHypothesis2019}
\bibinfo{author}{Frankle, J.}, \bibinfo{author}{Carbin, M.}, \bibinfo{year}{2019}.
\newblock \bibinfo{title}{The {{Lottery Ticket Hypothesis}}: {{Finding Sparse}}, {{Trainable Neural Networks}}}.
\newblock \DOIprefix\doi{10.48550/arXiv.1803.03635}, \href{http://arxiv.org/abs/1803.03635}{{\tt arXiv:1803.03635}}.
%Type = Inproceedings
\bibitem[{Fung et~al.(2017)Fung, Chen and Chen}]{fungSensorFusionReview2017}
\bibinfo{author}{Fung, M.L.}, \bibinfo{author}{Chen, M.Z.Q.}, \bibinfo{author}{Chen, Y.H.}, \bibinfo{year}{2017}.
\newblock \bibinfo{title}{Sensor fusion: {{A}} review of methods and applications}, in: \bibinfo{booktitle}{2017 29th {{Chinese Control And Decision Conference}} ({{CCDC}})}, pp. \bibinfo{pages}{3853--3860}.
\newblock \DOIprefix\doi{10.1109/CCDC.2017.7979175}.
%Type = Article
\bibitem[{Gambella et~al.(2025)Gambella, Pomponi, Scardapane and Roveri}]{gambellaNACHOSNeuralArchitecture2024a}
\bibinfo{author}{Gambella, M.}, \bibinfo{author}{Pomponi, J.}, \bibinfo{author}{Scardapane, S.}, \bibinfo{author}{Roveri, M.}, \bibinfo{year}{2025}.
\newblock \bibinfo{title}{{{NACHOS}}: {{Neural Architecture Search}} for {{Hardware-Constrained Early-Exit Neural Networks}}}.
\newblock \bibinfo{journal}{IEEE Transactions on Neural Networks and Learning Systems} , \bibinfo{pages}{1--14}\DOIprefix\doi{10.1109/TNNLS.2025.3588558}.
%Type = Inproceedings
\bibitem[{Gambella and Roveri(2023)}]{gambellaEDANASAdaptiveNeural2023}
\bibinfo{author}{Gambella, M.}, \bibinfo{author}{Roveri, M.}, \bibinfo{year}{2023}.
\newblock \bibinfo{title}{{{EDANAS}}: {{Adaptive Neural Architecture Search}} for {{Early Exit Neural Networks}}}, in: \bibinfo{booktitle}{2023 {{International Joint Conference}} on {{Neural Networks}} ({{IJCNN}})}, pp. \bibinfo{pages}{1--8}.
\newblock \DOIprefix\doi{10.1109/IJCNN54540.2023.10191876}.
%Type = Inproceedings
\bibitem[{Gao et~al.()Gao, Zhao, Dudziak, Mullins and Xu}]{gaoDynamicChannelPruning2019}
\bibinfo{author}{Gao, X.}, \bibinfo{author}{Zhao, Y.}, \bibinfo{author}{Dudziak, {\L}.}, \bibinfo{author}{Mullins, R.}, \bibinfo{author}{Xu, C.z.}, .
\newblock \bibinfo{title}{Dynamic channel pruning: Feature boosting and suppression}, in: \bibinfo{booktitle}{International Conference on Learning Representations}.
%Type = Inproceedings
\bibitem[{Gao et~al.(2023)Gao, Zhang, Qi and So}]{gaoDPACSHardwareAccelerated2023a}
\bibinfo{author}{Gao, Y.}, \bibinfo{author}{Zhang, B.}, \bibinfo{author}{Qi, X.}, \bibinfo{author}{So, H.K.H.}, \bibinfo{year}{2023}.
\newblock \bibinfo{title}{{{DPACS}}: {{Hardware Accelerated Dynamic Neural Network Pruning}} through {{Algorithm-Architecture Co-design}}}, in: \bibinfo{booktitle}{Proceedings of the 28th {{ACM International Conference}} on {{Architectural Support}} for {{Programming Languages}} and {{Operating Systems}}, {{Volume}} 2}, \bibinfo{publisher}{ACM}, \bibinfo{address}{Vancouver BC Canada}. pp. \bibinfo{pages}{237--251}.
\newblock \DOIprefix\doi{10.1145/3575693.3575728}.
%Type = Inproceedings
\bibitem[{Ghodrati et~al.(2021)Ghodrati, Bejnordi and Habibian}]{ghodratiFrameExitConditionalEarly2021a}
\bibinfo{author}{Ghodrati, A.}, \bibinfo{author}{Bejnordi, B.E.}, \bibinfo{author}{Habibian, A.}, \bibinfo{year}{2021}.
\newblock \bibinfo{title}{{{FrameExit}}: {{Conditional Early Exiting}} for {{Efficient Video Recognition}}}, in: \bibinfo{booktitle}{Proceedings of the {{IEEE}}/{{CVF Conference}} on {{Computer Vision}} and {{Pattern Recognition}}}, pp. \bibinfo{pages}{15608--15618}.
%Type = Misc
\bibitem[{Gholami et~al.(2021)Gholami, Kim, Dong, Yao, Mahoney and Keutzer}]{gholamiSurveyQuantizationMethods2021}
\bibinfo{author}{Gholami, A.}, \bibinfo{author}{Kim, S.}, \bibinfo{author}{Dong, Z.}, \bibinfo{author}{Yao, Z.}, \bibinfo{author}{Mahoney, M.W.}, \bibinfo{author}{Keutzer, K.}, \bibinfo{year}{2021}.
\newblock \bibinfo{title}{A {{Survey}} of {{Quantization Methods}} for {{Efficient Neural Network Inference}}}.
\newblock \DOIprefix\doi{10.48550/arXiv.2103.13630}, \href{http://arxiv.org/abs/2103.13630}{{\tt arXiv:2103.13630}}.
%Type = Article
\bibitem[{Goodale and Milner(1992)}]{goodaleSeparateVisualPathways1992}
\bibinfo{author}{Goodale, M.A.}, \bibinfo{author}{Milner, A.D.}, \bibinfo{year}{1992}.
\newblock \bibinfo{title}{Separate visual pathways for perception and action}.
\newblock \bibinfo{journal}{Trends in Neurosciences} \bibinfo{volume}{15}, \bibinfo{pages}{20--25}.
\newblock \DOIprefix\doi{10.1016/0166-2236(92)90344-8}.
%Type = Article
\bibitem[{G{\"o}rmez and Koyuncu(2024)}]{gormezClassBasedThresholding2024a}
\bibinfo{author}{G{\"o}rmez, A.}, \bibinfo{author}{Koyuncu, E.}, \bibinfo{year}{2024}.
\newblock \bibinfo{title}{Class {{Based Thresholding}} in {{Early Exit Semantic Segmentation Networks}}}.
\newblock \bibinfo{journal}{IEEE Signal Processing Letters} \bibinfo{volume}{31}, \bibinfo{pages}{1184--1188}.
\newblock \DOIprefix\doi{10.1109/LSP.2024.3386110}.
%Type = Article
\bibitem[{Gou et~al.(2021)Gou, Yu, Maybank and Tao}]{gouKnowledgeDistillationSurvey2021}
\bibinfo{author}{Gou, J.}, \bibinfo{author}{Yu, B.}, \bibinfo{author}{Maybank, S.J.}, \bibinfo{author}{Tao, D.}, \bibinfo{year}{2021}.
\newblock \bibinfo{title}{Knowledge {{Distillation}}: {{A Survey}}}.
\newblock \bibinfo{journal}{International Journal of Computer Vision} \bibinfo{volume}{129}, \bibinfo{pages}{1789--1819}.
\newblock \DOIprefix\doi{10.1007/s11263-021-01453-z}.
%Type = Inproceedings
\bibitem[{Gross et~al.(2017)Gross, Ranzato and Szlam}]{grossHardMixturesExperts2017}
\bibinfo{author}{Gross, S.}, \bibinfo{author}{Ranzato, M.}, \bibinfo{author}{Szlam, A.}, \bibinfo{year}{2017}.
\newblock \bibinfo{title}{Hard {{Mixtures}} of {{Experts}} for {{Large Scale Weakly Supervised Vision}}}, in: \bibinfo{booktitle}{2017 {{IEEE Conference}} on {{Computer Vision}} and {{Pattern Recognition}} ({{CVPR}})}, \bibinfo{publisher}{IEEE}, \bibinfo{address}{Honolulu, HI}. pp. \bibinfo{pages}{5085--5093}.
\newblock \DOIprefix\doi{10.1109/CVPR.2017.540}.
%Type = Misc
\bibitem[{Han et~al.(2024)Han, Wei, Dou, Wang, Qiang, He, Sun, Han and Tian}]{hanViMoEEmpiricalStudy2024}
\bibinfo{author}{Han, X.}, \bibinfo{author}{Wei, L.}, \bibinfo{author}{Dou, Z.}, \bibinfo{author}{Wang, Z.}, \bibinfo{author}{Qiang, C.}, \bibinfo{author}{He, X.}, \bibinfo{author}{Sun, Y.}, \bibinfo{author}{Han, Z.}, \bibinfo{author}{Tian, Q.}, \bibinfo{year}{2024}.
\newblock \bibinfo{title}{{{ViMoE}}: {{An Empirical Study}} of {{Designing Vision Mixture-of-Experts}}}.
\newblock \DOIprefix\doi{10.48550/arXiv.2410.15732}, \href{http://arxiv.org/abs/2410.15732}{{\tt arXiv:2410.15732}}.
%Type = Inproceedings
\bibitem[{Han et~al.(2023)Han, Han, Liu, Wang, Pan, Pu, Deng, Feng, Song and Huang}]{hanDynamicPerceiverEfficient2023}
\bibinfo{author}{Han, Y.}, \bibinfo{author}{Han, D.}, \bibinfo{author}{Liu, Z.}, \bibinfo{author}{Wang, Y.}, \bibinfo{author}{Pan, X.}, \bibinfo{author}{Pu, Y.}, \bibinfo{author}{Deng, C.}, \bibinfo{author}{Feng, J.}, \bibinfo{author}{Song, S.}, \bibinfo{author}{Huang, G.}, \bibinfo{year}{2023}.
\newblock \bibinfo{title}{Dynamic {{Perceiver}} for {{Efficient Visual Recognition}}}, in: \bibinfo{booktitle}{2023 {{IEEE}}/{{CVF International Conference}} on {{Computer Vision}} ({{ICCV}})}, pp. \bibinfo{pages}{5969--5979}.
\newblock \DOIprefix\doi{10.1109/ICCV51070.2023.00551}.
%Type = Misc
\bibitem[{Han et~al.(2021)Han, Huang, Song, Yang, Wang and Wang}]{hanDynamicNeuralNetworks2021a}
\bibinfo{author}{Han, Y.}, \bibinfo{author}{Huang, G.}, \bibinfo{author}{Song, S.}, \bibinfo{author}{Yang, L.}, \bibinfo{author}{Wang, H.}, \bibinfo{author}{Wang, Y.}, \bibinfo{year}{2021}.
\newblock \bibinfo{title}{Dynamic {{Neural Networks}}: {{A Survey}}}.
\newblock \DOIprefix\doi{10.48550/arXiv.2102.04906}, \href{http://arxiv.org/abs/2102.04906}{{\tt arXiv:2102.04906}}.
%Type = Inproceedings
\bibitem[{Han et~al.(2022)Han, Pu, Lai, Wang, Song, Cao, Huang, Deng and Huang}]{hanLearningWeightSamples2022}
\bibinfo{author}{Han, Y.}, \bibinfo{author}{Pu, Y.}, \bibinfo{author}{Lai, Z.}, \bibinfo{author}{Wang, C.}, \bibinfo{author}{Song, S.}, \bibinfo{author}{Cao, J.}, \bibinfo{author}{Huang, W.}, \bibinfo{author}{Deng, C.}, \bibinfo{author}{Huang, G.}, \bibinfo{year}{2022}.
\newblock \bibinfo{title}{Learning to~{{Weight Samples}} for~{{Dynamic Early-Exiting Networks}}}, in: \bibinfo{editor}{Avidan, S.}, \bibinfo{editor}{Brostow, G.}, \bibinfo{editor}{Ciss{\'e}, M.}, \bibinfo{editor}{Farinella, G.M.}, \bibinfo{editor}{Hassner, T.} (Eds.), \bibinfo{booktitle}{Computer {{Vision}} -- {{ECCV}} 2022}, \bibinfo{publisher}{Springer Nature Switzerland}, \bibinfo{address}{Cham}. pp. \bibinfo{pages}{362--378}.
\newblock \DOIprefix\doi{10.1007/978-3-031-20083-0_22}.
%Type = Inproceedings
\bibitem[{Havtorn et~al.(2023)Havtorn, Royer, Blankevoort and Bejnordi}]{havtornMSViTDynamicMixedscale2023}
\bibinfo{author}{Havtorn, J.D.}, \bibinfo{author}{Royer, A.}, \bibinfo{author}{Blankevoort, T.}, \bibinfo{author}{Bejnordi, B.E.}, \bibinfo{year}{2023}.
\newblock \bibinfo{title}{{{MSViT}}: {{Dynamic Mixed-scale Tokenization}} for {{Vision Transformers}}}, in: \bibinfo{booktitle}{2023 {{IEEE}}/{{CVF International Conference}} on {{Computer Vision Workshops}} ({{ICCVW}})}, \bibinfo{publisher}{IEEE}, \bibinfo{address}{Paris, France}. pp. \bibinfo{pages}{838--848}.
\newblock \DOIprefix\doi{10.1109/ICCVW60793.2023.00091}.
%Type = Misc
\bibitem[{Hazimeh et~al.(2021)Hazimeh, Zhao, Chowdhery, Sathiamoorthy, Chen, Mazumder, Hong and Chi}]{hazimehDSelectkDifferentiableSelection2021a}
\bibinfo{author}{Hazimeh, H.}, \bibinfo{author}{Zhao, Z.}, \bibinfo{author}{Chowdhery, A.}, \bibinfo{author}{Sathiamoorthy, M.}, \bibinfo{author}{Chen, Y.}, \bibinfo{author}{Mazumder, R.}, \bibinfo{author}{Hong, L.}, \bibinfo{author}{Chi, E.H.}, \bibinfo{year}{2021}.
\newblock \bibinfo{title}{{{DSelect-k}}: {{Differentiable Selection}} in the {{Mixture}} of {{Experts}} with {{Applications}} to {{Multi-Task Learning}}}.
\newblock \DOIprefix\doi{10.48550/arXiv.2106.03760}, \href{http://arxiv.org/abs/2106.03760}{{\tt arXiv:2106.03760}}.
%Type = Inproceedings
\bibitem[{He et~al.(2016)He, Zhang, Ren and Sun}]{heDeepResidualLearning2015}
\bibinfo{author}{He, K.}, \bibinfo{author}{Zhang, X.}, \bibinfo{author}{Ren, S.}, \bibinfo{author}{Sun, J.}, \bibinfo{year}{2016}.
\newblock \bibinfo{title}{Deep {{Residual Learning}} for {{Image Recognition}}}, in: \bibinfo{booktitle}{2016 {{IEEE Conference}} on {{Computer Vision}} and {{Pattern Recognition}} ({{CVPR}})}, pp. \bibinfo{pages}{770--778}.
\newblock \DOIprefix\doi{10.1109/CVPR.2016.90}.
%Type = Misc
\bibitem[{Hua et~al.(2019)Hua, Zhou, Sa, Zhang and Suh}]{huaChannelGatingNeural2019}
\bibinfo{author}{Hua, W.}, \bibinfo{author}{Zhou, Y.}, \bibinfo{author}{Sa, C.D.}, \bibinfo{author}{Zhang, Z.}, \bibinfo{author}{Suh, G.E.}, \bibinfo{year}{2019}.
\newblock \bibinfo{title}{Channel {{Gating Neural Networks}}}.
\newblock \DOIprefix\doi{10.48550/arXiv.1805.12549}, \href{http://arxiv.org/abs/1805.12549}{{\tt arXiv:1805.12549}}.
%Type = Misc
\bibitem[{Huang et~al.(2018)Huang, Chen, Li, Wu, van~der Maaten and Weinberger}]{huangMultiScaleDenseNetworks2018}
\bibinfo{author}{Huang, G.}, \bibinfo{author}{Chen, D.}, \bibinfo{author}{Li, T.}, \bibinfo{author}{Wu, F.}, \bibinfo{author}{van~der Maaten, L.}, \bibinfo{author}{Weinberger, K.Q.}, \bibinfo{year}{2018}.
\newblock \bibinfo{title}{Multi-{{Scale Dense Networks}} for {{Resource Efficient Image Classification}}}.
\newblock \href{http://arxiv.org/abs/1703.09844}{{\tt arXiv:1703.09844}}.
%Type = Article
\bibitem[{Huang et~al.(2025)Huang, Huang, Zuo, Gong, Zhang, Liu and Fang}]{huangPSRegPriorguidedSparse2025}
\bibinfo{author}{Huang, X.}, \bibinfo{author}{Huang, Z.}, \bibinfo{author}{Zuo, Y.}, \bibinfo{author}{Gong, Y.}, \bibinfo{author}{Zhang, C.}, \bibinfo{author}{Liu, D.}, \bibinfo{author}{Fang, Y.}, \bibinfo{year}{2025}.
\newblock \bibinfo{title}{{{PSReg}}: {{Prior-guided Sparse Mixture}} of {{Experts}} for {{Point Cloud Registration}}}.
\newblock \bibinfo{journal}{Proceedings of the AAAI Conference on Artificial Intelligence} \bibinfo{volume}{39}, \bibinfo{pages}{3788--3796}.
\newblock \DOIprefix\doi{10.1609/aaai.v39i4.32395}.
%Type = Article
\bibitem[{Jacobs et~al.(1991)Jacobs, Jordan, Nowlan and Hinton}]{jacobsAdaptiveMixturesLocal1991}
\bibinfo{author}{Jacobs, R.A.}, \bibinfo{author}{Jordan, M.I.}, \bibinfo{author}{Nowlan, S.J.}, \bibinfo{author}{Hinton, G.E.}, \bibinfo{year}{1991}.
\newblock \bibinfo{title}{Adaptive {{Mixtures}} of {{Local Experts}}}.
\newblock \bibinfo{journal}{Neural Computation} \bibinfo{volume}{3}, \bibinfo{pages}{79--87}.
\newblock \DOIprefix\doi{10.1162/neco.1991.3.1.79}.
%Type = Article
\bibitem[{Jain et~al.(2024)Jain, Hegde, Kusupati, Nagrani, Buch, Jain, Arnab and Paul}]{jainMixtureNestedExperts2024}
\bibinfo{author}{Jain, G.}, \bibinfo{author}{Hegde, N.}, \bibinfo{author}{Kusupati, A.}, \bibinfo{author}{Nagrani, A.}, \bibinfo{author}{Buch, S.}, \bibinfo{author}{Jain, P.}, \bibinfo{author}{Arnab, A.}, \bibinfo{author}{Paul, S.}, \bibinfo{year}{2024}.
\newblock \bibinfo{title}{Mixture of {{Nested Experts}}: {{Adaptive Processing}} of {{Visual Tokens}}}.
\newblock \bibinfo{journal}{Advances in Neural Information Processing Systems} \bibinfo{volume}{37}, \bibinfo{pages}{58480--58497}.
%Type = Misc
\bibitem[{Jiang et~al.(2023)Jiang, Peng, Lian, Shao and Xu}]{jiangNeighborPatchesMerging2023a}
\bibinfo{author}{Jiang, K.}, \bibinfo{author}{Peng, P.}, \bibinfo{author}{Lian, Y.}, \bibinfo{author}{Shao, W.}, \bibinfo{author}{Xu, W.}, \bibinfo{year}{2023}.
\newblock \bibinfo{title}{Neighbor {{Patches Merging Reduces Spatial Redundancy}} of {{Nature Images}}}.
\newblock \DOIprefix\doi{10.2139/ssrn.4663091}, \href{http://arxiv.org/abs/4663091}{{\tt arXiv:4663091}}.
%Type = Article
\bibitem[{Jiang et~al.(2020)Jiang, Cheng, Lin and Fu}]{jiangLearningLayerSkippableInference2020}
\bibinfo{author}{Jiang, Y.G.}, \bibinfo{author}{Cheng, C.}, \bibinfo{author}{Lin, H.}, \bibinfo{author}{Fu, Y.}, \bibinfo{year}{2020}.
\newblock \bibinfo{title}{Learning {{Layer-Skippable Inference Network}}}.
\newblock \bibinfo{journal}{IEEE Transactions on Image Processing} \bibinfo{volume}{29}, \bibinfo{pages}{8747--8759}.
\newblock \DOIprefix\doi{10.1109/TIP.2020.3018269}.
%Type = Article
\bibitem[{Jie et~al.(2021)Jie, Sun, Li, Feng and Liu}]{jieAnytimeRecognitionRouting2021}
\bibinfo{author}{Jie, Z.}, \bibinfo{author}{Sun, P.}, \bibinfo{author}{Li, X.}, \bibinfo{author}{Feng, J.}, \bibinfo{author}{Liu, W.}, \bibinfo{year}{2021}.
\newblock \bibinfo{title}{Anytime {{Recognition}} with {{Routing Convolutional Networks}}}.
\newblock \bibinfo{journal}{IEEE Transactions on Pattern Analysis and Machine Intelligence} \bibinfo{volume}{43}, \bibinfo{pages}{1875--1886}.
\newblock \DOIprefix\doi{10.1109/TPAMI.2019.2959322}.
%Type = Article
\bibitem[{John et~al.(2018)John, Boyali, Tehrani, Ishimaru, Konishi, Liu and Mita}]{johnEstimationSteeringAngle2018}
\bibinfo{author}{John, V.}, \bibinfo{author}{Boyali, A.}, \bibinfo{author}{Tehrani, H.}, \bibinfo{author}{Ishimaru, K.}, \bibinfo{author}{Konishi, M.}, \bibinfo{author}{Liu, Z.}, \bibinfo{author}{Mita, S.}, \bibinfo{year}{2018}.
\newblock \bibinfo{title}{Estimation of {{Steering Angle}} and {{Collision Avoidance}} for {{Automated Driving Using Deep Mixture}} of {{Experts}}}.
\newblock \bibinfo{journal}{IEEE Transactions on Intelligent Vehicles} \bibinfo{volume}{3}, \bibinfo{pages}{571--584}.
\newblock \DOIprefix\doi{10.1109/TIV.2018.2874555}.
%Type = Inproceedings
\bibitem[{Ju et~al.(2021)Ju, Bao, Ge and Yuan}]{juDynamicEarlyExit2021}
\bibinfo{author}{Ju, W.}, \bibinfo{author}{Bao, W.}, \bibinfo{author}{Ge, L.}, \bibinfo{author}{Yuan, D.}, \bibinfo{year}{2021}.
\newblock \bibinfo{title}{Dynamic {{Early Exit Scheduling}} for {{Deep Neural Network Inference}} through {{Contextual Bandits}}}, in: \bibinfo{booktitle}{Proceedings of the 30th {{ACM International Conference}} on {{Information}} \& {{Knowledge Management}}}, \bibinfo{publisher}{ACM}, \bibinfo{address}{Virtual Event Queensland Australia}. pp. \bibinfo{pages}{823--832}.
\newblock \DOIprefix\doi{10.1145/3459637.3482335}.
%Type = Misc
\bibitem[{Kaiser and Bengio(2018)}]{kaiserDiscreteAutoencodersSequence2018}
\bibinfo{author}{Kaiser, {\L}.}, \bibinfo{author}{Bengio, S.}, \bibinfo{year}{2018}.
\newblock \bibinfo{title}{Discrete {{Autoencoders}} for {{Sequence Models}}}.
\newblock \DOIprefix\doi{10.48550/arXiv.1801.09797}, \href{http://arxiv.org/abs/1801.09797}{{\tt arXiv:1801.09797}}.
%Type = Article
\bibitem[{Kandel et~al.()Kandel, Koester, Mack and Siegelbaum}]{kandelPrinciplesNeuralScience}
\bibinfo{author}{Kandel, E.R.}, \bibinfo{author}{Koester, J.D.}, \bibinfo{author}{Mack, S.H.}, \bibinfo{author}{Siegelbaum, S.A.}, .
\newblock \bibinfo{title}{Principles of {{Neural Science}}, {{Sixth Edition}}} .
%Type = Inproceedings
\bibitem[{Kaya et~al.(2019)Kaya, Hong and Dumitras}]{kayaShallowDeepNetworksUnderstanding2019}
\bibinfo{author}{Kaya, Y.}, \bibinfo{author}{Hong, S.}, \bibinfo{author}{Dumitras, T.}, \bibinfo{year}{2019}.
\newblock \bibinfo{title}{Shallow-{{Deep Networks}}: {{Understanding}} and {{Mitigating Network Overthinking}}}, in: \bibinfo{booktitle}{Proceedings of the 36th {{International Conference}} on {{Machine Learning}}}, \bibinfo{publisher}{PMLR}. pp. \bibinfo{pages}{3301--3310}.
%Type = Article
\bibitem[{Kim and Lee(2024)}]{kimExitNotExit2024}
\bibinfo{author}{Kim, K.S.}, \bibinfo{author}{Lee, H.S.}, \bibinfo{year}{2024}.
\newblock \bibinfo{title}{To {{Exit}} or {{Not}} to {{Exit}}: {{Cost-Effective Early-Exit Architecture Based}} on {{Markov Decision Process}}}.
\newblock \bibinfo{journal}{Mathematics} \bibinfo{volume}{12}, \bibinfo{pages}{2263}.
\newblock \DOIprefix\doi{10.3390/math12142263}.
%Type = Inproceedings
\bibitem[{Kim et~al.(2024)Kim, Gao, Hsu, Shen and Jin}]{kimTokenFusionBridging2024}
\bibinfo{author}{Kim, M.}, \bibinfo{author}{Gao, S.}, \bibinfo{author}{Hsu, Y.C.}, \bibinfo{author}{Shen, Y.}, \bibinfo{author}{Jin, H.}, \bibinfo{year}{2024}.
\newblock \bibinfo{title}{Token {{Fusion}}: {{Bridging}} the {{Gap}} between {{Token Pruning}} and {{Token Merging}}}, in: \bibinfo{booktitle}{2024 {{IEEE}}/{{CVF Winter Conference}} on {{Applications}} of {{Computer Vision}} ({{WACV}})}, \bibinfo{publisher}{IEEE}, \bibinfo{address}{Waikoloa, HI, USA}. pp. \bibinfo{pages}{1372--1381}.
\newblock \DOIprefix\doi{10.1109/WACV57701.2024.00141}.
%Type = Misc
\bibitem[{Kornblith et~al.(2019)Kornblith, Norouzi, Lee and Hinton}]{kornblithSimilarityNeuralNetwork2019}
\bibinfo{author}{Kornblith, S.}, \bibinfo{author}{Norouzi, M.}, \bibinfo{author}{Lee, H.}, \bibinfo{author}{Hinton, G.}, \bibinfo{year}{2019}.
\newblock \bibinfo{title}{Similarity of {{Neural Network Representations Revisited}}}.
\newblock \href{http://arxiv.org/abs/1905.00414}{{\tt arXiv:1905.00414}}.
%Type = Incollection
\bibitem[{Kouris et~al.(2022)Kouris, Venieris, Laskaridis and Lane}]{kourisMultiExitSemanticSegmentation2022a}
\bibinfo{author}{Kouris, A.}, \bibinfo{author}{Venieris, S.I.}, \bibinfo{author}{Laskaridis, S.}, \bibinfo{author}{Lane, N.}, \bibinfo{year}{2022}.
\newblock \bibinfo{title}{Multi-{{Exit Semantic Segmentation Networks}}}, in: \bibinfo{editor}{Avidan, S.}, \bibinfo{editor}{Brostow, G.}, \bibinfo{editor}{Ciss{\'e}, M.}, \bibinfo{editor}{Farinella, G.M.}, \bibinfo{editor}{Hassner, T.} (Eds.), \bibinfo{booktitle}{Computer {{Vision}} -- {{ECCV}} 2022}. \bibinfo{publisher}{Springer Nature Switzerland}, \bibinfo{address}{Cham}. volume \bibinfo{volume}{13681}, pp. \bibinfo{pages}{330--349}.
\newblock \DOIprefix\doi{10.1007/978-3-031-19803-8_20}.
%Type = Inproceedings
\bibitem[{Krizhevsky et~al.(2012)Krizhevsky, Sutskever and Hinton}]{krizhevskyImageNetClassificationDeep2012}
\bibinfo{author}{Krizhevsky, A.}, \bibinfo{author}{Sutskever, I.}, \bibinfo{author}{Hinton, G.E.}, \bibinfo{year}{2012}.
\newblock \bibinfo{title}{{{ImageNet Classification}} with {{Deep Convolutional Neural Networks}}}, in: \bibinfo{booktitle}{Advances in {{Neural Information Processing Systems}}}, \bibinfo{publisher}{Curran Associates, Inc.}
%Type = Misc
\bibitem[{Kuhse et~al.(2025)Kuhse, Teper, Buschj{\"a}ger, Wang and Chen}]{kuhseYouOnlyLook2025}
\bibinfo{author}{Kuhse, D.}, \bibinfo{author}{Teper, H.}, \bibinfo{author}{Buschj{\"a}ger, S.}, \bibinfo{author}{Wang, C.Y.}, \bibinfo{author}{Chen, J.J.}, \bibinfo{year}{2025}.
\newblock \bibinfo{title}{You {{Only Look Once}} at {{Anytime}} ({{AnytimeYOLO}}): {{Analysis}} and {{Optimization}} of {{Early-Exits}} for {{Object-Detection}}}.
\newblock \DOIprefix\doi{10.48550/arXiv.2503.17497}, \href{http://arxiv.org/abs/2503.17497}{{\tt arXiv:2503.17497}}.
%Type = Inproceedings
\bibitem[{Langford and Zhang(2007)}]{langfordEpochGreedyAlgorithmMultiarmed2007}
\bibinfo{author}{Langford, J.}, \bibinfo{author}{Zhang, T.}, \bibinfo{year}{2007}.
\newblock \bibinfo{title}{The {{Epoch-Greedy Algorithm}} for {{Multi-armed Bandits}} with {{Side Information}}}, in: \bibinfo{booktitle}{Advances in {{Neural Information Processing Systems}}}, \bibinfo{publisher}{Curran Associates, Inc.}
%Type = Inproceedings
\bibitem[{Laskaridis et~al.(2020a)Laskaridis, Venieris, Almeida, Leontiadis and Lane}]{laskaridisSPINNSynergisticProgressive2020}
\bibinfo{author}{Laskaridis, S.}, \bibinfo{author}{Venieris, S.I.}, \bibinfo{author}{Almeida, M.}, \bibinfo{author}{Leontiadis, I.}, \bibinfo{author}{Lane, N.D.}, \bibinfo{year}{2020}a.
\newblock \bibinfo{title}{{{SPINN}}: Synergistic progressive inference of neural networks over device and cloud}, in: \bibinfo{booktitle}{Proceedings of the 26th {{Annual International Conference}} on {{Mobile Computing}} and {{Networking}}}, \bibinfo{publisher}{ACM}, \bibinfo{address}{London United Kingdom}. pp. \bibinfo{pages}{1--15}.
\newblock \DOIprefix\doi{10.1145/3372224.3419194}.
%Type = Inproceedings
\bibitem[{Laskaridis et~al.(2020b)Laskaridis, Venieris, Kim and Lane}]{laskaridisHAPIHardwareAwareProgressive2020}
\bibinfo{author}{Laskaridis, S.}, \bibinfo{author}{Venieris, S.I.}, \bibinfo{author}{Kim, H.}, \bibinfo{author}{Lane, N.D.}, \bibinfo{year}{2020}b.
\newblock \bibinfo{title}{{{HAPI}}: {{Hardware-Aware Progressive Inference}}}, in: \bibinfo{booktitle}{2020 {{IEEE}}/{{ACM International Conference On Computer Aided Design}} ({{ICCAD}})}, pp. \bibinfo{pages}{1--9}.
%Type = Book
\bibitem[{Lei et~al.()Lei, Bai, Brahma, Ainslie, Lee, Zhou, Du, Zhao, Wu, Li, Zhang and Google}]{leiConditionalAdaptersParameterefficient2023a}
\bibinfo{author}{Lei, T.}, \bibinfo{author}{Bai, J.}, \bibinfo{author}{Brahma, S.}, \bibinfo{author}{Ainslie, J.}, \bibinfo{author}{Lee, K.}, \bibinfo{author}{Zhou, Y.}, \bibinfo{author}{Du, N.}, \bibinfo{author}{Zhao, V.}, \bibinfo{author}{Wu, Y.}, \bibinfo{author}{Li, B.}, \bibinfo{author}{Zhang, Y.}, \bibinfo{author}{Google, M.W.}, .
\newblock \bibinfo{title}{Conditional Adapters: Parameter-efficient Transfer Learning with Fast Inference}.
\newblock \URLprefix \url{https://proceedings.neurips.cc/paper_files/paper/2023/file/19d7204af519eae9993f7f72377a0ec0-Paper-Conference.pdf}.
%Type = Article
\bibitem[{Li et~al.(2020a)Li, Zeng, Zhou and Chen}]{liEdgeAIOnDemand2020}
\bibinfo{author}{Li, E.}, \bibinfo{author}{Zeng, L.}, \bibinfo{author}{Zhou, Z.}, \bibinfo{author}{Chen, X.}, \bibinfo{year}{2020}a.
\newblock \bibinfo{title}{Edge {{AI}}: {{On-Demand Accelerating Deep Neural Network Inference}} via {{Edge Computing}}}.
\newblock \bibinfo{journal}{IEEE Transactions on Wireless Communications} \bibinfo{volume}{19}, \bibinfo{pages}{447--457}.
\newblock \DOIprefix\doi{10.1109/TWC.2019.2946140}.
%Type = Article
\bibitem[{Li et~al.(2021)Li, Li, He and Cheng}]{liDynamicDualGating2021}
\bibinfo{author}{Li, F.}, \bibinfo{author}{Li, G.}, \bibinfo{author}{He, X.}, \bibinfo{author}{Cheng, J.}, \bibinfo{year}{2021}.
\newblock \bibinfo{title}{Dynamic dual gating neural networks}.
\newblock \bibinfo{journal}{2021 IEEE/CVF International Conference on Computer Vision (ICCV)} , \bibinfo{pages}{5310–5319}\URLprefix \url{https://ieeexplore.ieee.org/document/9710448}, \DOIprefix\doi{https://doi.org/10.1109/iccv48922.2021.00528}.
%Type = Inproceedings
\bibitem[{Li et~al.(2019)Li, Zhang, Qi, Ruigang and Huang}]{liImprovedTechniquesTraining2019}
\bibinfo{author}{Li, H.}, \bibinfo{author}{Zhang, H.}, \bibinfo{author}{Qi, X.}, \bibinfo{author}{Ruigang, Y.}, \bibinfo{author}{Huang, G.}, \bibinfo{year}{2019}.
\newblock \bibinfo{title}{Improved {{Techniques}} for {{Training Adaptive Deep Networks}}}, in: \bibinfo{booktitle}{2019 {{IEEE}}/{{CVF International Conference}} on {{Computer Vision}} ({{ICCV}})}, pp. \bibinfo{pages}{1891--1900}.
\newblock \DOIprefix\doi{10.1109/ICCV.2019.00198}.
%Type = Misc
\bibitem[{Li et~al.(2022)Li, Thorsley and Hassoun}]{liSaiTSparseVision2022}
\bibinfo{author}{Li, L.}, \bibinfo{author}{Thorsley, D.}, \bibinfo{author}{Hassoun, J.}, \bibinfo{year}{2022}.
\newblock \bibinfo{title}{{{SaiT}}: {{Sparse Vision Transformers}} through {{Adaptive Token Pruning}}}.
\newblock \DOIprefix\doi{10.48550/arXiv.2210.05832}, \href{http://arxiv.org/abs/2210.05832}{{\tt arXiv:2210.05832}}.
%Type = Inproceedings
\bibitem[{Li et~al.(2023a)Li, Lou, Chen, Zhu, Shen, Ma and Zou}]{liPredictiveExitPrediction2023a}
\bibinfo{author}{Li, X.}, \bibinfo{author}{Lou, C.}, \bibinfo{author}{Chen, Y.}, \bibinfo{author}{Zhu, Z.}, \bibinfo{author}{Shen, Y.}, \bibinfo{author}{Ma, Y.}, \bibinfo{author}{Zou, A.}, \bibinfo{year}{2023}a.
\newblock \bibinfo{title}{Predictive exit: Prediction of fine-grained early exits for computation-and energy-efficient inference}, in: \bibinfo{booktitle}{Proceedings of the {{Thirty-Seventh AAAI Conference}} on {{Artificial Intelligence}} and {{Thirty-Fifth Conference}} on {{Innovative Applications}} of {{Artificial Intelligence}} and {{Thirteenth Symposium}} on {{Educational Advances}} in {{Artificial Intelligence}}}, \bibinfo{publisher}{AAAI Press}. pp. \bibinfo{pages}{8657--8665}.
\newblock \DOIprefix\doi{10.1609/aaai.v37i7.26042}.
%Type = Misc
\bibitem[{Li et~al.(2023b)Li, Geller, Kim and Panda}]{liSEENNTemporalSpiking2023a}
\bibinfo{author}{Li, Y.}, \bibinfo{author}{Geller, T.}, \bibinfo{author}{Kim, Y.}, \bibinfo{author}{Panda, P.}, \bibinfo{year}{2023}b.
\newblock \bibinfo{title}{{{SEENN}}: {{Towards Temporal Spiking Early-Exit Neural Networks}}}.
\newblock \href{http://arxiv.org/abs/2304.01230}{{\tt arXiv:2304.01230}}.
%Type = Inproceedings
\bibitem[{Li et~al.(2020b)Li, Song, Chen, Li, Zhang, Wang and Sun}]{liLearningDynamicRouting2020}
\bibinfo{author}{Li, Y.}, \bibinfo{author}{Song, L.}, \bibinfo{author}{Chen, Y.}, \bibinfo{author}{Li, Z.}, \bibinfo{author}{Zhang, X.}, \bibinfo{author}{Wang, X.}, \bibinfo{author}{Sun, J.}, \bibinfo{year}{2020}b.
\newblock \bibinfo{title}{Learning {{Dynamic Routing}} for {{Semantic Segmentation}}}, in: \bibinfo{booktitle}{2020 {{IEEE}}/{{CVF Conference}} on {{Computer Vision}} and {{Pattern Recognition}} ({{CVPR}})}, \bibinfo{publisher}{IEEE}, \bibinfo{address}{Seattle, WA, USA}. pp. \bibinfo{pages}{8550--8559}.
\newblock \DOIprefix\doi{10.1109/CVPR42600.2020.00858}.
%Type = Article
\bibitem[{Liang et~al.(2022a)Liang, Fan, Sarkar, Jiang, Chen, Zou, Cheng, Hao and Wang}]{liangM3ViTMixtureofExpertsVision2022}
\bibinfo{author}{Liang, H.}, \bibinfo{author}{Fan, Z.}, \bibinfo{author}{Sarkar, R.}, \bibinfo{author}{Jiang, Z.}, \bibinfo{author}{Chen, T.}, \bibinfo{author}{Zou, K.}, \bibinfo{author}{Cheng, Y.}, \bibinfo{author}{Hao, C.}, \bibinfo{author}{Wang, Z.}, \bibinfo{year}{2022}a.
\newblock \bibinfo{title}{{{M3ViT}}: {{Mixture-of-Experts Vision Transformer}} for {{Efficient Multi-task Learning}} with {{Model-Accelerator Co-design}}} .
%Type = Misc
\bibitem[{Liang et~al.(2022b)Liang, Ge, Tong, Song, Wang and Xie}]{liangNotAllPatches2022}
\bibinfo{author}{Liang, Y.}, \bibinfo{author}{Ge, C.}, \bibinfo{author}{Tong, Z.}, \bibinfo{author}{Song, Y.}, \bibinfo{author}{Wang, J.}, \bibinfo{author}{Xie, P.}, \bibinfo{year}{2022}b.
\newblock \bibinfo{title}{Not {{All Patches}} are {{What You Need}}: {{Expediting Vision Transformers}} via {{Token Reorganizations}}}.
\newblock \href{http://arxiv.org/abs/2202.07800}{{\tt arXiv:2202.07800}}.
%Type = Inproceedings
\bibitem[{Lin et~al.(2017)Lin, Rao, Lu and Zhou}]{linRuntimeNeuralPruning2017}
\bibinfo{author}{Lin, J.}, \bibinfo{author}{Rao, Y.}, \bibinfo{author}{Lu, J.}, \bibinfo{author}{Zhou, J.}, \bibinfo{year}{2017}.
\newblock \bibinfo{title}{Runtime {{Neural Pruning}}}, in: \bibinfo{booktitle}{Advances in {{Neural Information Processing Systems}}}, \bibinfo{publisher}{Curran Associates, Inc.}
%Type = Article
\bibitem[{Lin et~al.(2024)Lin, Ji, Ji and Yao}]{linCloserLookBranch2024}
\bibinfo{author}{Lin, S.}, \bibinfo{author}{Ji, B.}, \bibinfo{author}{Ji, R.}, \bibinfo{author}{Yao, A.}, \bibinfo{year}{2024}.
\newblock \bibinfo{title}{A closer look at branch classifiers of multi-exit architectures}.
\newblock \bibinfo{journal}{Computer Vision and Image Understanding} \bibinfo{volume}{239}, \bibinfo{pages}{103900}.
\newblock \DOIprefix\doi{10.1016/j.cviu.2023.103900}.
%Type = Article
\bibitem[{Liu et~al.(2022)Liu, Parajuli, Hostetler, Chai and Bhanu}]{liuDynamicallyThrottleableNeural2022}
\bibinfo{author}{Liu, H.}, \bibinfo{author}{Parajuli, S.}, \bibinfo{author}{Hostetler, J.}, \bibinfo{author}{Chai, S.}, \bibinfo{author}{Bhanu, B.}, \bibinfo{year}{2022}.
\newblock \bibinfo{title}{Dynamically throttleable neural networks}.
\newblock \bibinfo{journal}{Machine Vision and Applications} \bibinfo{volume}{33}, \bibinfo{pages}{59}.
\newblock \DOIprefix\doi{10.1007/s00138-022-01311-z}.
%Type = Article
\bibitem[{Liu and Moon(2023)}]{liuSelfsupervisedEfficientSample2023a}
\bibinfo{author}{Liu, K.}, \bibinfo{author}{Moon, S.}, \bibinfo{year}{2023}.
\newblock \bibinfo{title}{Self-supervised efficient sample weighting for multi-exit networks}.
\newblock \bibinfo{journal}{Knowledge-Based Systems} \bibinfo{volume}{280}, \bibinfo{pages}{111003}.
\newblock \DOIprefix\doi{10.1016/j.knosys.2023.111003}.
%Type = Article
\bibitem[{Liu and Deng(2018)}]{liuDynamicDeepNeural2018}
\bibinfo{author}{Liu, L.}, \bibinfo{author}{Deng, J.}, \bibinfo{year}{2018}.
\newblock \bibinfo{title}{Dynamic {{Deep Neural Networks}}: {{Optimizing Accuracy-Efficiency Trade-Offs}} by {{Selective Execution}}}.
\newblock \bibinfo{journal}{Proceedings of the AAAI Conference on Artificial Intelligence} \bibinfo{volume}{32}.
\newblock \DOIprefix\doi{10.1609/aaai.v32i1.11630}.
%Type = Misc
\bibitem[{Liu et~al.(2025)Liu, Su, Jiang, Wu, Guo, Xu and Yang}]{liuAccurateEfficient3D2025}
\bibinfo{author}{Liu, L.}, \bibinfo{author}{Su, B.}, \bibinfo{author}{Jiang, J.}, \bibinfo{author}{Wu, G.}, \bibinfo{author}{Guo, C.}, \bibinfo{author}{Xu, C.}, \bibinfo{author}{Yang, H.F.}, \bibinfo{year}{2025}.
\newblock \bibinfo{title}{Towards {{Accurate}} and {{Efficient 3D Object Detection}} for {{Autonomous Driving}}: {{A Mixture}} of {{Experts Computing System}} on {{Edge}}}.
\newblock \DOIprefix\doi{10.48550/arXiv.2507.04123}, \href{http://arxiv.org/abs/2507.04123}{{\tt arXiv:2507.04123}}.
%Type = Misc
\bibitem[{Liu et~al.(2024a)Liu, Blondel, Riquelme and Puigcerver}]{liuRoutersVisionMixture2024}
\bibinfo{author}{Liu, T.}, \bibinfo{author}{Blondel, M.}, \bibinfo{author}{Riquelme, C.}, \bibinfo{author}{Puigcerver, J.}, \bibinfo{year}{2024}a.
\newblock \bibinfo{title}{Routers in {{Vision Mixture}} of {{Experts}}: {{An Empirical Study}}}.
\newblock \href{http://arxiv.org/abs/2401.15969}{{\tt arXiv:2401.15969}}.
%Type = Inproceedings
\bibitem[{Liu et~al.(2024b)Liu, Gehrig, Messikommer, Cannici and Scaramuzza}]{liuRevisitingTokenPruning2024a}
\bibinfo{author}{Liu, Y.}, \bibinfo{author}{Gehrig, M.}, \bibinfo{author}{Messikommer, N.}, \bibinfo{author}{Cannici, M.}, \bibinfo{author}{Scaramuzza, D.}, \bibinfo{year}{2024}b.
\newblock \bibinfo{title}{Revisiting {{Token Pruning}} for {{Object Detection}} and {{Instance Segmentation}}}, in: \bibinfo{booktitle}{2024 {{IEEE}}/{{CVF Winter Conference}} on {{Applications}} of {{Computer Vision}} ({{WACV}})}, \bibinfo{publisher}{IEEE}, \bibinfo{address}{Waikoloa, HI, USA}. pp. \bibinfo{pages}{2646--2656}.
\newblock \DOIprefix\doi{10.1109/WACV57701.2024.00264}.
%Type = Inproceedings
\bibitem[{Liu et~al.(2021a)Liu, Lin, Cao, Hu, Wei, Zhang, Lin and Guo}]{liuSwinTransformerHierarchical2021}
\bibinfo{author}{Liu, Z.}, \bibinfo{author}{Lin, Y.}, \bibinfo{author}{Cao, Y.}, \bibinfo{author}{Hu, H.}, \bibinfo{author}{Wei, Y.}, \bibinfo{author}{Zhang, Z.}, \bibinfo{author}{Lin, S.}, \bibinfo{author}{Guo, B.}, \bibinfo{year}{2021}a.
\newblock \bibinfo{title}{Swin {{Transformer}}: {{Hierarchical Vision Transformer}} using {{Shifted Windows}}}, in: \bibinfo{booktitle}{2021 {{IEEE}}/{{CVF International Conference}} on {{Computer Vision}} ({{ICCV}})}, \bibinfo{publisher}{IEEE}, \bibinfo{address}{Montreal, QC, Canada}. pp. \bibinfo{pages}{9992--10002}.
\newblock \DOIprefix\doi{10.1109/ICCV48922.2021.00986}.
%Type = Misc
\bibitem[{Liu et~al.(2021b)Liu, Xu, Wang, Darrell and Shelhamer}]{liuAnytimeDensePrediction2021}
\bibinfo{author}{Liu, Z.}, \bibinfo{author}{Xu, Z.}, \bibinfo{author}{Wang, H.J.}, \bibinfo{author}{Darrell, T.}, \bibinfo{author}{Shelhamer, E.}, \bibinfo{year}{2021}b.
\newblock \bibinfo{title}{Anytime {{Dense Prediction}} with {{Confidence Adaptivity}}}.
\newblock \bibinfo{howpublished}{https://arxiv.org/abs/2104.00749v2}.
%Type = Misc
\bibitem[{Liu et~al.(2024c)Liu, Zhu, Li and Huang}]{liuMultipleExitTuningInferenceEfficient2024}
\bibinfo{author}{Liu, Z.}, \bibinfo{author}{Zhu, J.}, \bibinfo{author}{Li, N.}, \bibinfo{author}{Huang, G.}, \bibinfo{year}{2024}c.
\newblock \bibinfo{title}{Multiple-{{Exit Tuning}}: {{Towards Inference-Efficient Adaptation}} for {{Vision Transformer}}}.
\newblock \href{http://arxiv.org/abs/2409.13999}{{\tt arXiv:2409.13999}}.
%Type = Inproceedings
\bibitem[{Long et~al.(2023)Long, Zhao, Pi, Wang and Wang}]{longAttentiveTokensIncorporating2023}
\bibinfo{author}{Long, S.}, \bibinfo{author}{Zhao, Z.}, \bibinfo{author}{Pi, J.}, \bibinfo{author}{Wang, S.}, \bibinfo{author}{Wang, J.}, \bibinfo{year}{2023}.
\newblock \bibinfo{title}{Beyond {{Attentive Tokens}}: {{Incorporating Token Importance}} and {{Diversity}} for {{Efficient Vision Transformers}}}, in: \bibinfo{booktitle}{2023 {{IEEE}}/{{CVF Conference}} on {{Computer Vision}} and {{Pattern Recognition}} ({{CVPR}})}, \bibinfo{publisher}{IEEE}, \bibinfo{address}{Vancouver, BC, Canada}. pp. \bibinfo{pages}{10334--10343}.
\newblock \DOIprefix\doi{10.1109/CVPR52729.2023.00996}.
%Type = Inproceedings
\bibitem[{Lu et~al.(2023)Lu, De~Geus and Dubbelman}]{luContentawareTokenSharing2023a}
\bibinfo{author}{Lu, C.}, \bibinfo{author}{De~Geus, D.}, \bibinfo{author}{Dubbelman, G.}, \bibinfo{year}{2023}.
\newblock \bibinfo{title}{Content-aware {{Token Sharing}} for {{Efficient Semantic Segmentation}} with {{Vision Transformers}}}, in: \bibinfo{booktitle}{2023 {{IEEE}}/{{CVF Conference}} on {{Computer Vision}} and {{Pattern Recognition}} ({{CVPR}})}, \bibinfo{publisher}{IEEE}, \bibinfo{address}{Vancouver, BC, Canada}. pp. \bibinfo{pages}{23631--23640}.
\newblock \DOIprefix\doi{10.1109/CVPR52729.2023.02263}.
%Type = Inproceedings
\bibitem[{Lu et~al.(2025)Lu, Zheng, Xia and Wang}]{luToMATokenMerge2025}
\bibinfo{author}{Lu, W.}, \bibinfo{author}{Zheng, S.}, \bibinfo{author}{Xia, Y.}, \bibinfo{author}{Wang, S.}, \bibinfo{year}{2025}.
\newblock \bibinfo{title}{{{ToMA}}: {{Token Merge}} with {{Attention}} for {{Diffusion Models}}}, in: \bibinfo{booktitle}{Forty-Second {{International Conference}} on {{Machine Learning}}}.
%Type = Inproceedings
\bibitem[{Malawade et~al.(2022a)Malawade, Mortlock and Al~Faruque}]{malawadeHydraFusionContextAwareSelective2022}
\bibinfo{author}{Malawade, A.V.}, \bibinfo{author}{Mortlock, T.}, \bibinfo{author}{Al~Faruque, M.A.}, \bibinfo{year}{2022}a.
\newblock \bibinfo{title}{{{HydraFusion}}: {{Context-Aware Selective Sensor Fusion}} for {{Robust}} and {{Efficient Autonomous Vehicle Perception}}}, in: \bibinfo{booktitle}{2022 {{ACM}}/{{IEEE}} 13th {{International Conference}} on {{Cyber-Physical Systems}} ({{ICCPS}})}, pp. \bibinfo{pages}{68--79}.
\newblock \DOIprefix\doi{10.1109/ICCPS54341.2022.00013}.
%Type = Inproceedings
\bibitem[{Malawade et~al.(2022b)Malawade, Mortlock and Faruque}]{malawadeEcoFusionEnergyawareAdaptive2022}
\bibinfo{author}{Malawade, A.V.}, \bibinfo{author}{Mortlock, T.}, \bibinfo{author}{Faruque, M.A.A.}, \bibinfo{year}{2022}b.
\newblock \bibinfo{title}{{{EcoFusion}}: Energy-aware adaptive sensor fusion for efficient autonomous vehicle perception}, in: \bibinfo{booktitle}{Proceedings of the 59th {{ACM}}/{{IEEE Design Automation Conference}}}, \bibinfo{publisher}{Association for Computing Machinery}, \bibinfo{address}{New York, NY, USA}. pp. \bibinfo{pages}{481--486}.
\newblock \DOIprefix\doi{10.1145/3489517.3530489}.
%Type = Article
\bibitem[{Matsubara et~al.(2022)Matsubara, Levorato and Restuccia}]{matsubaraSplitComputingEarly2022a}
\bibinfo{author}{Matsubara, Y.}, \bibinfo{author}{Levorato, M.}, \bibinfo{author}{Restuccia, F.}, \bibinfo{year}{2022}.
\newblock \bibinfo{title}{Split {{Computing}} and {{Early Exiting}} for {{Deep Learning Applications}}: {{Survey}} and {{Research Challenges}}}.
\newblock \bibinfo{journal}{ACM Comput. Surv.} \bibinfo{volume}{55}, \bibinfo{pages}{90:1--90:30}.
\newblock \DOIprefix\doi{10.1145/3527155}.
%Type = Inproceedings
\bibitem[{Mees et~al.(2016)Mees, Eitel and Burgard}]{meesChoosingSmartlyAdaptive2016b}
\bibinfo{author}{Mees, O.}, \bibinfo{author}{Eitel, A.}, \bibinfo{author}{Burgard, W.}, \bibinfo{year}{2016}.
\newblock \bibinfo{title}{Choosing smartly: {{Adaptive}} multimodal fusion for object detection in changing environments}, in: \bibinfo{booktitle}{2016 {{IEEE}}/{{RSJ International Conference}} on {{Intelligent Robots}} and {{Systems}} ({{IROS}})}, pp. \bibinfo{pages}{151--156}.
\newblock \DOIprefix\doi{10.1109/IROS.2016.7759048}.
%Type = Inproceedings
\bibitem[{Meng et~al.(2022)Meng, Li, Chen, Lan, Wu, Jiang and Lim}]{mengAdaViTAdaptiveVision2022}
\bibinfo{author}{Meng, L.}, \bibinfo{author}{Li, H.}, \bibinfo{author}{Chen, B.C.}, \bibinfo{author}{Lan, S.}, \bibinfo{author}{Wu, Z.}, \bibinfo{author}{Jiang, Y.G.}, \bibinfo{author}{Lim, S.N.}, \bibinfo{year}{2022}.
\newblock \bibinfo{title}{{{AdaViT}}: {{Adaptive Vision Transformers}} for {{Efficient Image Recognition}}}, in: \bibinfo{booktitle}{2022 {{IEEE}}/{{CVF Conference}} on {{Computer Vision}} and {{Pattern Recognition}} ({{CVPR}})}, \bibinfo{publisher}{IEEE}, \bibinfo{address}{New Orleans, LA, USA}. pp. \bibinfo{pages}{12299--12308}.
\newblock \DOIprefix\doi{10.1109/CVPR52688.2022.01199}.
%Type = Inproceedings
\bibitem[{Meronen et~al.(2024)Meronen, Trapp, Pilzer, Yang and Solin}]{meronenFixingOverconfidenceDynamic2024a}
\bibinfo{author}{Meronen, L.}, \bibinfo{author}{Trapp, M.}, \bibinfo{author}{Pilzer, A.}, \bibinfo{author}{Yang, L.}, \bibinfo{author}{Solin, A.}, \bibinfo{year}{2024}.
\newblock \bibinfo{title}{Fixing {{Overconfidence}} in {{Dynamic Neural Networks}}}, in: \bibinfo{booktitle}{2024 {{IEEE}}/{{CVF Winter Conference}} on {{Applications}} of {{Computer Vision}} ({{WACV}})}, pp. \bibinfo{pages}{2668--2678}.
\newblock \DOIprefix\doi{10.1109/WACV57701.2024.00266}.
%Type = Misc
\bibitem[{Mersha(2022)}]{mershaDynamicTransformerNetworks2022}
\bibinfo{author}{Mersha, A.}, \bibinfo{year}{2022}.
\newblock \bibinfo{title}{Dynamic {{Transformer Networks}}}.
\newblock \URLprefix \url{https://dynn-icml2022.github.io/papers/paper_19.pdf}.
%Type = Article
\bibitem[{Morra et~al.(2023)Morra, Biondo, Poerio and Lamberti}]{morraMIXOMixtureExpertsBased2023}
\bibinfo{author}{Morra, L.}, \bibinfo{author}{Biondo, A.}, \bibinfo{author}{Poerio, N.}, \bibinfo{author}{Lamberti, F.}, \bibinfo{year}{2023}.
\newblock \bibinfo{title}{{{MIXO}}: {{Mixture Of Experts-Based Visual Odometry}} for {{Multicamera Autonomous Systems}}}.
\newblock \bibinfo{journal}{IEEE Transactions on Consumer Electronics} \bibinfo{volume}{69}, \bibinfo{pages}{261--270}.
\newblock \DOIprefix\doi{10.1109/TCE.2023.3238655}.
%Type = Article
\bibitem[{Mortlock and Al~Faruque(2024)}]{mortlockAdaptiveDataFusion2024}
\bibinfo{author}{Mortlock, T.}, \bibinfo{author}{Al~Faruque, M.A.}, \bibinfo{year}{2024}.
\newblock \bibinfo{title}{Adaptive {{Data Fusion}} for {{State Estimation}} and {{Control}} of {{Power Grids Under Attack}}}.
\newblock \bibinfo{journal}{IEEE Transactions on Industrial Informatics} \bibinfo{volume}{20}, \bibinfo{pages}{11115--11126}.
\newblock \DOIprefix\doi{10.1109/TII.2024.3399885}.
%Type = Book
\bibitem[{Odena et~al.(2017)Odena, Lawson and Olah}]{odenaChangingModelBehavior2017}
\bibinfo{author}{Odena, A.}, \bibinfo{author}{Lawson, D.}, \bibinfo{author}{Olah, C.}, \bibinfo{year}{2017}.
\newblock \bibinfo{title}{Workshop track -ICLR 2017 Changing {{Model Behavior}} at {{Test-Time Using Reinforcement Learning}}}.
\newblock \URLprefix \url{https://arxiv.org/pdf/1702.07780}.
%Type = Misc
\bibitem[{O'Shea and Nash(2015)}]{osheaIntroductionConvolutionalNeural2015}
\bibinfo{author}{O'Shea, K.}, \bibinfo{author}{Nash, R.}, \bibinfo{year}{2015}.
\newblock \bibinfo{title}{An {{Introduction}} to {{Convolutional Neural Networks}}}.
\newblock \DOIprefix\doi{10.48550/arXiv.1511.08458}, \href{http://arxiv.org/abs/1511.08458}{{\tt arXiv:1511.08458}}.
%Type = Article
\bibitem[{Pan et~al.(2021)Pan, Panda, Jiang, Wang, Feris and Oliva}]{panIARED2InterpretabilityAwareRedundancy2021}
\bibinfo{author}{Pan, B.}, \bibinfo{author}{Panda, R.}, \bibinfo{author}{Jiang, Y.}, \bibinfo{author}{Wang, Z.}, \bibinfo{author}{Feris, R.}, \bibinfo{author}{Oliva, A.}, \bibinfo{year}{2021}.
\newblock \bibinfo{title}{{{IA-RED2}}: {{Interpretability-Aware Redundancy Reduction}} for {{Vision Transformers}} ({{Supplementary Material}})} \URLprefix \url{https://proceedings.neurips.cc/paper/2021/file/d072677d210ac4c03ba046120f0802ec-Paper.pdf}.
%Type = Article
\bibitem[{Pan et~al.(2023)Pan, Foo, Zheng, Fan, Rahmani, Ke and Liu}]{panGradMDMAdversarialAttack2023a}
\bibinfo{author}{Pan, J.}, \bibinfo{author}{Foo, L.G.}, \bibinfo{author}{Zheng, Q.}, \bibinfo{author}{Fan, Z.}, \bibinfo{author}{Rahmani, H.}, \bibinfo{author}{Ke, Q.}, \bibinfo{author}{Liu, J.}, \bibinfo{year}{2023}.
\newblock \bibinfo{title}{{{GradMDM}}: {{Adversarial Attack}} on {{Dynamic Networks}}}.
\newblock \bibinfo{journal}{IEEE Transactions on Pattern Analysis and Machine Intelligence} \bibinfo{volume}{45}, \bibinfo{pages}{11374--11381}.
\newblock \DOIprefix\doi{10.1109/TPAMI.2023.3263619}.
%Type = Inproceedings
\bibitem[{Panda et~al.(2016)Panda, Sengupta and Roy}]{pandaConditionalDeepLearning2016}
\bibinfo{author}{Panda, P.}, \bibinfo{author}{Sengupta, A.}, \bibinfo{author}{Roy, K.}, \bibinfo{year}{2016}.
\newblock \bibinfo{title}{Conditional {{Deep Learning}} for {{Energy-Efficient}} and {{Enhanced Pattern Recognition}}}, in: \bibinfo{booktitle}{Proceedings of the 2016 {{Design}}, {{Automation}} \& {{Test}} in {{Europe Conference}} \& {{Exhibition}} ({{DATE}})}, \bibinfo{publisher}{Research Publishing Services}. pp. \bibinfo{pages}{475--480}.
\newblock \DOIprefix\doi{10.3850/9783981537079_0819}.
%Type = Inproceedings
\bibitem[{Park et~al.(2021)Park, Yoo, Jeong, Venkatesh and Kwak}]{parkLearningDynamicNetwork2021a}
\bibinfo{author}{Park, H.}, \bibinfo{author}{Yoo, J.}, \bibinfo{author}{Jeong, S.}, \bibinfo{author}{Venkatesh, G.}, \bibinfo{author}{Kwak, N.}, \bibinfo{year}{2021}.
\newblock \bibinfo{title}{Learning {{Dynamic Network Using}} a {{Reuse Gate Function}} in {{Semi-supervised Video Object Segmentation}}}, in: \bibinfo{booktitle}{2021 {{IEEE}}/{{CVF Conference}} on {{Computer Vision}} and {{Pattern Recognition}} ({{CVPR}})}, \bibinfo{publisher}{IEEE}, \bibinfo{address}{Nashville, TN, USA}. pp. \bibinfo{pages}{8401--8410}.
\newblock \DOIprefix\doi{10.1109/CVPR46437.2021.00830}.
%Type = Inproceedings
\bibitem[{Passalis et~al.(2019)Passalis, Raitoharju, Tefas and Gabbouj}]{passalisAdaptiveInferenceUsing2019}
\bibinfo{author}{Passalis, N.}, \bibinfo{author}{Raitoharju, J.}, \bibinfo{author}{Tefas, A.}, \bibinfo{author}{Gabbouj, M.}, \bibinfo{year}{2019}.
\newblock \bibinfo{title}{Adaptive {{Inference Using Hierarchical Convolutional Bag-of-Features}} for {{Low-Power Embedded Platforms}}}, in: \bibinfo{booktitle}{2019 {{IEEE International Conference}} on {{Image Processing}} ({{ICIP}})}, pp. \bibinfo{pages}{3048--3052}.
\newblock \DOIprefix\doi{10.1109/ICIP.2019.8803283}.
%Type = Article
\bibitem[{Passalis et~al.(2020)Passalis, Raitoharju, Tefas and Gabbouj}]{passalisEfficientAdaptiveInference2020}
\bibinfo{author}{Passalis, N.}, \bibinfo{author}{Raitoharju, J.}, \bibinfo{author}{Tefas, A.}, \bibinfo{author}{Gabbouj, M.}, \bibinfo{year}{2020}.
\newblock \bibinfo{title}{Efficient adaptive inference for deep convolutional neural networks using hierarchical early exits}.
\newblock \bibinfo{journal}{Pattern Recognition} \bibinfo{volume}{105}, \bibinfo{pages}{107346}.
\newblock \DOIprefix\doi{10.1016/j.patcog.2020.107346}.
%Type = Inproceedings
\bibitem[{Passalis and Tefas(2021)}]{passalisAdaptiveInferenceFace2021a}
\bibinfo{author}{Passalis, N.}, \bibinfo{author}{Tefas, A.}, \bibinfo{year}{2021}.
\newblock \bibinfo{title}{Adaptive {{Inference}} for {{Face Recognition}} leveraging {{Deep Metric Learning-enabled Early Exits}}}, in: \bibinfo{booktitle}{2021 29th {{European Signal Processing Conference}} ({{EUSIPCO}})}, pp. \bibinfo{pages}{1346--1350}.
\newblock \DOIprefix\doi{10.23919/EUSIPCO54536.2021.9616231}.
%Type = Inproceedings
\bibitem[{Pavlitskaya et~al.(2020)Pavlitskaya, Hubschneider, Weber, Moritz, Huger, Schlicht and Zollner}]{pavlitskayaUsingMixtureExpert2020}
\bibinfo{author}{Pavlitskaya, S.}, \bibinfo{author}{Hubschneider, C.}, \bibinfo{author}{Weber, M.}, \bibinfo{author}{Moritz, R.}, \bibinfo{author}{Huger, F.}, \bibinfo{author}{Schlicht, P.}, \bibinfo{author}{Zollner, J.M.}, \bibinfo{year}{2020}.
\newblock \bibinfo{title}{Using {{Mixture}} of {{Expert Models}} to {{Gain Insights}} into {{Semantic Segmentation}}}, in: \bibinfo{booktitle}{2020 {{IEEE}}/{{CVF Conference}} on {{Computer Vision}} and {{Pattern Recognition Workshops}} ({{CVPRW}})}, \bibinfo{publisher}{IEEE}, \bibinfo{address}{Seattle, WA, USA}. pp. \bibinfo{pages}{1399--1406}.
\newblock \DOIprefix\doi{10.1109/CVPRW50498.2020.00179}.
%Type = Inproceedings
\bibitem[{Peng et~al.(2024)Peng, Li, Zhang, Sun and Wu}]{pengSceneAdaptiveSparse2024}
\bibinfo{author}{Peng, Y.}, \bibinfo{author}{Li, H.}, \bibinfo{author}{Zhang, Y.}, \bibinfo{author}{Sun, X.}, \bibinfo{author}{Wu, F.}, \bibinfo{year}{2024}.
\newblock \bibinfo{title}{Scene {{Adaptive Sparse Transformer}} for {{Event-based Object Detection}}}, in: \bibinfo{booktitle}{2024 {{IEEE}}/{{CVF Conference}} on {{Computer Vision}} and {{Pattern Recognition}} ({{CVPR}})}, \bibinfo{publisher}{IEEE}, \bibinfo{address}{Seattle, WA, USA}. pp. \bibinfo{pages}{16794--16804}.
\newblock \DOIprefix\doi{10.1109/CVPR52733.2024.01589}.
%Type = Inproceedings
\bibitem[{Phuong and Lampert(2019)}]{phuongDistillationBasedTrainingMultiExit2019}
\bibinfo{author}{Phuong, M.}, \bibinfo{author}{Lampert, C.}, \bibinfo{year}{2019}.
\newblock \bibinfo{title}{Distillation-{{Based Training}} for {{Multi-Exit Architectures}}}, in: \bibinfo{booktitle}{2019 {{IEEE}}/{{CVF International Conference}} on {{Computer Vision}} ({{ICCV}})}, \bibinfo{publisher}{IEEE}, \bibinfo{address}{Seoul, Korea (South)}. pp. \bibinfo{pages}{1355--1364}.
\newblock \DOIprefix\doi{10.1109/ICCV.2019.00144}.
%Type = Inproceedings
\bibitem[{Polina et~al.(2023)Polina, Ekaterina, Anastasia, Andrei, Leonid, Riccardo and Aleksei}]{polinaFIANCEEFasterInference2023}
\bibinfo{author}{Polina, K.}, \bibinfo{author}{Ekaterina, R.}, \bibinfo{author}{Anastasia, Y.}, \bibinfo{author}{Andrei, S.}, \bibinfo{author}{Leonid, K.}, \bibinfo{author}{Riccardo, F.}, \bibinfo{author}{Aleksei, I.}, \bibinfo{year}{2023}.
\newblock \bibinfo{title}{{{FIANCEE}}: {{Faster Inference}} of {{Adversarial Networks}} via {{Conditional Early Exits}}}, in: \bibinfo{booktitle}{2023 {{IEEE}}/{{CVF Conference}} on {{Computer Vision}} and {{Pattern Recognition}} ({{CVPR}})}, \bibinfo{publisher}{IEEE}, \bibinfo{address}{Vancouver, BC, Canada}. pp. \bibinfo{pages}{12032--12043}.
\newblock \DOIprefix\doi{10.1109/CVPR52729.2023.01158}.
%Type = Article
\bibitem[{Pomponi et~al.(2022)Pomponi, Scardapane and Uncini}]{pomponiProbabilisticReIntepretationConfidence2022}
\bibinfo{author}{Pomponi, J.}, \bibinfo{author}{Scardapane, S.}, \bibinfo{author}{Uncini, A.}, \bibinfo{year}{2022}.
\newblock \bibinfo{title}{A {{Probabilistic Re-Intepretation}} of {{Confidence Scores}} in {{Multi-Exit Models}}}.
\newblock \bibinfo{journal}{Entropy} \bibinfo{volume}{24}, \bibinfo{pages}{1}.
\newblock \DOIprefix\doi{10.3390/e24010001}.
%Type = Article
\bibitem[{Rabinowitch et~al.(2016)Rabinowitch, Laurent, Zhao, Walker, Beets, Schoofs, Bai, Schafer and Treinin}]{rabinowitchNeuropeptideDrivenCrossModalPlasticity2016}
\bibinfo{author}{Rabinowitch, I.}, \bibinfo{author}{Laurent, P.}, \bibinfo{author}{Zhao, B.}, \bibinfo{author}{Walker, D.}, \bibinfo{author}{Beets, I.}, \bibinfo{author}{Schoofs, L.}, \bibinfo{author}{Bai, J.}, \bibinfo{author}{Schafer, W.R.}, \bibinfo{author}{Treinin, M.}, \bibinfo{year}{2016}.
\newblock \bibinfo{title}{Neuropeptide-{{Driven Cross-Modal Plasticity}} following {{Sensory Loss}} in {{Caenorhabditis}} elegans}.
\newblock \bibinfo{journal}{PLOS Biology} \bibinfo{volume}{14}, \bibinfo{pages}{e1002348}.
\newblock \DOIprefix\doi{10.1371/journal.pbio.1002348}.
%Type = Inproceedings
\bibitem[{Rao et~al.(2021)Rao, Zhao, Liu, Lu, Zhou and Hsieh}]{raoDynamicViTEfficientVision2021}
\bibinfo{author}{Rao, Y.}, \bibinfo{author}{Zhao, W.}, \bibinfo{author}{Liu, B.}, \bibinfo{author}{Lu, J.}, \bibinfo{author}{Zhou, J.}, \bibinfo{author}{Hsieh, C.J.}, \bibinfo{year}{2021}.
\newblock \bibinfo{title}{{{DynamicViT}}: {{Efficient Vision Transformers}} with {{Dynamic Token Sparsification}}}, in: \bibinfo{booktitle}{Advances in {{Neural Information Processing Systems}}}, \bibinfo{publisher}{Curran Associates, Inc.}. pp. \bibinfo{pages}{13937--13949}.
%Type = Inproceedings
\bibitem[{Rashid et~al.(2022)Rashid, Mortlock and Al~Faruque}]{rashidSELFCARESelectiveFusion2022}
\bibinfo{author}{Rashid, N.}, \bibinfo{author}{Mortlock, T.}, \bibinfo{author}{Al~Faruque, M.A.}, \bibinfo{year}{2022}.
\newblock \bibinfo{title}{{{SELF-CARE}}: {{Selective Fusion}} with {{Context-Aware Low-Power Edge Computing}} for {{Stress Detection}}}, in: \bibinfo{booktitle}{2022 18th {{International Conference}} on {{Distributed Computing}} in {{Sensor Systems}} ({{DCOSS}})}, pp. \bibinfo{pages}{49--52}.
\newblock \DOIprefix\doi{10.1109/DCOSS54816.2022.00019}.
%Type = Article
\bibitem[{Rashid et~al.(2023)Rashid, Mortlock and Faruque}]{rashidStressDetectionUsing2023a}
\bibinfo{author}{Rashid, N.}, \bibinfo{author}{Mortlock, T.}, \bibinfo{author}{Faruque, M.A.A.}, \bibinfo{year}{2023}.
\newblock \bibinfo{title}{Stress {{Detection Using Context-Aware Sensor Fusion From Wearable Devices}}}.
\newblock \bibinfo{journal}{IEEE Internet of Things Journal} \bibinfo{volume}{10}, \bibinfo{pages}{14114--14127}.
\newblock \DOIprefix\doi{10.1109/JIOT.2023.3265768}.
%Type = Misc
\bibitem[{Regol et~al.(2024a)Regol, Chataoui, Charpentier, Coates, Piantanida and Gunnemann}]{regolPredictingProbabilitiesError2024}
\bibinfo{author}{Regol, F.}, \bibinfo{author}{Chataoui, J.}, \bibinfo{author}{Charpentier, B.}, \bibinfo{author}{Coates, M.}, \bibinfo{author}{Piantanida, P.}, \bibinfo{author}{Gunnemann, S.}, \bibinfo{year}{2024}a.
\newblock \bibinfo{title}{Predicting {{Probabilities}} of {{Error}} to {{Combine Quantization}} and {{Early Exiting}}: {{QuEE}}}.
\newblock \href{http://arxiv.org/abs/2406.14404}{{\tt arXiv:2406.14404}}.
%Type = Misc
\bibitem[{Regol et~al.(2024b)Regol, Chataoui and Coates}]{regolJointlyLearnedExitInference2024}
\bibinfo{author}{Regol, F.}, \bibinfo{author}{Chataoui, J.}, \bibinfo{author}{Coates, M.}, \bibinfo{year}{2024}b.
\newblock \bibinfo{title}{Jointly-{{Learned Exit}} and {{Inference}} for a {{Dynamic Neural Network}} : {{JEI-DNN}}}.
\newblock \DOIprefix\doi{10.48550/arXiv.2310.09163}, \href{http://arxiv.org/abs/2310.09163}{{\tt arXiv:2310.09163}}.
%Type = Inproceedings
\bibitem[{Riquelme et~al.(2021)Riquelme, Puigcerver, Mustafa, Neumann, Jenatton, Susano~Pinto, Keysers and Houlsby}]{riquelmeScalingVisionSparse2021a}
\bibinfo{author}{Riquelme, C.}, \bibinfo{author}{Puigcerver, J.}, \bibinfo{author}{Mustafa, B.}, \bibinfo{author}{Neumann, M.}, \bibinfo{author}{Jenatton, R.}, \bibinfo{author}{Susano~Pinto, A.}, \bibinfo{author}{Keysers, D.}, \bibinfo{author}{Houlsby, N.}, \bibinfo{year}{2021}.
\newblock \bibinfo{title}{Scaling {{Vision}} with {{Sparse Mixture}} of {{Experts}}}, in: \bibinfo{booktitle}{Advances in {{Neural Information Processing Systems}}}, \bibinfo{publisher}{Curran Associates, Inc.}. pp. \bibinfo{pages}{8583--8595}.
%Type = Misc
\bibitem[{Rombach et~al.(2022)Rombach, Blattmann, Lorenz, Esser and Ommer}]{rombachHighResolutionImageSynthesis2022}
\bibinfo{author}{Rombach, R.}, \bibinfo{author}{Blattmann, A.}, \bibinfo{author}{Lorenz, D.}, \bibinfo{author}{Esser, P.}, \bibinfo{author}{Ommer, B.}, \bibinfo{year}{2022}.
\newblock \bibinfo{title}{High-{{Resolution Image Synthesis}} with {{Latent Diffusion Models}}}.
\newblock \DOIprefix\doi{10.48550/arXiv.2112.10752}, \href{http://arxiv.org/abs/2112.10752}{{\tt arXiv:2112.10752}}.
%Type = Article
\bibitem[{Rosenbaum et~al.(2018)Rosenbaum, Klinger and Riemer}]{rosenbaumRoutingNetworksAdaptive2018}
\bibinfo{author}{Rosenbaum, C.}, \bibinfo{author}{Klinger, T.}, \bibinfo{author}{Riemer, M.}, \bibinfo{year}{2018}.
\newblock \bibinfo{title}{Published as a conference paper at iclr 2018 routing networks: Routing {{Networks}}: {{Adaptive Selection}} of {{Non-linear Functions}} for {{Multi-Task Learning}}} \URLprefix \url{https://all.cs.umass.edu/pubs/2018/Rosenbaum%20et%20al%20-%20Routing%20Networks%20Adaptive%20Selection%20of%20Non-Linear%20Functions%20for%20Multi-Task%20Learning.pdf}.
%Type = Misc
\bibitem[{Sabet et~al.(2022)Sabet, Hare, {Al-Hashimi} and Merrett}]{sabetTemporalEarlyExits2022}
\bibinfo{author}{Sabet, A.}, \bibinfo{author}{Hare, J.}, \bibinfo{author}{{Al-Hashimi}, B.}, \bibinfo{author}{Merrett, G.V.}, \bibinfo{year}{2022}.
\newblock \bibinfo{title}{Temporal {{Early Exits}} for {{Efficient Video Object Detection}}}.
\newblock \DOIprefix\doi{10.2139/ssrn.4015043}, \href{http://arxiv.org/abs/4015043}{{\tt arXiv:4015043}}.
%Type = Article
\bibitem[{Scardapane et~al.(2024)Scardapane, Baiocchi, Devoto, Marsocci, Minervini and Pomponi}]{scardapaneConditionalComputationNeural2024a}
\bibinfo{author}{Scardapane, S.}, \bibinfo{author}{Baiocchi, A.}, \bibinfo{author}{Devoto, A.}, \bibinfo{author}{Marsocci, V.}, \bibinfo{author}{Minervini, P.}, \bibinfo{author}{Pomponi, J.}, \bibinfo{year}{2024}.
\newblock \bibinfo{title}{Conditional computation in neural networks: {{Principles}} and research trends}.
\newblock \bibinfo{journal}{Intelligenza Artificiale} \bibinfo{volume}{18}, \bibinfo{pages}{175--190}.
\newblock \DOIprefix\doi{10.3233/IA-240035}.
%Type = Inproceedings
\bibitem[{Scardapane et~al.(2020a)Scardapane, Comminiello, Scarpiniti, Baccarelli and Uncini}]{scardapaneDifferentiableBranchingDeep2020a}
\bibinfo{author}{Scardapane, S.}, \bibinfo{author}{Comminiello, D.}, \bibinfo{author}{Scarpiniti, M.}, \bibinfo{author}{Baccarelli, E.}, \bibinfo{author}{Uncini, A.}, \bibinfo{year}{2020}a.
\newblock \bibinfo{title}{Differentiable {{Branching In Deep Networks}} for {{Fast Inference}}}, in: \bibinfo{booktitle}{{{ICASSP}} 2020 - 2020 {{IEEE International Conference}} on {{Acoustics}}, {{Speech}} and {{Signal Processing}} ({{ICASSP}})}, pp. \bibinfo{pages}{4167--4171}.
\newblock \DOIprefix\doi{10.1109/ICASSP40776.2020.9054209}.
%Type = Article
\bibitem[{Scardapane et~al.(2020b)Scardapane, Scarpiniti, Baccarelli and Uncini}]{scardapaneWhyShouldWe2020a}
\bibinfo{author}{Scardapane, S.}, \bibinfo{author}{Scarpiniti, M.}, \bibinfo{author}{Baccarelli, E.}, \bibinfo{author}{Uncini, A.}, \bibinfo{year}{2020}b.
\newblock \bibinfo{title}{Why {{Should We Add Early Exits}} to {{Neural Networks}}?}
\newblock \bibinfo{journal}{Cognitive Computation} \bibinfo{volume}{12}, \bibinfo{pages}{954--966}.
\newblock \DOIprefix\doi{10.1007/s12559-020-09734-4}.
%Type = Article
\bibitem[{Schiebener et~al.(2013)Schiebener, Morimoto, Asfour and Ude}]{schiebenerIntegratingVisualPerception2013}
\bibinfo{author}{Schiebener, D.}, \bibinfo{author}{Morimoto, J.}, \bibinfo{author}{Asfour, T.}, \bibinfo{author}{Ude, A.}, \bibinfo{year}{2013}.
\newblock \bibinfo{title}{Integrating visual perception and manipulation for autonomous learning of object representations}.
\newblock \bibinfo{journal}{Adaptive Behavior} \DOIprefix\doi{10.1177/1059712313484502}.
%Type = Inproceedings
\bibitem[{Seol et~al.(2023)Seol, Roh and Chung}]{seolTokenMergingClass2023a}
\bibinfo{author}{Seol, K.S.}, \bibinfo{author}{Roh, S.D.}, \bibinfo{author}{Chung, K.S.}, \bibinfo{year}{2023}.
\newblock \bibinfo{title}{Token {{Merging}} with {{Class Importance Score}}}, in: \bibinfo{booktitle}{{{IECON}} 2023- 49th {{Annual Conference}} of the {{IEEE Industrial Electronics Society}}}, pp. \bibinfo{pages}{1--6}.
\newblock \DOIprefix\doi{10.1109/IECON51785.2023.10312420}.
%Type = Inproceedings
\bibitem[{Shao et~al.(2021)Shao, Zhang, Mao and Zhang}]{shaoBranchyGNNDeviceEdgeCoInference2021}
\bibinfo{author}{Shao, J.}, \bibinfo{author}{Zhang, H.}, \bibinfo{author}{Mao, Y.}, \bibinfo{author}{Zhang, J.}, \bibinfo{year}{2021}.
\newblock \bibinfo{title}{Branchy-{{GNN}}: {{A Device-Edge Co-Inference Framework}} for {{Efficient Point Cloud Processing}}}, in: \bibinfo{booktitle}{{{ICASSP}} 2021 - 2021 {{IEEE International Conference}} on {{Acoustics}}, {{Speech}} and {{Signal Processing}} ({{ICASSP}})}, pp. \bibinfo{pages}{8488--8492}.
\newblock \DOIprefix\doi{10.1109/ICASSP39728.2021.9414831}.
%Type = Inproceedings
\bibitem[{Shazeer et~al.(2018)Shazeer, Fatahalian, Mark and Mullapudi}]{shazeerHydraNetsSpecializedDynamic2018}
\bibinfo{author}{Shazeer, N.}, \bibinfo{author}{Fatahalian, K.}, \bibinfo{author}{Mark, W.R.}, \bibinfo{author}{Mullapudi, R.T.}, \bibinfo{year}{2018}.
\newblock \bibinfo{title}{{{HydraNets}}: {{Specialized Dynamic Architectures}} for {{Efficient Inference}}}, in: \bibinfo{booktitle}{2018 {{IEEE}}/{{CVF Conference}} on {{Computer Vision}} and {{Pattern Recognition}}}, \bibinfo{publisher}{IEEE}, \bibinfo{address}{Salt Lake City, UT}. pp. \bibinfo{pages}{8080--8089}.
\newblock \DOIprefix\doi{10.1109/CVPR.2018.00843}.
%Type = Article
\bibitem[{Shen et~al.(2020)Shen, Wang, Xu, Fu, Wang and Lin}]{shenFractionalSkippingFinerGrained2020}
\bibinfo{author}{Shen, J.}, \bibinfo{author}{Wang, Y.}, \bibinfo{author}{Xu, P.}, \bibinfo{author}{Fu, Y.}, \bibinfo{author}{Wang, Z.}, \bibinfo{author}{Lin, Y.}, \bibinfo{year}{2020}.
\newblock \bibinfo{title}{Fractional {{Skipping}}: {{Towards Finer-Grained Dynamic CNN Inference}}}.
\newblock \bibinfo{journal}{Proceedings of the AAAI Conference on Artificial Intelligence} \bibinfo{volume}{34}, \bibinfo{pages}{5700--5708}.
\newblock \DOIprefix\doi{10.1609/aaai.v34i04.6025}.
%Type = Article
\bibitem[{Sun et~al.(2025)Sun, Tan, Li, Hou, Li, Shao, Wang and Song}]{sunHotMoEExploringSparse2025}
\bibinfo{author}{Sun, W.}, \bibinfo{author}{Tan, Y.}, \bibinfo{author}{Li, J.}, \bibinfo{author}{Hou, S.}, \bibinfo{author}{Li, X.}, \bibinfo{author}{Shao, Y.}, \bibinfo{author}{Wang, Z.}, \bibinfo{author}{Song, B.}, \bibinfo{year}{2025}.
\newblock \bibinfo{title}{{{HotMoE}}: {{Exploring Sparse Mixture-of-Experts}} for {{Hyperspectral Object Tracking}}}.
\newblock \bibinfo{journal}{IEEE Transactions on Multimedia} \bibinfo{volume}{27}, \bibinfo{pages}{4072--4083}.
\newblock \DOIprefix\doi{10.1109/TMM.2025.3535339}.
%Type = Inproceedings
\bibitem[{Sun et~al.(2021)Sun, Panda, Chen, Oliva, Feris and Saenko}]{sunDynamicNetworkQuantization2021a}
\bibinfo{author}{Sun, X.}, \bibinfo{author}{Panda, R.}, \bibinfo{author}{Chen, C.F.R.}, \bibinfo{author}{Oliva, A.}, \bibinfo{author}{Feris, R.}, \bibinfo{author}{Saenko, K.}, \bibinfo{year}{2021}.
\newblock \bibinfo{title}{Dynamic {{Network Quantization}} for {{Efficient Video Inference}}}, in: \bibinfo{booktitle}{2021 {{IEEE}}/{{CVF International Conference}} on {{Computer Vision}} ({{ICCV}})}, \bibinfo{publisher}{IEEE}, \bibinfo{address}{Montreal, QC, Canada}. pp. \bibinfo{pages}{7355--7365}.
\newblock \DOIprefix\doi{10.1109/ICCV48922.2021.00728}.
%Type = Incollection
\bibitem[{Sun et~al.(2022)Sun, Li and Xu}]{sunMetaGFTrainingDynamicDepth2022a}
\bibinfo{author}{Sun, Y.}, \bibinfo{author}{Li, J.}, \bibinfo{author}{Xu, X.}, \bibinfo{year}{2022}.
\newblock \bibinfo{title}{Meta-{{GF}}: {{Training Dynamic-Depth Neural Networks Harmoniously}}}, in: \bibinfo{editor}{Avidan, S.}, \bibinfo{editor}{Brostow, G.}, \bibinfo{editor}{Ciss{\'e}, M.}, \bibinfo{editor}{Farinella, G.M.}, \bibinfo{editor}{Hassner, T.} (Eds.), \bibinfo{booktitle}{Computer {{Vision}} -- {{ECCV}} 2022}. \bibinfo{publisher}{Springer Nature Switzerland}, \bibinfo{address}{Cham}. volume \bibinfo{volume}{13671}, pp. \bibinfo{pages}{691--708}.
\newblock \DOIprefix\doi{10.1007/978-3-031-20083-0_41}.
%Type = Article
\bibitem[{Swaminathan et~al.(2020)Swaminathan, Garg, Kannan and Andres}]{swaminathanSparseLowRank2020}
\bibinfo{author}{Swaminathan, S.}, \bibinfo{author}{Garg, D.}, \bibinfo{author}{Kannan, R.}, \bibinfo{author}{Andres, F.}, \bibinfo{year}{2020}.
\newblock \bibinfo{title}{Sparse low rank factorization for deep neural network compression}.
\newblock \bibinfo{journal}{Neurocomputing} \bibinfo{volume}{398}, \bibinfo{pages}{185--196}.
\newblock \DOIprefix\doi{10.1016/j.neucom.2020.02.035}.
%Type = Inproceedings
\bibitem[{Szegedy et~al.(2015)Szegedy, {Wei Liu}, {Yangqing Jia}, Sermanet, Reed, Anguelov, Erhan, Vanhoucke and Rabinovich}]{szegedyGoingDeeperConvolutions2015}
\bibinfo{author}{Szegedy, C.}, \bibinfo{author}{{Wei Liu}}, \bibinfo{author}{{Yangqing Jia}}, \bibinfo{author}{Sermanet, P.}, \bibinfo{author}{Reed, S.}, \bibinfo{author}{Anguelov, D.}, \bibinfo{author}{Erhan, D.}, \bibinfo{author}{Vanhoucke, V.}, \bibinfo{author}{Rabinovich, A.}, \bibinfo{year}{2015}.
\newblock \bibinfo{title}{Going deeper with convolutions}, in: \bibinfo{booktitle}{2015 {{IEEE Conference}} on {{Computer Vision}} and {{Pattern Recognition}} ({{CVPR}})}, \bibinfo{publisher}{IEEE}, \bibinfo{address}{Boston, MA, USA}. pp. \bibinfo{pages}{1--9}.
\newblock \DOIprefix\doi{10.1109/CVPR.2015.7298594}.
%Type = Article
\bibitem[{Tan et~al.(2021)Tan, Li, Wang, Huang and Xu}]{tanEmpoweringAdaptiveEarlyExit2021}
\bibinfo{author}{Tan, X.}, \bibinfo{author}{Li, H.}, \bibinfo{author}{Wang, L.}, \bibinfo{author}{Huang, X.}, \bibinfo{author}{Xu, Z.}, \bibinfo{year}{2021}.
\newblock \bibinfo{title}{Empowering {{Adaptive Early-Exit Inference}} with {{Latency Awareness}}}.
\newblock \bibinfo{journal}{Proceedings of the AAAI Conference on Artificial Intelligence} \bibinfo{volume}{35}, \bibinfo{pages}{9825--9833}.
\newblock \DOIprefix\doi{10.1609/aaai.v35i11.17181}.
%Type = Inproceedings
\bibitem[{Tang et~al.(2023)Tang, Zhang, Liu, Liu and Liu}]{tangDynamicTokenPruning2023a}
\bibinfo{author}{Tang, Q.}, \bibinfo{author}{Zhang, B.}, \bibinfo{author}{Liu, J.}, \bibinfo{author}{Liu, F.}, \bibinfo{author}{Liu, Y.}, \bibinfo{year}{2023}.
\newblock \bibinfo{title}{Dynamic {{Token Pruning}} in {{Plain Vision Transformers}} for {{Semantic Segmentation}}}, in: \bibinfo{booktitle}{2023 {{IEEE}}/{{CVF International Conference}} on {{Computer Vision}} ({{ICCV}})}, \bibinfo{publisher}{IEEE}, \bibinfo{address}{Paris, France}. pp. \bibinfo{pages}{777--786}.
\newblock \DOIprefix\doi{10.1109/ICCV51070.2023.00078}.
%Type = Inproceedings
\bibitem[{Teerapittayanon et~al.(2016)Teerapittayanon, McDanel and Kung}]{teerapittayanonBranchyNetFastInference2016a}
\bibinfo{author}{Teerapittayanon, S.}, \bibinfo{author}{McDanel, B.}, \bibinfo{author}{Kung, H.}, \bibinfo{year}{2016}.
\newblock \bibinfo{title}{{{BranchyNet}}: {{Fast}} inference via early exiting from deep neural networks}, in: \bibinfo{booktitle}{2016 23rd {{International Conference}} on {{Pattern Recognition}} ({{ICPR}})}, pp. \bibinfo{pages}{2464--2469}.
\newblock \DOIprefix\doi{10.1109/ICPR.2016.7900006}.
%Type = Inproceedings
\bibitem[{Valada et~al.(2017)Valada, Vertens, Dhall and Burgard}]{valadaAdapNetAdaptiveSemantic2017}
\bibinfo{author}{Valada, A.}, \bibinfo{author}{Vertens, J.}, \bibinfo{author}{Dhall, A.}, \bibinfo{author}{Burgard, W.}, \bibinfo{year}{2017}.
\newblock \bibinfo{title}{{{AdapNet}}: {{Adaptive}} semantic segmentation in adverse environmental conditions}, in: \bibinfo{booktitle}{2017 {{IEEE International Conference}} on {{Robotics}} and {{Automation}} ({{ICRA}})}, pp. \bibinfo{pages}{4644--4651}.
\newblock \DOIprefix\doi{10.1109/ICRA.2017.7989540}.
%Type = Misc
\bibitem[{Valade et~al.(2024)Valade, Hebiri and Gay}]{valadeEEROEarlyExit2024a}
\bibinfo{author}{Valade, F.}, \bibinfo{author}{Hebiri, M.}, \bibinfo{author}{Gay, P.}, \bibinfo{year}{2024}.
\newblock \bibinfo{title}{{{EERO}}: {{Early Exit}} with {{Reject Option}} for {{Efficient Classification}} with limited budget}.
\newblock \href{http://arxiv.org/abs/2402.03779}{{\tt arXiv:2402.03779}}.
%Type = Article
\bibitem[{Veit and Belongie(2019)}]{veitConvolutionalNetworksAdaptive2018}
\bibinfo{author}{Veit, A.}, \bibinfo{author}{Belongie, S.}, \bibinfo{year}{2019}.
\newblock \bibinfo{title}{Convolutional networks with adaptive inference graphs}.
\newblock \bibinfo{journal}{International Journal of Computer Vision} \bibinfo{volume}{128}, \bibinfo{pages}{730–741}.
\newblock \DOIprefix\doi{https://doi.org/10.1007/s11263-019-01190-4}.
%Type = Inproceedings
\bibitem[{Verelst and Tuytelaars(2020)}]{verelstDynamicConvolutionsExploiting2020}
\bibinfo{author}{Verelst, T.}, \bibinfo{author}{Tuytelaars, T.}, \bibinfo{year}{2020}.
\newblock \bibinfo{title}{Dynamic {{Convolutions}}: {{Exploiting Spatial Sparsity}} for {{Faster Inference}}}, in: \bibinfo{booktitle}{Proceedings of the {{IEEE}}/{{CVF Conference}} on {{Computer Vision}} and {{Pattern Recognition}}}, pp. \bibinfo{pages}{2320--2329}.
%Type = Article
\bibitem[{Verelst and Tuytelaars(2023)}]{verelstSegBlocksBlockBasedDynamic2023}
\bibinfo{author}{Verelst, T.}, \bibinfo{author}{Tuytelaars, T.}, \bibinfo{year}{2023}.
\newblock \bibinfo{title}{{{SegBlocks}}: {{Block-Based Dynamic Resolution Networks}} for {{Real-Time Segmentation}}}.
\newblock \bibinfo{journal}{IEEE Transactions on Pattern Analysis and Machine Intelligence} \bibinfo{volume}{45}, \bibinfo{pages}{2400--2411}.
\newblock \DOIprefix\doi{10.1109/TPAMI.2022.3162528}.
%Type = Article
\bibitem[{Walther et~al.(2010)Walther, Chai, Caddigan, Beck and {Fei-Fei}}]{waltherFMRIDecodingNatural2010}
\bibinfo{author}{Walther, D.}, \bibinfo{author}{Chai, B.}, \bibinfo{author}{Caddigan, E.}, \bibinfo{author}{Beck, D.}, \bibinfo{author}{{Fei-Fei}, L.}, \bibinfo{year}{2010}.
\newblock \bibinfo{title}{{{fMRI Decoding}} of {{Natural Scene Categories}} from {{Line Drawings}}}.
\newblock \bibinfo{journal}{Journal of Vision} \bibinfo{volume}{10}, \bibinfo{pages}{1221}.
\newblock \DOIprefix\doi{10.1167/10.7.1221}.
%Type = Inproceedings
\bibitem[{Wang et~al.(2019a)Wang, Mo, Lin, Wang and Du}]{wangDynExitDynamicEarlyExit2019}
\bibinfo{author}{Wang, M.}, \bibinfo{author}{Mo, J.}, \bibinfo{author}{Lin, J.}, \bibinfo{author}{Wang, Z.}, \bibinfo{author}{Du, L.}, \bibinfo{year}{2019}a.
\newblock \bibinfo{title}{{{DynExit}}: {{A Dynamic Early-Exit Strategy}} for {{Deep Residual Networks}}}, in: \bibinfo{booktitle}{2019 {{IEEE International Workshop}} on {{Signal Processing Systems}} ({{SiPS}})}, pp. \bibinfo{pages}{178--183}.
\newblock \DOIprefix\doi{10.1109/SiPS47522.2019.9020551}.
%Type = Article
\bibitem[{Wang and Li(2021)}]{wangHarmonizedDenseKnowledge2021}
\bibinfo{author}{Wang, X.}, \bibinfo{author}{Li, Y.}, \bibinfo{year}{2021}.
\newblock \bibinfo{title}{Harmonized {{Dense Knowledge Distillation Training}} for {{Multi-Exit Architectures}}}.
\newblock \bibinfo{journal}{Proceedings of the AAAI Conference on Artificial Intelligence} \bibinfo{volume}{35}, \bibinfo{pages}{10218--10226}.
\newblock \DOIprefix\doi{10.1609/aaai.v35i11.17225}.
%Type = Inproceedings
\bibitem[{Wang et~al.(2018a)Wang, Yu, Dou, Darrell and Gonzalez}]{wangSkipNetLearningDynamic2018}
\bibinfo{author}{Wang, X.}, \bibinfo{author}{Yu, F.}, \bibinfo{author}{Dou, Z.Y.}, \bibinfo{author}{Darrell, T.}, \bibinfo{author}{Gonzalez, J.E.}, \bibinfo{year}{2018}a.
\newblock \bibinfo{title}{{{SkipNet}}: {{Learning Dynamic Routing}} in {{Convolutional Networks}}}, in: \bibinfo{booktitle}{Proceedings of the {{European Conference}} on {{Computer Vision}} ({{ECCV}})}, pp. \bibinfo{pages}{409--424}.
%Type = Inproceedings
\bibitem[{Wang et~al.(2020a)Wang, Yu, Dunlap, Ma, Wang, Mirhoseini, Darrell and Gonzalez}]{wangDeepMixtureExperts2020}
\bibinfo{author}{Wang, X.}, \bibinfo{author}{Yu, F.}, \bibinfo{author}{Dunlap, L.}, \bibinfo{author}{Ma, Y.A.}, \bibinfo{author}{Wang, R.}, \bibinfo{author}{Mirhoseini, A.}, \bibinfo{author}{Darrell, T.}, \bibinfo{author}{Gonzalez, J.E.}, \bibinfo{year}{2020}a.
\newblock \bibinfo{title}{Deep {{Mixture}} of {{Experts}} via {{Shallow Embedding}}}, in: \bibinfo{booktitle}{Proceedings of {{The}} 35th {{Uncertainty}} in {{Artificial Intelligence Conference}}}, \bibinfo{publisher}{PMLR}. pp. \bibinfo{pages}{552--562}.
%Type = Article
\bibitem[{Wang et~al.(2025)Wang, Zheng, Shao, Duan and Deng}]{wangAdaptiveRectangularConvolution}
\bibinfo{author}{Wang, X.}, \bibinfo{author}{Zheng, Z.}, \bibinfo{author}{Shao, J.}, \bibinfo{author}{Duan, Y.}, \bibinfo{author}{Deng, L.J.}, \bibinfo{year}{2025}.
\newblock \bibinfo{title}{Adaptive {{Rectangular Convolution}} for {{Remote Sensing Pansharpening}}} \URLprefix \url{https://openaccess.thecvf.com/content/CVPR2025/papers/Wang_Adaptive_Rectangular_Convolution_for_Remote_Sensing_Pansharpening_CVPR_2025_paper.pdf}.
%Type = Inproceedings
\bibitem[{Wang et~al.(2021a)Wang, Chen, Jiang, Song, Han and Huang}]{wangAdaptiveFocusEfficient2021}
\bibinfo{author}{Wang, Y.}, \bibinfo{author}{Chen, Z.}, \bibinfo{author}{Jiang, H.}, \bibinfo{author}{Song, S.}, \bibinfo{author}{Han, Y.}, \bibinfo{author}{Huang, G.}, \bibinfo{year}{2021}a.
\newblock \bibinfo{title}{Adaptive {{Focus}} for {{Efficient Video Recognition}}}, in: \bibinfo{booktitle}{2021 {{IEEE}}/{{CVF International Conference}} on {{Computer Vision}} ({{ICCV}})}, \bibinfo{publisher}{IEEE}, \bibinfo{address}{Montreal, QC, Canada}. pp. \bibinfo{pages}{16229--16238}.
\newblock \DOIprefix\doi{10.1109/ICCV48922.2021.01594}.
%Type = Inproceedings
\bibitem[{Wang et~al.(2021b)Wang, Huang, Song, Huang and Huang}]{wangNotAllImages2021a}
\bibinfo{author}{Wang, Y.}, \bibinfo{author}{Huang, R.}, \bibinfo{author}{Song, S.}, \bibinfo{author}{Huang, Z.}, \bibinfo{author}{Huang, G.}, \bibinfo{year}{2021}b.
\newblock \bibinfo{title}{Not {{All Images}} are {{Worth}} 16x16 {{Words}}: {{Dynamic Transformers}} for {{Efficient Image Recognition}}}, in: \bibinfo{booktitle}{Advances in {{Neural Information Processing Systems}}}, \bibinfo{publisher}{Curran Associates, Inc.}. pp. \bibinfo{pages}{11960--11973}.
%Type = Article
\bibitem[{Wang et~al.(2018b)Wang, Nguyen, Zhao, Wang, Lin and Baraniuk}]{wangEnergyNetEnergyEfficientDynamic2018}
\bibinfo{author}{Wang, Y.}, \bibinfo{author}{Nguyen, T.}, \bibinfo{author}{Zhao, Y.}, \bibinfo{author}{Wang, Z.}, \bibinfo{author}{Lin, Y.}, \bibinfo{author}{Baraniuk, R.}, \bibinfo{year}{2018}b.
\newblock \bibinfo{title}{{{EnergyNet}}: {{Energy-Efficient Dynamic Inference}}}.
\newblock \bibinfo{journal}{2018} \URLprefix \url{https://openreview.net/pdf?id=Syxp2bgKoX}.
%Type = Article
\bibitem[{Wang et~al.(2020b)Wang, Shen, Hu, Xu, Nguyen, Baraniuk, Wang and Lin}]{wangDualDynamicInference2020}
\bibinfo{author}{Wang, Y.}, \bibinfo{author}{Shen, J.}, \bibinfo{author}{Hu, T.K.}, \bibinfo{author}{Xu, P.}, \bibinfo{author}{Nguyen, T.}, \bibinfo{author}{Baraniuk, R.}, \bibinfo{author}{Wang, Z.}, \bibinfo{author}{Lin, Y.}, \bibinfo{year}{2020}b.
\newblock \bibinfo{title}{Dual {{Dynamic Inference}}: {{Enabling More Efficient}}, {{Adaptive}}, and {{Controllable Deep Inference}}}.
\newblock \bibinfo{journal}{IEEE Journal of Selected Topics in Signal Processing} \bibinfo{volume}{14}, \bibinfo{pages}{623--633}.
\newblock \DOIprefix\doi{10.1109/JSTSP.2020.2979669}.
%Type = Inproceedings
\bibitem[{Wang et~al.(2022a)Wang, Yue, Lin, Jiang, Lai, Kulikov, Orlov, Shi and Huang}]{wangAdaFocusV2EndtoEnd2022}
\bibinfo{author}{Wang, Y.}, \bibinfo{author}{Yue, Y.}, \bibinfo{author}{Lin, Y.}, \bibinfo{author}{Jiang, H.}, \bibinfo{author}{Lai, Z.}, \bibinfo{author}{Kulikov, V.}, \bibinfo{author}{Orlov, N.}, \bibinfo{author}{Shi, H.}, \bibinfo{author}{Huang, G.}, \bibinfo{year}{2022}a.
\newblock \bibinfo{title}{{{AdaFocus V2}}: {{End-to-End Training}} of {{Spatial Dynamic Networks}} for {{Video Recognition}}}, in: \bibinfo{booktitle}{2022 {{IEEE}}/{{CVF Conference}} on {{Computer Vision}} and {{Pattern Recognition}} ({{CVPR}})}, \bibinfo{publisher}{IEEE}, \bibinfo{address}{New Orleans, LA, USA}. pp. \bibinfo{pages}{20030--20040}.
\newblock \DOIprefix\doi{10.1109/CVPR52688.2022.01943}.
%Type = Incollection
\bibitem[{Wang et~al.(2022b)Wang, Yue, Xu, Hassani, Kulikov, Orlov, Song, Shi and Huang}]{wangAdaFocusV3UnifiedSpatialTemporal2022}
\bibinfo{author}{Wang, Y.}, \bibinfo{author}{Yue, Y.}, \bibinfo{author}{Xu, X.}, \bibinfo{author}{Hassani, A.}, \bibinfo{author}{Kulikov, V.}, \bibinfo{author}{Orlov, N.}, \bibinfo{author}{Song, S.}, \bibinfo{author}{Shi, H.}, \bibinfo{author}{Huang, G.}, \bibinfo{year}{2022}b.
\newblock \bibinfo{title}{{{AdaFocusV3}}: {{On Unified Spatial-Temporal Dynamic Video Recognition}}}, in: \bibinfo{editor}{Avidan, S.}, \bibinfo{editor}{Brostow, G.}, \bibinfo{editor}{Ciss{\'e}, M.}, \bibinfo{editor}{Farinella, G.M.}, \bibinfo{editor}{Hassner, T.} (Eds.), \bibinfo{booktitle}{Computer {{Vision}} -- {{ECCV}} 2022}. \bibinfo{publisher}{Springer Nature Switzerland}, \bibinfo{address}{Cham}. volume \bibinfo{volume}{13664}, pp. \bibinfo{pages}{226--243}.
\newblock \DOIprefix\doi{10.1007/978-3-031-19772-7_14}.
%Type = Inproceedings
\bibitem[{Wang et~al.(2019b)Wang, Bao, Yuan, Ge, Tran and Zomaya}]{wangSEESchedulingEarly2019}
\bibinfo{author}{Wang, Z.}, \bibinfo{author}{Bao, W.}, \bibinfo{author}{Yuan, D.}, \bibinfo{author}{Ge, L.}, \bibinfo{author}{Tran, N.H.}, \bibinfo{author}{Zomaya, A.Y.}, \bibinfo{year}{2019}b.
\newblock \bibinfo{title}{{{SEE}}: {{Scheduling Early Exit}} for {{Mobile DNN Inference}} during {{Service Outage}}}, in: \bibinfo{booktitle}{Proceedings of the 22nd {{International ACM Conference}} on {{Modeling}}, {{Analysis}} and {{Simulation}} of {{Wireless}} and {{Mobile Systems}}}, \bibinfo{publisher}{ACM}, \bibinfo{address}{Miami Beach FL USA}. pp. \bibinfo{pages}{279--288}.
\newblock \DOIprefix\doi{10.1145/3345768.3355917}.
%Type = Inproceedings
\bibitem[{Wei et~al.(2023)Wei, Ye, Zhang, Tang and Liang}]{weiJointTokenPruning2023}
\bibinfo{author}{Wei, S.}, \bibinfo{author}{Ye, T.}, \bibinfo{author}{Zhang, S.}, \bibinfo{author}{Tang, Y.}, \bibinfo{author}{Liang, J.}, \bibinfo{year}{2023}.
\newblock \bibinfo{title}{Joint {{Token Pruning}} and {{Squeezing Towards More Aggressive Compression}} of {{Vision Transformers}}}, in: \bibinfo{booktitle}{2023 {{IEEE}}/{{CVF Conference}} on {{Computer Vision}} and {{Pattern Recognition}} ({{CVPR}})}, \bibinfo{publisher}{IEEE}, \bibinfo{address}{Vancouver, BC, Canada}. pp. \bibinfo{pages}{2092--2101}.
\newblock \DOIprefix\doi{10.1109/CVPR52729.2023.00208}.
%Type = Misc
\bibitem[{White et~al.(2023)White, Safari, Sukthanker, Ru, Elsken, Zela, Dey and Hutter}]{whiteNeuralArchitectureSearch2023}
\bibinfo{author}{White, C.}, \bibinfo{author}{Safari, M.}, \bibinfo{author}{Sukthanker, R.}, \bibinfo{author}{Ru, B.}, \bibinfo{author}{Elsken, T.}, \bibinfo{author}{Zela, A.}, \bibinfo{author}{Dey, D.}, \bibinfo{author}{Hutter, F.}, \bibinfo{year}{2023}.
\newblock \bibinfo{title}{Neural {{Architecture Search}}: {{Insights}} from 1000 {{Papers}}}.
\newblock \DOIprefix\doi{10.48550/arXiv.2301.08727}, \href{http://arxiv.org/abs/2301.08727}{{\tt arXiv:2301.08727}}.
%Type = Article
\bibitem[{Williams(1992)}]{williamsSimpleStatisticalGradientfollowing1992}
\bibinfo{author}{Williams, R.J.}, \bibinfo{year}{1992}.
\newblock \bibinfo{title}{Simple statistical gradient-following algorithms for connectionist reinforcement learning}.
\newblock \bibinfo{journal}{Machine Learning} \bibinfo{volume}{8}, \bibinfo{pages}{229--256}.
\newblock \DOIprefix\doi{10.1007/BF00992696}.
%Type = Article
\bibitem[{W{\'o}jcik et~al.(2023)W{\'o}jcik, Przewi{\c e}{\'z}likowski, Szatkowski, Wo{\l}czyk, Ba{\l}azy, Krzepkowski, Podolak, Tabor, {\'S}mieja and Trzci{\'n}ski}]{wojcikZeroTimeWaste2023a}
\bibinfo{author}{W{\'o}jcik, B.}, \bibinfo{author}{Przewi{\c e}{\'z}likowski, M.}, \bibinfo{author}{Szatkowski, F.}, \bibinfo{author}{Wo{\l}czyk, M.}, \bibinfo{author}{Ba{\l}azy, K.}, \bibinfo{author}{Krzepkowski, B.}, \bibinfo{author}{Podolak, I.}, \bibinfo{author}{Tabor, J.}, \bibinfo{author}{{\'S}mieja, M.}, \bibinfo{author}{Trzci{\'n}ski, T.}, \bibinfo{year}{2023}.
\newblock \bibinfo{title}{Zero time waste in pre-trained early exit neural networks}.
\newblock \bibinfo{journal}{Neural Networks} \bibinfo{volume}{168}, \bibinfo{pages}{580--601}.
\newblock \DOIprefix\doi{10.1016/j.neunet.2023.10.003}.
%Type = Inproceedings
\bibitem[{{Wo{\l}czyk} et~al.(2021){Wo{\l}czyk}, W{\'o}jcik, {Ba{\l} azy}, Podolak, Tabor, {\'S}mieja and Trzcinski}]{wolczykZeroTimeWaste2021}
\bibinfo{author}{{Wo{\l}czyk}, M.}, \bibinfo{author}{W{\'o}jcik, B.}, \bibinfo{author}{{Ba{\l} azy}, K.}, \bibinfo{author}{Podolak, I.T.}, \bibinfo{author}{Tabor, J.}, \bibinfo{author}{{\'S}mieja, M.}, \bibinfo{author}{Trzcinski, T.}, \bibinfo{year}{2021}.
\newblock \bibinfo{title}{Zero {{Time Waste}}: {{Recycling Predictions}} in {{Early Exit Neural Networks}}}, in: \bibinfo{booktitle}{Advances in {{Neural Information Processing Systems}}}, pp. \bibinfo{pages}{2516--2528}.
%Type = Inproceedings
\bibitem[{Wu et~al.(2018)Wu, Nagarajan, Kumar, Rennie, Davis, Grauman and Feris}]{wuBlockDropDynamicInference2018}
\bibinfo{author}{Wu, Z.}, \bibinfo{author}{Nagarajan, T.}, \bibinfo{author}{Kumar, A.}, \bibinfo{author}{Rennie, S.}, \bibinfo{author}{Davis, L.S.}, \bibinfo{author}{Grauman, K.}, \bibinfo{author}{Feris, R.}, \bibinfo{year}{2018}.
\newblock \bibinfo{title}{{{BlockDrop}}: {{Dynamic Inference Paths}} in {{Residual Networks}}}, in: \bibinfo{booktitle}{2018 {{IEEE}}/{{CVF Conference}} on {{Computer Vision}} and {{Pattern Recognition}}}, \bibinfo{publisher}{IEEE}, \bibinfo{address}{Salt Lake City, UT}. pp. \bibinfo{pages}{8817--8826}.
\newblock \DOIprefix\doi{10.1109/CVPR.2018.00919}.
%Type = Inproceedings
\bibitem[{Xia and Bouganis(2023)}]{xiaWindowBasedEarlyExitCascades2023}
\bibinfo{author}{Xia, G.}, \bibinfo{author}{Bouganis, C.S.}, \bibinfo{year}{2023}.
\newblock \bibinfo{title}{Window-{{Based Early-Exit Cascades}} for {{Uncertainty Estimation}}: {{When Deep Ensembles}} are {{More Efficient}} than {{Single Models}}}, in: \bibinfo{booktitle}{2023 {{IEEE}}/{{CVF International Conference}} on {{Computer Vision}} ({{ICCV}})}, \bibinfo{publisher}{IEEE}, \bibinfo{address}{Paris, France}. pp. \bibinfo{pages}{17322--17334}.
\newblock \DOIprefix\doi{10.1109/ICCV51070.2023.01593}.
%Type = Article
\bibitem[{Xie et~al.(2024)Xie, Zhang, Zhuang, Shi, Liu, Gu and Zhang}]{xieMoDEMixtureofExpertsModel2024}
\bibinfo{author}{Xie, Z.}, \bibinfo{author}{Zhang, Y.}, \bibinfo{author}{Zhuang, C.}, \bibinfo{author}{Shi, Q.}, \bibinfo{author}{Liu, Z.}, \bibinfo{author}{Gu, J.}, \bibinfo{author}{Zhang, G.}, \bibinfo{year}{2024}.
\newblock \bibinfo{title}{{{MoDE}}: {{A Mixture-of-Experts Model}} with {{Mutual Distillation}} among the {{Experts}}}.
\newblock \bibinfo{journal}{Proceedings of the AAAI Conference on Artificial Intelligence} \bibinfo{volume}{38}, \bibinfo{pages}{16067--16075}.
\newblock \DOIprefix\doi{10.1609/aaai.v38i14.29539}.
%Type = Incollection
\bibitem[{Xing et~al.(2020)Xing, Xu, Li and Guan}]{xingEarlyExitNot2020}
\bibinfo{author}{Xing, Q.}, \bibinfo{author}{Xu, M.}, \bibinfo{author}{Li, T.}, \bibinfo{author}{Guan, Z.}, \bibinfo{year}{2020}.
\newblock \bibinfo{title}{Early {{Exit}} or {{Not}}: {{Resource-Efficient Blind Quality Enhancement}} for {{Compressed Images}}}, in: \bibinfo{editor}{Vedaldi, A.}, \bibinfo{editor}{Bischof, H.}, \bibinfo{editor}{Brox, T.}, \bibinfo{editor}{Frahm, J.M.} (Eds.), \bibinfo{booktitle}{Computer {{Vision}} -- {{ECCV}} 2020}. \bibinfo{publisher}{Springer International Publishing}, \bibinfo{address}{Cham}. volume \bibinfo{volume}{12361}, pp. \bibinfo{pages}{275--292}.
\newblock \DOIprefix\doi{10.1007/978-3-030-58517-4_17}.
%Type = Inproceedings
\bibitem[{Xu and McAuley(2023)}]{xuSurveyDynamicNeural2023a}
\bibinfo{author}{Xu, C.}, \bibinfo{author}{McAuley, J.}, \bibinfo{year}{2023}.
\newblock \bibinfo{title}{A {{Survey}} on {{Dynamic Neural Networks}} for {{Natural Language Processing}}}, in: \bibinfo{editor}{Vlachos, A.}, \bibinfo{editor}{Augenstein, I.} (Eds.), \bibinfo{booktitle}{Findings of the {{Association}} for {{Computational Linguistics}}: {{EACL}} 2023}, \bibinfo{publisher}{Association for Computational Linguistics}, \bibinfo{address}{Dubrovnik, Croatia}. pp. \bibinfo{pages}{2370--2381}.
\newblock \DOIprefix\doi{10.18653/v1/2023.findings-eacl.180}.
%Type = Misc
\bibitem[{Xu et~al.(2023)Xu, Hao, Shen, Hu, Luo, Lin and Shen}]{xuLGViTDynamicEarly2023a}
\bibinfo{author}{Xu, G.}, \bibinfo{author}{Hao, J.}, \bibinfo{author}{Shen, L.}, \bibinfo{author}{Hu, H.}, \bibinfo{author}{Luo, Y.}, \bibinfo{author}{Lin, H.}, \bibinfo{author}{Shen, J.}, \bibinfo{year}{2023}.
\newblock \bibinfo{title}{{{LGViT}}: {{Dynamic Early Exiting}} for {{Accelerating Vision Transformer}}}.
\newblock \DOIprefix\doi{10.48550/arXiv.2308.00255}, \href{http://arxiv.org/abs/2308.00255}{{\tt arXiv:2308.00255}}.
%Type = Article
\bibitem[{Xu et~al.(2024a)Xu, Wang and Guo}]{xuATFTransAttentionweightedToken2024a}
\bibinfo{author}{Xu, L.}, \bibinfo{author}{Wang, L.}, \bibinfo{author}{Guo, Z.}, \bibinfo{year}{2024}a.
\newblock \bibinfo{title}{{{ATFTrans}}: Attention-weighted token fusion transformer for robust and efficient object tracking}.
\newblock \bibinfo{journal}{Neural Computing and Applications} \bibinfo{volume}{36}, \bibinfo{pages}{7043--7056}.
\newblock \DOIprefix\doi{10.1007/s00521-024-09444-0}.
%Type = Inproceedings
\bibitem[{Xu et~al.(2024b)Xu, Li, Chen, Chang, Liu and Wang}]{xuNoTokenLeft2024a}
\bibinfo{author}{Xu, X.}, \bibinfo{author}{Li, C.}, \bibinfo{author}{Chen, Y.}, \bibinfo{author}{Chang, X.}, \bibinfo{author}{Liu, J.}, \bibinfo{author}{Wang, S.}, \bibinfo{year}{2024}b.
\newblock \bibinfo{title}{No {{Token Left Behind}}: {{Efficient Vision Transformer}} via~{{Dynamic Token Idling}}}, in: \bibinfo{editor}{Liu, T.}, \bibinfo{editor}{Webb, G.}, \bibinfo{editor}{Yue, L.}, \bibinfo{editor}{Wang, D.} (Eds.), \bibinfo{booktitle}{{{AI}} 2023: {{Advances}} in {{Artificial Intelligence}}}, \bibinfo{publisher}{Springer Nature}, \bibinfo{address}{Singapore}. pp. \bibinfo{pages}{28--41}.
\newblock \DOIprefix\doi{10.1007/978-981-99-8388-9_3}.
%Type = Inproceedings
\bibitem[{Xu et~al.(2024c)Xu, Wang, Chen, Zheng, Wei and Liu}]{xuGTPViTEfficientVision2024}
\bibinfo{author}{Xu, X.}, \bibinfo{author}{Wang, S.}, \bibinfo{author}{Chen, Y.}, \bibinfo{author}{Zheng, Y.}, \bibinfo{author}{Wei, Z.}, \bibinfo{author}{Liu, J.}, \bibinfo{year}{2024}c.
\newblock \bibinfo{title}{{{GTP-ViT}}: {{Efficient Vision Transformers}} via {{Graph-based Token Propagation}}}, in: \bibinfo{booktitle}{2024 {{IEEE}}/{{CVF Winter Conference}} on {{Applications}} of {{Computer Vision}} ({{WACV}})}, \bibinfo{publisher}{IEEE}, \bibinfo{address}{Waikoloa, HI, USA}. pp. \bibinfo{pages}{86--95}.
\newblock \DOIprefix\doi{10.1109/WACV57701.2024.00016}.
%Type = Article
\bibitem[{Xu et~al.(2022)Xu, Zhang, Zhang, Sheng, Li, Dong, Zhang, Xu and Sun}]{xuEvoViTSlowFastToken2022}
\bibinfo{author}{Xu, Y.}, \bibinfo{author}{Zhang, Z.}, \bibinfo{author}{Zhang, M.}, \bibinfo{author}{Sheng, K.}, \bibinfo{author}{Li, K.}, \bibinfo{author}{Dong, W.}, \bibinfo{author}{Zhang, L.}, \bibinfo{author}{Xu, C.}, \bibinfo{author}{Sun, X.}, \bibinfo{year}{2022}.
\newblock \bibinfo{title}{Evo-{{ViT}}: {{Slow-Fast Token Evolution}} for {{Dynamic Vision Transformer}}}.
\newblock \bibinfo{journal}{Proceedings of the AAAI Conference on Artificial Intelligence} \bibinfo{volume}{36}, \bibinfo{pages}{2964--2972}.
\newblock \DOIprefix\doi{10.1609/aaai.v36i3.20202}.
%Type = Misc
\bibitem[{Xue et~al.(2023)Xue, Likhosherstov, Arnab, Houlsby, Dehghani and You}]{xueAdaptiveComputationElastic2023}
\bibinfo{author}{Xue, F.}, \bibinfo{author}{Likhosherstov, V.}, \bibinfo{author}{Arnab, A.}, \bibinfo{author}{Houlsby, N.}, \bibinfo{author}{Dehghani, M.}, \bibinfo{author}{You, Y.}, \bibinfo{year}{2023}.
\newblock \bibinfo{title}{Adaptive {{Computation}} with {{Elastic Input Sequence}}}.
\newblock \href{http://arxiv.org/abs/2301.13195}{{\tt arXiv:2301.13195}}.
%Type = Inproceedings
\bibitem[{Yang et~al.(2020)Yang, Han, Chen, Song, Dai and Huang}]{yangResolutionAdaptiveNetworks2020}
\bibinfo{author}{Yang, L.}, \bibinfo{author}{Han, Y.}, \bibinfo{author}{Chen, X.}, \bibinfo{author}{Song, S.}, \bibinfo{author}{Dai, J.}, \bibinfo{author}{Huang, G.}, \bibinfo{year}{2020}.
\newblock \bibinfo{title}{Resolution {{Adaptive Networks}} for {{Efficient Inference}}}, in: \bibinfo{booktitle}{2020 {{IEEE}}/{{CVF Conference}} on {{Computer Vision}} and {{Pattern Recognition}} ({{CVPR}})}, \bibinfo{publisher}{IEEE}, \bibinfo{address}{Seattle, WA, USA}. pp. \bibinfo{pages}{2366--2375}.
\newblock \DOIprefix\doi{10.1109/CVPR42600.2020.00244}.
%Type = Incollection
\bibitem[{Yang et~al.(2024a)Yang, Zheng, Han, Cheng, Song, Huang and Li}]{yangDyFADetDynamicFeature2024}
\bibinfo{author}{Yang, L.}, \bibinfo{author}{Zheng, Z.}, \bibinfo{author}{Han, Y.}, \bibinfo{author}{Cheng, H.}, \bibinfo{author}{Song, S.}, \bibinfo{author}{Huang, G.}, \bibinfo{author}{Li, F.}, \bibinfo{year}{2024}a.
\newblock \bibinfo{title}{{{DyFADet}}: {{Dynamic Feature Aggregation}} for {{Temporal Action Detection}}}, in: \bibinfo{editor}{Leonardis, A.}, \bibinfo{editor}{Ricci, E.}, \bibinfo{editor}{Roth, S.}, \bibinfo{editor}{Russakovsky, O.}, \bibinfo{editor}{Sattler, T.}, \bibinfo{editor}{Varol, G.} (Eds.), \bibinfo{booktitle}{Computer {{Vision}} -- {{ECCV}} 2024}. \bibinfo{publisher}{Springer Nature Switzerland}, \bibinfo{address}{Cham}. volume \bibinfo{volume}{15104}, pp. \bibinfo{pages}{305--322}.
\newblock \DOIprefix\doi{10.1007/978-3-031-72952-2_18}.
%Type = Article
\bibitem[{Yang et~al.(2024b)Yang, Zheng, Wang, Song, Huang and Li}]{yangAdaDetAdaptiveObject2024}
\bibinfo{author}{Yang, L.}, \bibinfo{author}{Zheng, Z.}, \bibinfo{author}{Wang, J.}, \bibinfo{author}{Song, S.}, \bibinfo{author}{Huang, G.}, \bibinfo{author}{Li, F.}, \bibinfo{year}{2024}b.
\newblock \bibinfo{title}{{{AdaDet}}: {{An Adaptive Object Detection System Based}} on {{Early-Exit Neural Networks}}}.
\newblock \bibinfo{journal}{IEEE Transactions on Cognitive and Developmental Systems} \bibinfo{volume}{16}, \bibinfo{pages}{332--345}.
\newblock \DOIprefix\doi{10.1109/TCDS.2023.3274214}.
%Type = Inproceedings
\bibitem[{Yin et~al.(2022)Yin, Vahdat, Alvarez, Mallya, Kautz and Molchanov}]{yinAViTAdaptiveTokens2022}
\bibinfo{author}{Yin, H.}, \bibinfo{author}{Vahdat, A.}, \bibinfo{author}{Alvarez, J.M.}, \bibinfo{author}{Mallya, A.}, \bibinfo{author}{Kautz, J.}, \bibinfo{author}{Molchanov, P.}, \bibinfo{year}{2022}.
\newblock \bibinfo{title}{A-{{ViT}}: {{Adaptive Tokens}} for {{Efficient Vision Transformer}}}, in: \bibinfo{booktitle}{2022 {{IEEE}}/{{CVF Conference}} on {{Computer Vision}} and {{Pattern Recognition}} ({{CVPR}})}, \bibinfo{publisher}{IEEE}, \bibinfo{address}{New Orleans, LA, USA}. pp. \bibinfo{pages}{10799--10808}.
\newblock \DOIprefix\doi{10.1109/CVPR52688.2022.01054}.
%Type = Misc
\bibitem[{Yu et~al.(2022)Yu, Li, Hua, Huang and Shi}]{yuBoostedDynamicNeural2022}
\bibinfo{author}{Yu, H.}, \bibinfo{author}{Li, H.}, \bibinfo{author}{Hua, G.}, \bibinfo{author}{Huang, G.}, \bibinfo{author}{Shi, H.}, \bibinfo{year}{2022}.
\newblock \bibinfo{title}{Boosted {{Dynamic Neural Networks}}}.
\newblock \DOIprefix\doi{10.48550/arXiv.2211.16726}, \href{http://arxiv.org/abs/2211.16726}{{\tt arXiv:2211.16726}}.
%Type = Inproceedings
\bibitem[{Yuan et~al.(2024)Yuan, Fei and Baek}]{yuanEfficientTransformerAdaptation2024}
\bibinfo{author}{Yuan, X.}, \bibinfo{author}{Fei, H.}, \bibinfo{author}{Baek, J.}, \bibinfo{year}{2024}.
\newblock \bibinfo{title}{Efficient {{Transformer Adaptation}} with {{Soft Token Merging}}}, in: \bibinfo{booktitle}{Proceedings of the {{IEEE}}/{{CVF Conference}} on {{Computer Vision}} and {{Pattern Recognition}}}, pp. \bibinfo{pages}{3658--3668}.
%Type = Article
\bibitem[{Zhang et~al.(2022)Zhang, Bao and Ma}]{zhangSelfDistillationEfficientCompact2022}
\bibinfo{author}{Zhang, L.}, \bibinfo{author}{Bao, C.}, \bibinfo{author}{Ma, K.}, \bibinfo{year}{2022}.
\newblock \bibinfo{title}{Self-{{Distillation}}: {{Towards Efficient}} and {{Compact Neural Networks}}}.
\newblock \bibinfo{journal}{IEEE Transactions on Pattern Analysis and Machine Intelligence} \bibinfo{volume}{44}, \bibinfo{pages}{4388--4403}.
\newblock \DOIprefix\doi{10.1109/TPAMI.2021.3067100}.
%Type = Article
\bibitem[{Zhang et~al.(2024a)Zhang, Li, Zhang and Guo}]{zhangMultilevelCollaborativeSelfdistillation2024a}
\bibinfo{author}{Zhang, L.}, \bibinfo{author}{Li, J.}, \bibinfo{author}{Zhang, B.}, \bibinfo{author}{Guo, Y.}, \bibinfo{year}{2024}a.
\newblock \bibinfo{title}{A multi-level collaborative self-distillation learning for improving adaptive inference efficiency}.
\newblock \bibinfo{journal}{Complex \& Intelligent Systems} \bibinfo{volume}{10}, \bibinfo{pages}{8043--8061}.
\newblock \DOIprefix\doi{10.1007/s40747-024-01572-3}.
%Type = Inproceedings
\bibitem[{Zhang et~al.(2019)Zhang, Song, Gao, Chen, Bao and Ma}]{zhangBeYourOwn2019}
\bibinfo{author}{Zhang, L.}, \bibinfo{author}{Song, J.}, \bibinfo{author}{Gao, A.}, \bibinfo{author}{Chen, J.}, \bibinfo{author}{Bao, C.}, \bibinfo{author}{Ma, K.}, \bibinfo{year}{2019}.
\newblock \bibinfo{title}{Be {{Your Own Teacher}}: {{Improve}} the {{Performance}} of {{Convolutional Neural Networks}} via {{Self Distillation}}}, in: \bibinfo{booktitle}{2019 {{IEEE}}/{{CVF International Conference}} on {{Computer Vision}} ({{ICCV}})}, \bibinfo{publisher}{IEEE}, \bibinfo{address}{Seoul, Korea (South)}. pp. \bibinfo{pages}{3712--3721}.
\newblock \DOIprefix\doi{10.1109/ICCV.2019.00381}.
%Type = Inproceedings
\bibitem[{Zhang et~al.(2023a)Zhang, Cai, Chen, Zhang, Zhang, Chen, Chang, Wang and Liu}]{zhangRobustMixtureofExpertTraining2023a}
\bibinfo{author}{Zhang, Y.}, \bibinfo{author}{Cai, R.}, \bibinfo{author}{Chen, T.}, \bibinfo{author}{Zhang, G.}, \bibinfo{author}{Zhang, H.}, \bibinfo{author}{Chen, P.Y.}, \bibinfo{author}{Chang, S.}, \bibinfo{author}{Wang, Z.}, \bibinfo{author}{Liu, S.}, \bibinfo{year}{2023}a.
\newblock \bibinfo{title}{Robust {{Mixture-of-Expert Training}} for {{Convolutional Neural Networks}}}, in: \bibinfo{booktitle}{2023 {{IEEE}}/{{CVF International Conference}} on {{Computer Vision}} ({{ICCV}})}, \bibinfo{publisher}{IEEE}, \bibinfo{address}{Paris, France}. pp. \bibinfo{pages}{90--101}.
\newblock \DOIprefix\doi{10.1109/ICCV51070.2023.00015}.
%Type = Book
\bibitem[{Zhang et~al.(2023b)Zhang, Malawade, Zhang, Li, Seong, Al~Faruque and Huang}]{zhangCARMAContextAwareRuntime2023}
\bibinfo{author}{Zhang, Y.}, \bibinfo{author}{Malawade, A.}, \bibinfo{author}{Zhang, X.}, \bibinfo{author}{Li, Y.}, \bibinfo{author}{Seong, D.}, \bibinfo{author}{Al~Faruque, M.A.}, \bibinfo{author}{Huang, S.}, \bibinfo{year}{2023}b.
\newblock \bibinfo{title}{{{CARMA}}: {{Context-Aware Runtime Reconfiguration}} for {{Energy-Efficient Sensor Fusion}}}.
\newblock \DOIprefix\doi{10.48550/arXiv.2306.15748}.
%Type = Article
\bibitem[{Zhang et~al.(2024b)Zhang, Yang, Zhang, Yue, Liu, Ou, Gong and Sun}]{zhangTMFormerTokenMerging2024b}
\bibinfo{author}{Zhang, Z.}, \bibinfo{author}{Yang, G.}, \bibinfo{author}{Zhang, Y.}, \bibinfo{author}{Yue, H.}, \bibinfo{author}{Liu, A.}, \bibinfo{author}{Ou, Y.}, \bibinfo{author}{Gong, J.}, \bibinfo{author}{Sun, X.}, \bibinfo{year}{2024}b.
\newblock \bibinfo{title}{{{TMFormer}}: {{Token Merging Transformer}} for {{Brain Tumor Segmentation}} with {{Missing Modalities}}}.
\newblock \bibinfo{journal}{Proceedings of the AAAI Conference on Artificial Intelligence} \bibinfo{volume}{38}, \bibinfo{pages}{7414--7422}.
\newblock \DOIprefix\doi{10.1609/aaai.v38i7.28572}.
%Type = Article
\bibitem[{Zhou et~al.(2017)Zhou, Gao, Wang, Yu, Wang and Chi}]{zhouEyeTrackingData2017}
\bibinfo{author}{Zhou, X.}, \bibinfo{author}{Gao, X.}, \bibinfo{author}{Wang, J.}, \bibinfo{author}{Yu, H.}, \bibinfo{author}{Wang, Z.}, \bibinfo{author}{Chi, Z.}, \bibinfo{year}{2017}.
\newblock \bibinfo{title}{Eye tracking data guided feature selection for image classification}.
\newblock \bibinfo{journal}{Pattern Recognition} \bibinfo{volume}{63}, \bibinfo{pages}{56--70}.
\newblock \DOIprefix\doi{10.1016/j.patcog.2016.09.007}.

\end{thebibliography}





% \bibitem[Lamport(1994)]{lamport94}
%   \textit{\LaTeX: a document preparation system},
%   Addison Wesley, Massachusetts,

% \end{thebibliography}
\end{document}